\PassOptionsToPackage{dvipsnames}{xcolor}
\documentclass{article}
\usepackage{iclr2026_conference,times}

\usepackage{amsmath,amsfonts,bm}

\def\eqref#1{equation~\ref{#1}}

\def\1{\bm{1}}

\def\va{{\bm{a}}}
\def\vb{{\bm{b}}}
\def\vc{{\bm{c}}}
\def\vd{{\bm{d}}}

\def\vh{{\bm{h}}}

\def\vk{{\bm{k}}}

\def\vm{{\bm{m}}}

\def\vq{{\bm{q}}}
\def\vr{{\bm{r}}}

\def\vu{{\bm{u}}}

\def\vx{{\bm{x}}}

\def\mA{{\bm{A}}}
\def\mB{{\bm{B}}}

\def\mG{{\bm{G}}}
\def\mH{{\bm{H}}}
\def\mI{{\bm{I}}}

\def\mL{{\bm{L}}}

\def\mQ{{\bm{Q}}}

\def\mU{{\bm{U}}}
\def\mV{{\bm{V}}}
\def\mW{{\bm{W}}}

\def\mY{{\bm{Y}}}

\def\mSigma{{\bm{\Sigma}}}

\DeclareMathAlphabet{\mathsfit}{\encodingdefault}{\sfdefault}{m}{sl}
\SetMathAlphabet{\mathsfit}{bold}{\encodingdefault}{\sfdefault}{bx}{n}

\usepackage{float}
\usepackage{placeins}
\usepackage{hyperref}
\usepackage{url}
\usepackage{graphicx}
\usepackage[markup=default]{changes}
\definechangesauthor[name=Me,color=blue]{ych}
\definechangesauthor[name=XH,color=red]{hxh}
\usepackage{enumitem}
\usepackage{array}
\usepackage{booktabs}
\usepackage{multirow}
\usepackage{tabularx}

\usepackage{amssymb}
\usepackage{tikz}
\usepackage{tcolorbox}
\usetikzlibrary{arrows.meta,positioning}
\colorlet{mylinkcolor}{RoyalBlue}
\colorlet{mycitecolor}{violet}
\colorlet{myurlcolor}{YellowOrange}
\definecolor{takeawaybg}{HTML}{F6F8FC}
\newtcolorbox{takeawaybox}{
  colback=takeawaybg,
  colframe=RoyalBlue!75!black,
  boxrule=0.5pt,
  arc=2pt,
  left=6pt,
  right=6pt,
  top=5pt,
  bottom=5pt,
  before skip=6pt,
  after skip=8pt,
  fontupper=\small
}
\newcommand{\takeaway}[1]{%
  \begin{takeawaybox}
    \textbf{\color{RoyalBlue!75!black}Takeaway.} #1
  \end{takeawaybox}%
}
\newcommand\myshade{85}
\hypersetup{
  linkcolor  = mylinkcolor!\myshade!black,
  citecolor  = mycitecolor!\myshade!black,
  urlcolor   = myurlcolor!\myshade!black,
  colorlinks = true,
}
\newcommand{\papertitle}{Sparse Weight Decomposition for Efficient Circuit Extraction}
\title{%
  \normalfont
  \hrule height 1pt
  \vspace{0.20in}
  {\centering\Large\bfseries \papertitle\par}
  \vspace{0.18in}
  \hrule height 1pt
  \vspace{-0.08in}
}

\author{
\makebox[0.95\textwidth][c]{{\small
\textbf{Chuanhao Yan}\textsuperscript{1}\thanks{Equal contribution.}\quad
\textbf{Xuhan Huang}\textsuperscript{1}\footnotemark[1]\quad
\textbf{Yawen Duan}\textsuperscript{2}\quad
\textbf{Zhenfei Yin}\textsuperscript{3,4}}} \\[-1pt]
\makebox[0.95\textwidth][c]{{\small
\textbf{Hang Zhao}\textsuperscript{5}\quad
\textbf{Bryan Dai}\textsuperscript{1}\thanks{Corresponding authors. Correspondence to \href{mailto:jie.fu@iquestlab.com}{jie.fu@iquestlab.com}}\quad
\textbf{Jie Fu}\textsuperscript{1}\footnotemark[2]}} \\[0pt]
\makebox[0.95\textwidth][c]{{\footnotesize\normalfont
\textsuperscript{1}IQuest Research\quad
\textsuperscript{2}Safe AI Forum\quad
\textsuperscript{3}University of Oxford\quad
\textsuperscript{4}Stanford University\quad
\textsuperscript{5}Tsinghua University}}
}

\iclrfinalcopy
\begin{document}

\maketitle
\fancyhead{}
\fancyhead[C]{\footnotesize\bfseries\papertitle}
\thispagestyle{plain}
\vspace{-0.32in}

\begin{abstract}\vspace{-0.75em}
Dense pretrained transformers do not naturally expose interpretable units for circuit extraction. 
Existing approaches obtain such units by learning auxiliary sparse representations or training sparse models, incurring substantial additional computation while potentially introducing a fidelity gap between the representation being analyzed and the original pretrained model. 
We propose Sparse Weight Decomposition (SWD), which reparameterizes pretrained linear projections by factorizing each weight matrix into two sparse factors whose shared intermediate coordinates serve as individually addressable circuit units. Without training a separate replacement network, this parametric representation supports the same scoring, selection, and ablation circuit extraction workflow used for methods that learn sparse features. 
Across single-matrix replacements, SWD matches the held-out fidelity achieved by Transcoder and other strong baselines while using less than 1\% of the data that those baselines use to train their replacements.
For matched replacement fidelity, SWD reaches the same circuit sufficiency and necessity targets with fewer active read/write edges and selected units across tasks on GPT-2, Qwen2.5, and Qwen3.5-27B.
We further show that SWD remains effective for full-model replacement of all attention and MLP weight matrices after fine-tuning the nonzero factor values. 
Finally, SWD also features a zero-data variant, allowing broader use of mechanistic interpretability analysis (\textit{e.g.}, per-step analysis).
\end{abstract}

\vspace{-0.08in}
\noindent\makebox[\textwidth][c]{%
\scriptsize
\begin{tabular}{@{}r@{\enspace}l@{\hspace{1.5em}}r@{\enspace}l@{\hspace{1.5em}}r@{\enspace}l@{}}
\textbf{Code:} & \href{https://github.com/veri-safe/SWD}{github.com/veri-safe/SWD} &
\textbf{Model:} & \href{https://huggingface.co/veri-safe/SWD}{huggingface.co/veri-safe/SWD} &
\textbf{Blog:} & \href{https://huggingface.co/spaces/veri-safe/SWD-Blog}{huggingface.co/spaces/veri-safe/SWD-Blog}
\end{tabular}}
\vspace{-0.24in}
\enlargethispage{0.25in}

\begin{figure}[H]
    \centering
    \includegraphics[
        width=0.9\linewidth,
        trim=13 18 12 13,
        clip
    ]{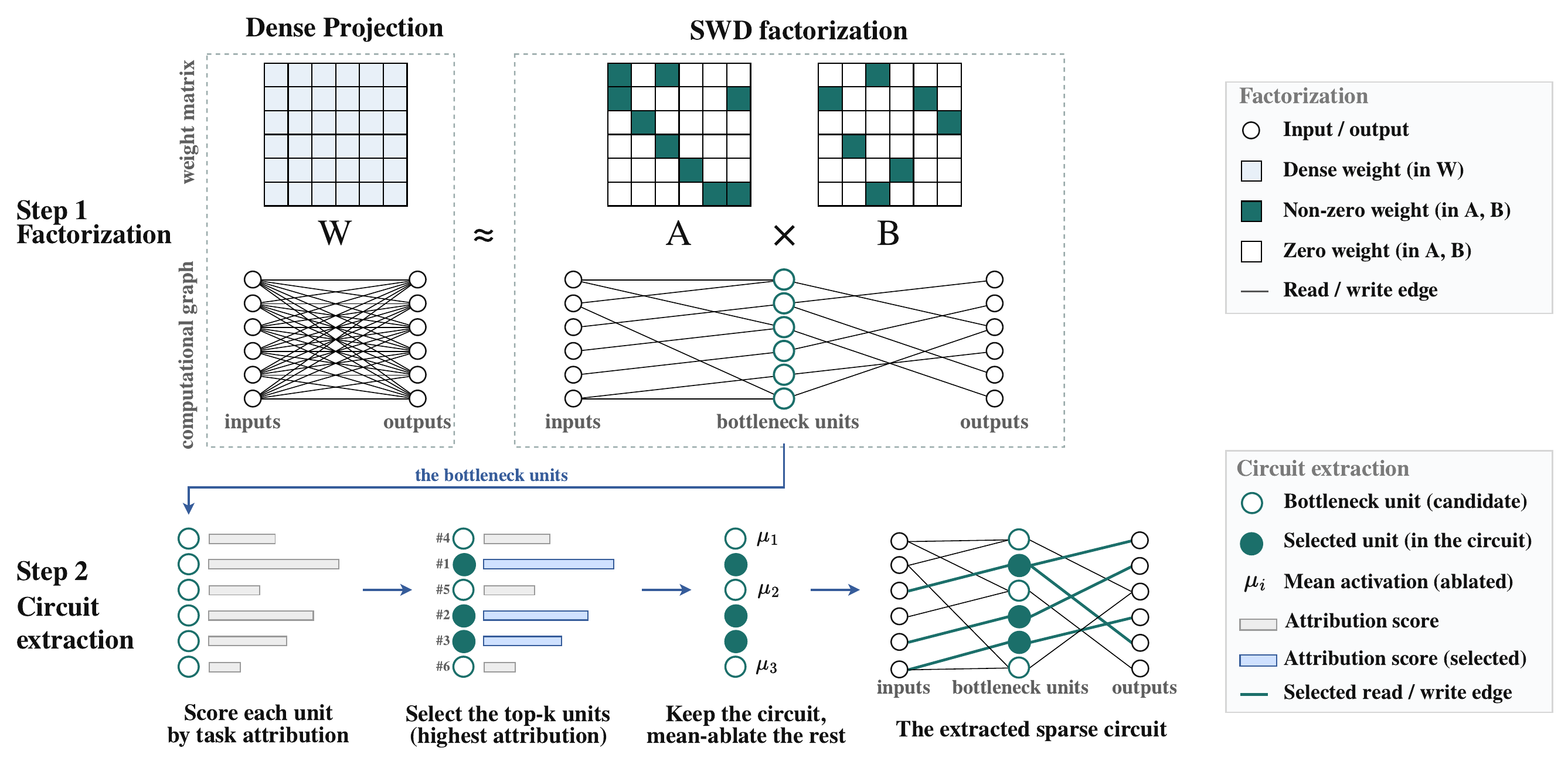}
    \caption{\footnotesize The SWD pipeline. \textbf{Step 1 (factorization).} A dense linear projection with weight matrix $\mW$ is reparameterized by two sparse factors, $\mW \approx \mA\mB$. Each shared intermediate coordinate $i$ is a \emph{bottleneck unit}: it reads from the input through $\mA_{:,i}$ and writes to the output through $\mB_{i,:}$, thereby defining the fixed rank-one path $\mA_{:,i}\mB_{i,:}$. In the graph representation, each nonzero scalar matrix entry corresponds to a directed edge whose weight is that entry, while a zero entry corresponds to no edge. Specifically, a nonzero entry $\mA_{pi}$ defines a read edge from input coordinate $p$ to unit $i$, while a nonzero entry $\mB_{iq}$ defines a write edge from unit $i$ to output coordinate $q$. For illustration, the diagram uses six bottleneck units.
    \textbf{Step 2 (circuit extraction).} These six units are scored and ranked by task attribution. The bar length represents the attribution score (longer is higher), and labels \#1--\,\#6 indicate the resulting ranking, with \#1 ranked highest. The top-$k$ units are selected for the circuit; in this illustration, $k=3$, so the three highest-ranked units are retained. The activation of each unselected unit $i$ is replaced by its mean $\mu_i$.
    }    
    \label{fig:swd_pipeline}
\end{figure}

\section{Introduction}

Circuit analysis seeks a small set of internal units that is sufficient to reproduce a behavior and necessary in the sense that ablating it predictably degrades that behavior \citep{conmy2023automated,syed2023attribution,bhaskar2024edge}.
Pretrained transformers, however, do not expose such units: they implement task behavior through dense linear projections, where individual neurons can be polysemantic and task-relevant effects are distributed across many activations and parameters \citep{elhage2022toy,cunningham2023sparse}.
This leaves a gap between the sparse causal explanations we seek and the dense computational substrate we aim to analyze.


Existing approaches obtain interpretable units for circuit analysis in several ways. Sparse autoencoders and Transcoders learn feature dictionaries or replacement modules in activation space; parameter-decomposition methods learn components of the model's weights; and sparse pretraining builds sparse connectivity into the model during pretraining \citep{cunningham2023sparse,dunefsky2024transcoders,braun2025apd,bushnaq2025spd,bushnaq2026interpreting,gao2025weightsparse}. Each of these three families provides useful units for analysis, but requires training or optimizing an additional representation or model beyond the dense pretrained checkpoint. The reported overheads can be substantial. For example, weight-sparse models require $100$--$1000\times$ more training and inference compute than dense models of comparable capability \citep{gao2025weightsparse}.

Alternatively, a natural question is whether circuit units can be obtained directly from a pretrained checkpoint, avoiding the cost of learning auxiliary representations altogether. NaNA \citep{xue2026svd} applies exact SVD directly to MLP weights, ranks the resulting components by their contribution to a specified target token, and shows that a small set of top-ranked components can recover the target prediction without training an auxiliary representation. However, each retained component still has dense input and output vectors, so retaining only a few components does not necessarily yield a compact circuit in terms of active connections. Weight-sparse transformers are trained from scratch with sparse weights and activations and show that sparse connectivity can yield substantially smaller and more human-understandable task circuits at comparable pretraining loss \citep{gao2025weightsparse}. This result suggests that sparse connectivity is a useful structural constraint for circuit analysis. We therefore ask whether weight-side sparse connectivity can be introduced after pretraining by reparameterizing an existing dense checkpoint. To pursue this question, we draw on Double Sparse Factorization from model compression, which approximates a dense matrix as the product of two sparse matrices \citep{boza2025double}. We evaluate whether this sparse decomposition can preserve the original model's behavior while yielding task circuits with fewer active connections.

\textbf{Our key idea is to obtain circuit units directly from pretrained weights via \emph{sparse factorization}, avoiding the cost of training an auxiliary representation.} Specifically, we introduce Sparse Weight Decomposition (SWD), which reparameterizes a pretrained dense linear projection with weight matrix $\mW$ as $\mW\approx \mA\mB$, where $\mA$ and $\mB$ are sparse factor matrices. Each intermediate coordinate $i$ is a \emph{bottleneck unit}: it reads through $\mA_{:,i}$ and writes through $\mB_{i,:}$, defining a fixed rank-one path $\mA_{:,i}\mB_{i,:}$ with sparse read and write connections. Its scalar activation can be independently scored, selected, and ablated using the same circuit-extraction workflow applied to learned sparse features~\citep{conmy2023automated,bhaskar2024edge,miller2024transformercircuitfaithfulnessmetrics,syed2023attribution,wang2022interpretability}.

In our experiments, we evaluate both replacement fidelity (how closely each replacement preserves dense-model behavior before circuit extraction) and the circuit cost-quality tradeoff (how many selected units or active read/write edges are required to reach each sufficiency or necessity target). We first replace one MLP matrix in each of GPT-2 Small~\citep{radford2019language} and three Qwen2.5~\citep{qwen2025qwen25technicalreport} models of different sizes, comparing SWD with activation- and parameter-space baselines under matched pre-pruning cross-entropy (CE) loss, as detailed in Section~\ref{sec:single_layer_replacement}. To test scalability to larger models, we further apply the same single-matrix replacement and circuit-extraction protocol to Qwen3.5-27B~\citep{qwen35blog}. We simultaneously replace all 48 attention and MLP weight matrices across the 12 transformer blocks of GPT-2 Small to test whether SWD works when local approximation errors accumulate (Section~\ref{sec:full-model-replacement}). As controls, exact full-rank dense factorizations reproduce the original weight matrix without error, testing whether low circuit cost comes from factorization alone or from sparse read/write connectivity (Section~\ref{sec:exact-dense-control-ablation}). Finally, a zero-data variant constructs factors from weights alone, omitting calibration activations (input activations collected on calibration data), to test whether the model alone can be turned into useful bottleneck units (Section~\ref{sec:zero_data}).

Our contributions are as follows:
\begin{itemize}
    \item We repurpose and evaluate sparse weight factorization as a circuit-extraction method: its bottleneck units can be scored and pruned without training a separate replacement network.
    \item Compared with baselines such as Transcoder and VPD, SWD reaches matched replacement fidelity using less than $1\%$ as much data and, at that fidelity, SWD requires fewer active read/write edges to attain the same sufficiency and necessity thresholds.
    \item SWD scales across model sizes and, after fixed-support fine-tuning, remains effective when all attention and MLP weight matrices are replaced simultaneously. It also admits a variant that requires no calibration data.
\end{itemize}


\section{Methodology}
\label{sec:methodology}


\subsection{Sparse Weight Decomposition}
\label{sec:swd-factorization}
\paragraph{Sparsity on the edges.} A dense matrix has many factorizations, but an exact dense reparameterization such as Singular Value Decomposition (SVD) still leaves every bottleneck unit connected to almost every input and output dimension. Motivated by evidence that weight-sparse models preserve more tractable circuits \citep{gao2025weightsparse}, SWD places sparsity on the factorization's read and write \emph{edges} (weights). Each bottleneck unit therefore reads from and writes to only a small set of input/output dimensions. Section~\ref{sec:exact-dense-control-ablation} shows that the resulting cost-quality tradeoff advantage does not come from exact factorization alone.

\paragraph{Objective.} For a dense matrix $\mW \in \mathbb{R}^{d_{\mathrm{in}} \times d_{\mathrm{out}}}$, SWD seeks sparse factors $\mA \in \mathbb{R}^{d_{\mathrm{in}} \times m}$ and $\mB \in \mathbb{R}^{m \times d_{\mathrm{out}}}$ such that $\mW \approx \widehat{\mW}=\mA\mB$, where $m$ is the number of bottleneck units. Table~\ref{tab:notation} in Appendix~\ref{app:notation} summarizes the recurring notation used throughout the paper. Under a fixed nonzero budget $K$, the sparse product cannot in general preserve the action of $\mW$ equally well in every input direction, so the optimization must decide where to allocate approximation error \citep{boza2025double}. The Frobenius objective $\|\mW-\mA\mB\|_F^2$ nevertheless weights all directions equally so that at the same sparsity, it can spend limited fitting capacity on directions rarely encountered by the model, and leave avoidably large error on the activations the model actually processes (Section \ref{sec:zero_data}). Our default objective therefore minimizes reconstruction error on calibration activations:
\[
    \min_{\mA,\mB}\;
    \mathbb{E}_{\vh \sim \mathcal{D}_{\mathrm{cal}}}
    \left\|\vh\mW - \vh\mA\mB\right\|_2^2
    \quad
    \mathrm{s.t.}
    \quad
    \|\mA\|_0 + \|\mB\|_0 \leq K,
\]
where $\vh$ is a calibration-activation vector and $K$ is the total number of nonzero factor entries. In experiments, we report $K$ as a sparsity level $s=1 - \frac{K}{\|\mW\|_0}$ (larger $s$, smaller $K$). Since
\begin{align*}
    \mathbb{E}_{\vh \sim \mathcal{D}_{\mathrm{cal}}}\|\vh\mW - \vh\mA\mB\|_2^2 = \mathbb{E}_{\vh \sim \mathcal{D}_{\mathrm{cal}}}(\mW-\mA\mB)^\top(\vh^\top\vh)(\mW-\mA\mB) \\
= (\mW-\mA\mB)^\top\mathbb{E}_{\vh \sim \mathcal{D}_{\mathrm{cal}}}[\vh^\top\vh](\mW-\mA\mB),
\end{align*}
 the Gram matrix $\mG=\mathbb{E}_{\vh\sim\mathcal{D}_{\mathrm{cal}}}[\vh^\top\vh]$ directs the limited fitting capacity toward input directions with greater mass under the model's activation distribution. When calibration activations are unavailable, we also have a zero-data variant: set $\mG=\mI$, reducing the objective to $\|\mW-\mA\mB\|_F^2$, which we evaluate in Section~\ref{sec:zero_data}.

\paragraph{Solver.} Because the $\ell_0$-constrained objective is nonconvex and NP-Hard, we approximately optimize it using the Double Sparse Factorization (DSF) heuristic \citep{boza2025double}. We split the total budget into fixed factor-specific budgets, $K_A+K_B=K$, and alternate between updating $\mA$ with $\mB$ fixed and updating $\mB$ with $\mA$ fixed. Each block update is an $\ell_0$-constrained regression problem, which DSF solves through Alternating Direction Method of Multipliers (ADMM) \citep{boyd2011distributed} iterations consisting of: 1. a regularized least-squares update, 2. A hard-thresholding projection onto the prescribed nonzero budget, and 3. a dual-variable update. After the prescribed outer iterations, DSF freezes the per-matrix supports and the matrix $\mA$ and locally refits the surviving entries of the matrix $\mB$ against the activation-weighted reconstruction objective.
For a single matrix, this is the entire SWD process. When many weight matrices are replaced simultaneously, however, their local approximation errors can accumulate. In the full-model replacement experiment, we therefore minimize next-token cross-entropy over the nonzero factor values while keeping all other pretrained parameters fixed. Here, \emph{fixed support} means that the zero/nonzero patterns of $\mA$ and $\mB$ are frozen, so optimization updates only values at existing nonzero entries. This preserves the sparse read/write connectivity and its corresponding edge count. We refer to the replacement after fixed-support fine-tuning as SWD-FT and evaluate it in Section~\ref{sec:full-model-replacement}; implementation and hyperparameter details are given in Appendix~\ref{app:swd-optimization}.

\takeaway{SWD turns a dense matrix into two sparse factors so that every bottleneck unit becomes a rank-one path with few read and write edges.}

\subsection{Circuit Extraction from Sparse Factors}

The factorized replacement can be written as a sum of additive read--write paths. For an input-activation vector $\vh$, define
\[
    z_i(\vh)=\vh\mA_{:,i},
    \qquad
    \vh\widehat{\mW}=\vh\mA\mB=\sum_{i=0}^{m-1}z_i(\vh)\mB_{i,:}.
\]
As shown in Figure \ref{fig:swd_pipeline}, bottleneck unit $i$ reads $\vh$ through the sparse vector $\mA_{:,i}$ and writes the resulting scalar along the sparse vector $\mB_{i,:}$; their supports therefore specify which input and output coordinates the path touches. Moreover, replacing $z_i(\vh)$ changes only the $i$th term in the sum, leaving the other paths unchanged. This makes each bottleneck unit individually addressable for circuit interventions.

We rank bottleneck units by a task-margin attribution score and evaluate nested top-$k$ prefixes (Appendix~\ref{app:scoring-protocol}). Given this ranking, a candidate circuit is a subset $S\subseteq\{0,\dots,m-1\}$. Let $\mu_i$ be the reference activation for bottleneck unit $i$. Following the mean-ablation convention used in prior circuit analysis \citep{wang2022interpretability}, we set $\mu_i$ to be the mean activation on the circuit extraction training split. We form two complementary interventions:
\[
    z_i^{\mathrm{keep}}(\vh;S)=
    \begin{cases}
        z_i(\vh), & i\in S,\\
        \mu_i, & i\notin S,
    \end{cases}
    \qquad
    z_i^{\mathrm{abl}}(\vh;S)=
    \begin{cases}
        \mu_i, & i\in S,\\
        z_i(\vh), & i\notin S.
    \end{cases}
\]
The keep intervention tests sufficiency: how much task behavior is retained by only keeping those units in $S$. The ablation intervention tests necessity: how much task behavior is lost by removing those units in $S$. In both cases, the intervened output is obtained by substituting the corresponding activations into $\sum_i z_i \mB_{i,:}$.

These interventions are comparable across methods only when the unpruned replacements share the same pre-pruning behavior; otherwise scores are measured against different reference computations. The detailed replacement-fidelity matching rule and the common attribution-based $S$ selection protocol are specified in Section~\ref{sec:circuit-extraction-protocol}.

\takeaway{Each bottleneck is an independently ablatable path. We rank paths on circuit training examples, and then test sufficiency and necessity on the test set.}

\section{Experiments}
Our experiments track two criteria throughout: replacement fidelity and circuit cost-quality tradeoff, which is defined in Section~\ref{sec:circuit-extraction-protocol}. We first replace one MLP matrix in each of GPT-2 Small and the Qwen2.5 models from 0.5B to 3B with SWD, Transcoder, and VPD variants, then extend this single-matrix comparison to Qwen3.5-27B to test scalability, and evaluate sufficiency and necessity under matched pre-pruning cross-entropy. We then report full-model replacement results, isolate the contribution of sparse read/write edges from exact dense controls, evaluate a zero-data variant, and qualitatively inspect selected units and circuit.

\subsection{Baselines}
We compare SWD with three approaches that provide units for circuit analysis: \textbf{Transcoders} learn sparse activation-space replacements for MLP computations \citep{dunefsky2024transcoders}; \textbf{sparse-pretrained models} impose weight sparsity during pretraining \citep{gao2025weightsparse}; and \textbf{VPD} learns parameter components with an input-dependent selector \citep{bushnaq2026interpreting}. We evaluate two VPD configurations. \textbf{VPD-KL} denotes the original VPD configuration, whose fidelity term is model-level Kullback--Leibler (KL) divergence. \textbf{VPD-Recon-CI} retains the same component parameterization and causal-importance (CI) selector but replaces this term with local activation reconstruction. VPD-KL does not reach the fidelity required for CE-matched circuit comparison in our single-matrix replacement experiments, so we report it only for replacement quality; VPD-Recon-CI is used for circuit comparisons (Appendix~\ref{app:vpd-recon-ci-setting} and~\ref{app:original-vpd-kl-longtail}). Since sparse pretraining produces an entire sparse model rather than a single-matrix replacement, we compare against it only in the full-model replacement experiment (Section~\ref{sec:full-model-replacement}). Table~\ref{tab:baseline-comparison} summarizes how these methods differ in representation, sparsity, optimization, data requirements, and applicable scope. Full configurations are collected in Appendix~\ref{app:baseline-settings}.

\begin{table}[H]
    \centering
    \footnotesize
    \setlength{\tabcolsep}{2.5pt}
    \renewcommand{\arraystretch}{1.35}
    \begin{tabularx}{\linewidth}{>{\raggedright\arraybackslash}p{0.175\linewidth}!{\vrule width 0.49pt}*{4}{>{\raggedright\arraybackslash}X}}
        \toprule
        Property & \textbf{SWD (ours)} & \textbf{Transcoder} & \textbf{VPD} & \textbf{Sparse pretrain} \\
        \midrule
        Separate model training (additional optimization cost)
            & \textbf{No}; fit sparse factors \textbf{post hoc}
            & \textbf{Yes}; train a replacement network for the pretrained model
            & \textbf{Yes}; train a decomposition and input-dependent selector for the pretrained model
            & \textbf{Yes}; train the complete sparse model \textbf{from scratch} \\
        Representation and unit of analysis
            & Two sparse factors; one bottleneck unit is a \textbf{rank-one read--write path}
            & A replacement network; one unit is a \textbf{hidden feature}
            & A parameter decomposition; one unit is a \textbf{rank-one parameter component}
            & A sparse pretrained model; one unit is a native \textbf{hidden channel} \\
        Type of sparsity
            & \textbf{Weight sparsity} in the factor matrices
            & \textbf{Activation sparsity} in hidden features
            & \textbf{Input-dependent component selection} over parameter components
            & \textbf{Weight sparsity} throughout the model; activation sparsity is disabled in our run \\
        Optimization objective
            & Reconstruct target weights or their outputs on calibration data under fixed factor sparsity
            & Reconstruct original MLP outputs while encouraging sparse feature activations
            & Match model outputs or replaced-matrix outputs, with parameter reconstruction and sparse component use
            & Minimize language-model loss while progressively sparsifying weights \\
        Data required
            & \textbf{None or a small amount} of calibration data\textsuperscript{*}
            & Text used to train the replacement network
            & Text used to train the decomposition and selector
            & \textbf{Full pretraining corpus} \\
        Applicable scope
            & MLP and attention matrices; from one matrix to all model projections
            & MLP computations; one MLP at a time in our comparisons
            & MLP and attention matrices; from one matrix up to full-model replacement
            & Whole model during pretraining \\
        \bottomrule
    \end{tabularx}
    \vspace{2pt}

    \begin{minipage}{\linewidth}
        \scriptsize
        \textsuperscript{*}At matched replacement CE, SWD uses \textbf{less than 1\%} of the data used by the corresponding trained baseline in our experiments. This greater data use does not consistently reduce circuit cost: SWD typically reaches the same sufficiency or necessity target with fewer active edges; see Sections~\ref{sec:single_layer_replacement} and~\ref{sec:full-model-replacement}.
    \end{minipage}
    \caption{Comparison of the sparse intervention methods evaluated in this paper.}
    \label{tab:baseline-comparison}
\end{table}

\subsection{Evaluation Protocol}
\label{sec:circuit-extraction-protocol}
Our evaluation protocol largely follows the task-based circuit-discovery setup used by ACDC/auto-circuit and Edge-Pruning \citep{conmy2023automated,bhaskar2024edge,miller2024transformercircuitfaithfulnessmetrics}, together with first-order attribution scoring from attribution patching \citep{syed2023attribution}. Evaluation proceeds in two stages. First, we evaluate replacement fidelity using CE delta (replacement CE minus dense-model CE), supplemented where applicable by KL and activation relative mean-squared error (relative MSE; the relative error between the dense and replacement outputs of the replaced projection). Detailed definitions of the companion metrics are provided in Appendix~\ref{app:original-vpd-kl-longtail}. All methods use FineWeb-Edu for training or calibration, with a disjoint split reserved for evaluation \citep{lozhkov2024fineweb-edu}. We report token--CE loss curves to evaluate data efficiency.
Since circuit scores are computed relative to each method's own replacement, unless explicitly stated otherwise, we compare circuit performance only between replacements whose pre-pruning CE differs by at most $0.001$. Second, for circuit extraction, we use greater-than, indirect-object identification (IOI), docstring, and gendered-pronoun task families from public Edge-Pruning and auto-circuit datasets \citep{bhaskar2024edge,conmy2023automated,miller2024transformercircuitfaithfulnessmetrics}. For each task $\tau$, we compute unit scores and mean-ablation statistics on the circuit extraction training split (\texttt{circuit\_train}). The resulting ranking defines a nested family of top-$k$ prefixes $S_k$. We then evaluate these fixed prefixes on the held-out test split (\texttt{circuit\_test}) and report held-out sufficiency and necessity as functions of circuit cost. Data sources, split sizes, and the Qwen task subsets are detailed in Appendix~\ref{app:data-sources}.

For each method, we apply this common selection protocol to the candidate units exposed by its representation. Units are ranked by positive first-order task-margin attribution (Appendix~\ref{app:scoring-protocol}). Ablated units are replaced by their mean activation on \texttt{circuit\_train}; thus $M_{\mathrm{keep}}(S)$ retains the selected units and mean-ablates the remainder, whereas $M_{\mathrm{abl}}(S)$ mean-ablates the selected units and retains the remainder. Following prior causal circuit analyses \citep{wang2022interpretability}, we use mean ablation for the main comparisons because it removes input-dependent variation while preserving each unit's average activation level. Because circuit evaluations can be sensitive to the ablation methodology, including the value assigned to ablated activations \citep{miller2024transformercircuitfaithfulnessmetrics}, we also test whether our conclusions depend on this choice. Appendix~\ref{app:gpt2-zero-ablation} keeps the checkpoints, unit rankings, and top-$k$ sets fixed and repeats the GPT-2 single-matrix evaluation using zero rather than mean ablation. Let $g_{\tau}(M,\vx)$ be the task logit margin for example $\vx$: the model's average logit for correct task answers minus its average logit for distractor answers. We define $Q_{\tau}(M;\mathcal{D})=\mathbb{E}_{\vx\in\mathcal{D}}[g_{\tau}(M,\vx)]$ as the average task score of model $M$ on split $\mathcal{D}$, with $M_{\mathrm{unpruned}}$ denoting the replacement before circuit pruning. We report method-relative sufficiency,
\[
    \mathrm{Suff}_{\tau}(S;\mathcal{D})
    =
    \frac{
        Q_{\tau}(M_{\mathrm{keep}}(S);\mathcal{D})
    }{
        Q_{\tau}(M_{\mathrm{unpruned}};\mathcal{D})
    },
\]
and
\[
    \mathrm{NecDrop}_{\tau}(S;\mathcal{D})
    =
    Q_{\tau}(M_{\mathrm{unpruned}};\mathcal{D})
    -
    Q_{\tau}(M_{\mathrm{abl}}(S);\mathcal{D}).
\]
We measure circuit size by selected units and active edges. An active edge is a nonzero weight in the read or write vector of a selected unit. For a selected set $S$, let $\mathbf{r}_i$ and $\mathbf{w}_i$ denote the read and write weight vectors of unit $i$, and let $S_{\mathrm{eff}}\subseteq S$ contain the selected units with at least one nonzero read edge and one nonzero write edge. We define
\[
    C_{\mathrm{unit}}(S)=|S|,
    \qquad
    C_{\mathrm{edge}}(S)
    =
    \sum_{i \in S_{\mathrm{eff}}}
    \left(
        \|\mathbf{r}_i\|_0 + \|\mathbf{w}_i\|_0
    \right),
\]
The active-edge count describes the structural connectivity of $S$ and is the same whether $S$ is retained for sufficiency or ablated for necessity. For each sufficiency or necessity value, we report the minimum selected units or minimum active edges among the evaluated top-$k$ sets that reach that value. Method-specific definitions are given in Appendices~\ref{app:swd-optimization} and~\ref{app:baseline-settings}.

\subsection{Single-Matrix Replacement}
\label{sec:single_layer_replacement}
\begin{figure}[t]
    \centering
    \includegraphics[width=0.82\linewidth]{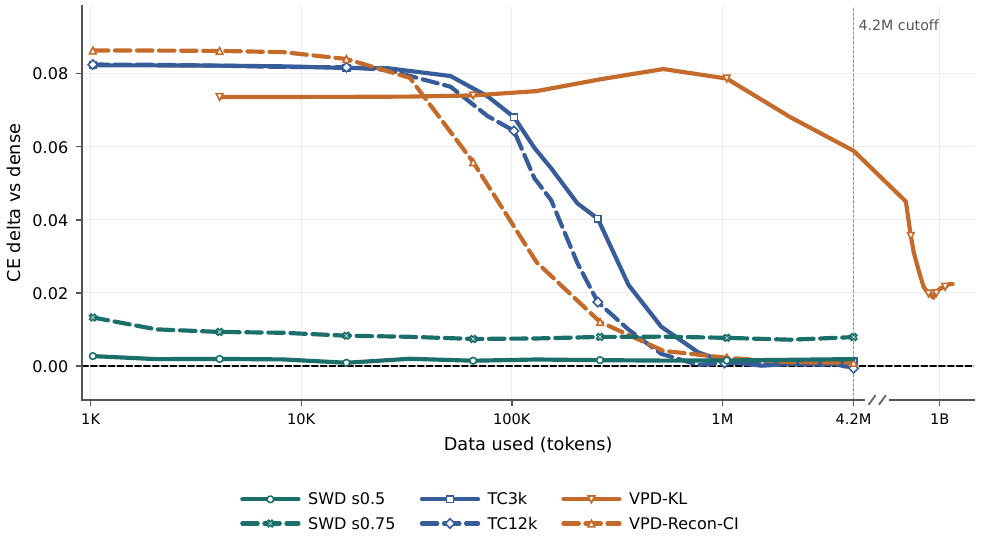}
    \caption{GPT-2 single-matrix replacement quality. SWD reaches low CE delta with far less data; VPD-KL is included for replacement-quality context and excluded from circuit extraction evaluation. See Appendix~\ref{app:original-vpd-kl-longtail} for companion metrics and settings.}
    \label{fig:gpt2-cproj-ce-delta}
\end{figure}

We begin by replacing the GPT-2 Small layer-8 MLP output projection, \texttt{mlp.c\_proj}, with each method's corresponding sparse representation. Figure~\ref{fig:gpt2-cproj-ce-delta} plots CE delta from the dense model against data used. SWD reaches low replacement error with substantially less data: SWD ($s{=}0.5$) reaches low CE delta after a few thousand tokens versus roughly $10^6$ for the baselines, whereas VPD-KL improves only after much longer optimizer replay and then plateaus. Because its CE cannot be matched to the other methods, we retain VPD-KL as a replacement-quality reference but exclude it from circuit extraction evaluation. Appendix~\ref{app:original-vpd-kl-longtail} documents activation relative MSE, KL, and full settings.

\begin{figure}[t]
    \centering
    \includegraphics[width=\linewidth]{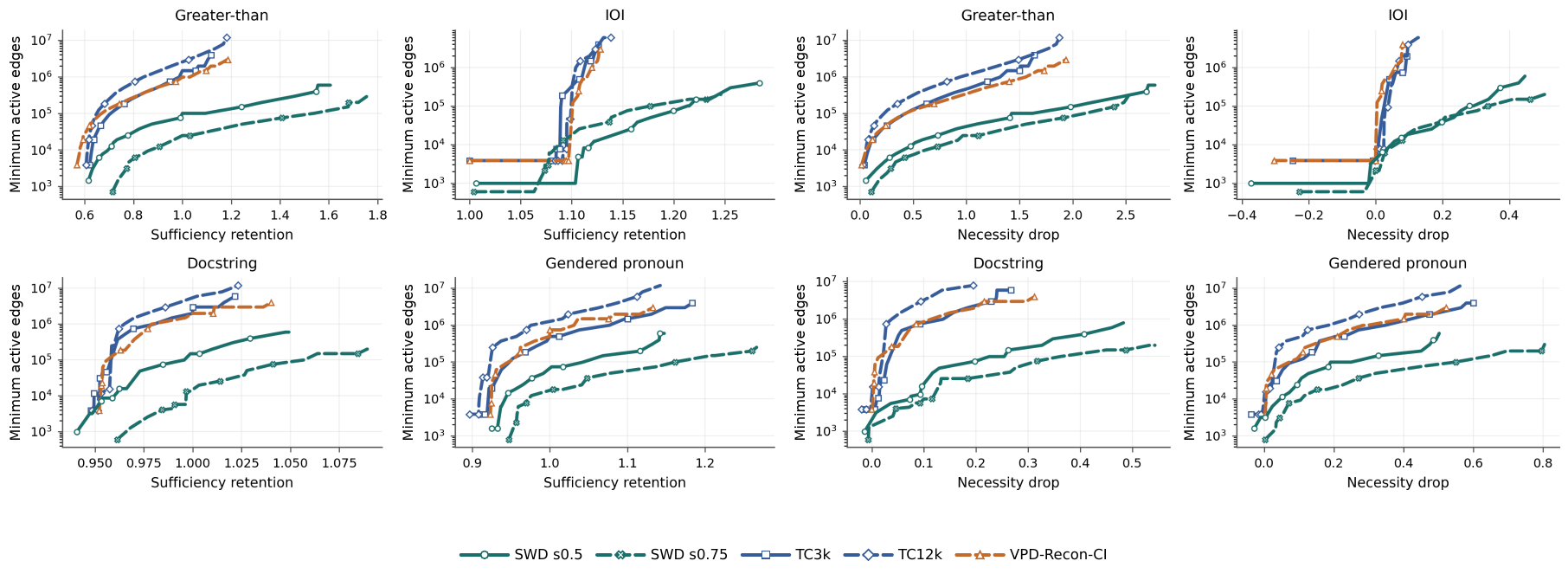}
    \caption{GPT-2 single-matrix circuit results. SWD reaches the target thresholds with markedly fewer active edges than the Transcoder and VPD-Recon-CI baselines.}
    \label{fig:gpt2-cproj-circuit}
\end{figure}

Figure~\ref{fig:gpt2-cproj-circuit} compares performance at matched pre-pruning CE, with one sparser SWD setting shown separately as a non-matched control. Across all four tasks, SWD attains the same sufficiency and necessity targets with substantially fewer active edges than Transcoder and VPD-Recon-CI. This advantage is not specific to mean ablation: when the checkpoints, unit rankings, and top-$k$ sets are held fixed and zero ablation is used instead, SWD continues to require fewer active edges across all four tasks (Appendix~\ref{app:gpt2-zero-ablation}). Selected-unit circuit results and settings are in Appendix~\ref{app:gpt2-cproj-unit-companion} and~\ref{app:baseline-settings}.

The two SWD settings expose a trade-off between replacement fidelity and circuit size. SWD with $s{=}0.5$ achieves lower replacement CE and is therefore used in the CE-matched comparison, whereas the more aggressive $s{=}0.75$ setting incurs a modest fidelity penalty but requires fewer active edges to retain the same task behavior or induce the same performance drop when ablated. Thus, increasing factor sparsity can reduce circuit size even when it slightly reduces replacement fidelity.

\begin{figure}[t]
    \centering
    \includegraphics[width=0.98\linewidth]{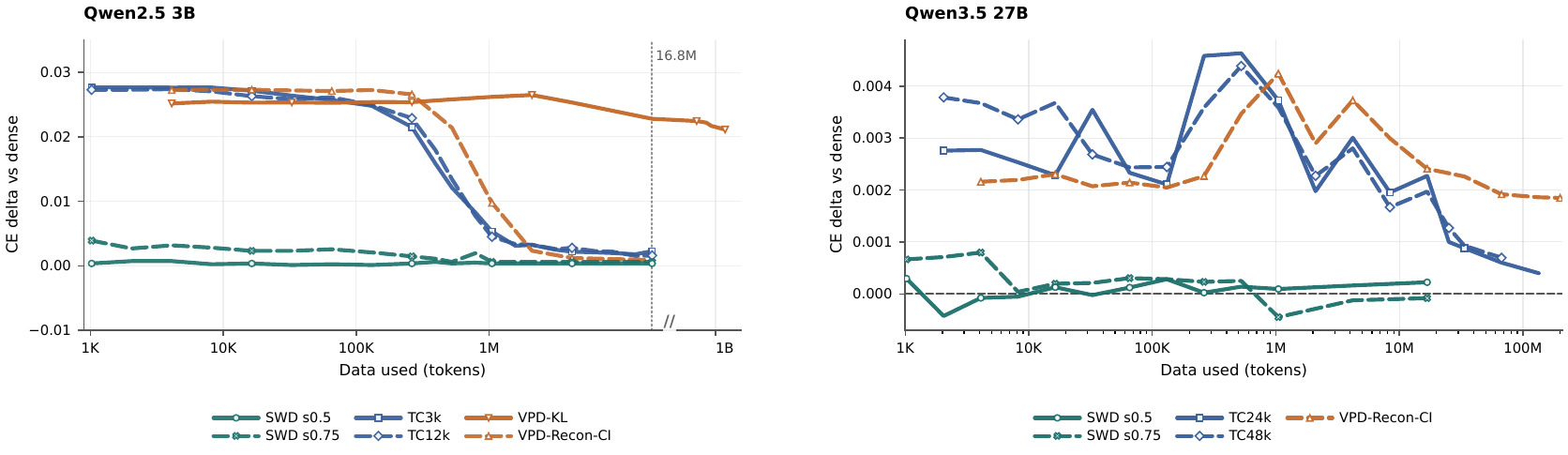}
    \caption{Qwen single-matrix replacement quality. Left: Qwen2.5-3B. Right: Qwen3.5-27B. SWD reaches low CE delta with substantially fewer tokens than the trained baselines. Results for the other Qwen2.5 model sizes are reported in Appendix~\ref{app:qwen-vpdrecon-extension}.}
    \label{fig:qwen-ce-scaling}
\end{figure}

\begin{figure}[t]
    \centering
    \includegraphics[width=\linewidth]{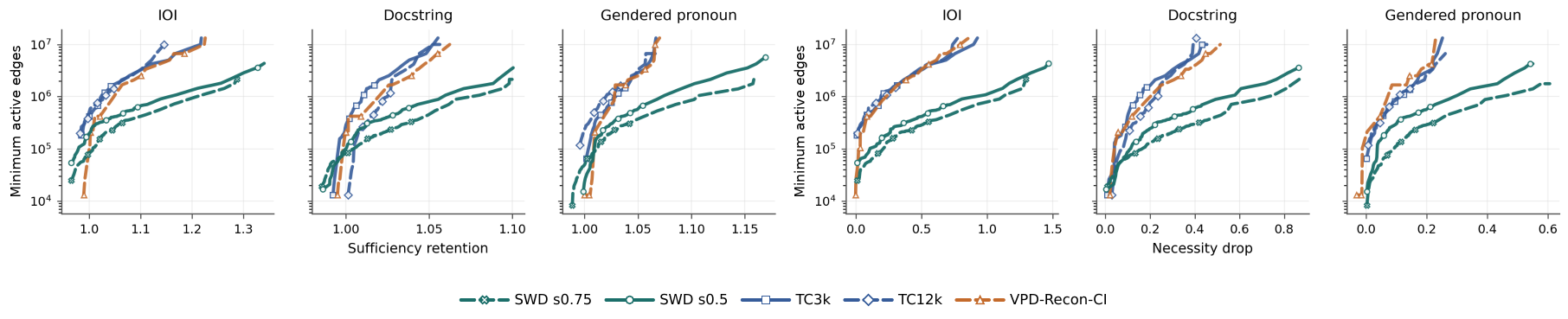}
    \caption{Qwen2.5 3B single-matrix circuit results. SWD reaches the same sufficiency and necessity thresholds at substantially fewer active edges. The 0.5B/1.5B results and their settings appear in Appendix~\ref{app:qwen-vpdrecon-extension}.}
    \label{fig:qwen-circuit}
\end{figure}

\begin{figure}[t]
    \centering
    \includegraphics[width=\linewidth]{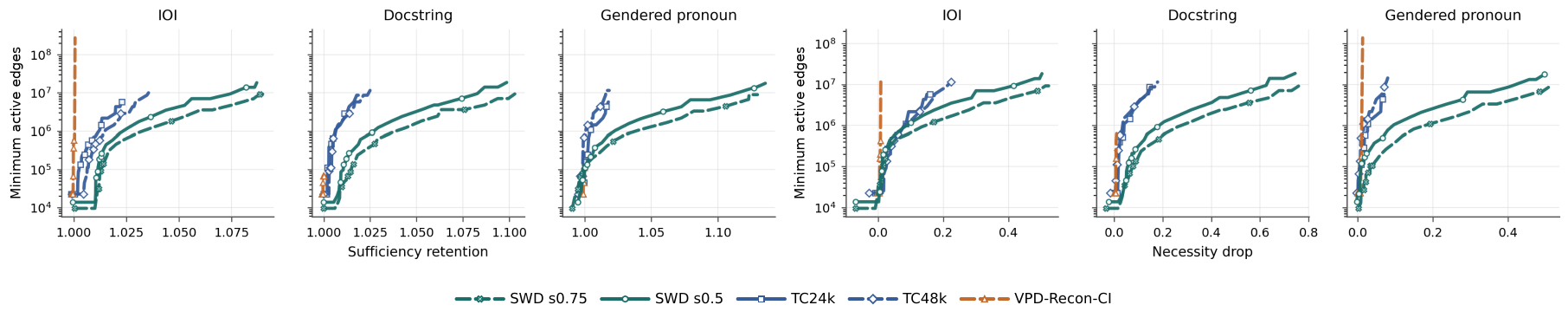}
    \caption{Qwen3.5-27B single-matrix circuit results. The vertical axis reports the minimum active edges. Checkpoint settings and the selected-unit companion are reported in Appendix~\ref{app:qwen35-circuit}.}
    \label{fig:qwen35-circuit}
\end{figure}

We repeat the single-matrix comparison on Qwen2.5 models from 0.5B to 3B and extend it to Qwen3.5-27B. Figure~\ref{fig:qwen-ce-scaling} reports replacement quality on Qwen2.5-3B and Qwen3.5-27B. In both models, SWD reaches low CE delta with substantially fewer tokens than the trained baselines. Figures~\ref{fig:qwen-circuit} and~\ref{fig:qwen35-circuit} show the same circuit-cost trend observed on GPT-2: SWD reaches comparable sufficiency and necessity with fewer active edges, extending this advantage to Qwen2.5-3B and Qwen3.5-27B. Results for the other Qwen2.5 model sizes are reported in Appendix~\ref{app:qwen-vpdrecon-extension}, with further details for Qwen3.5-27B in Appendix~\ref{app:qwen35-circuit}.

In the experiments above, each method replaces only one MLP matrix. Appendix~\ref{app:fullmlp-alignment} additionally reports an MLP-replacement experiment, where the standard Transcoder replaces the entire MLP and SWD replaces both MLP weight matrices with sparse factorizations while retaining the original GELU.

\takeaway{SWD requires substantially less data than trained replacement baselines to reach matched replacement fidelity, while achieving better circuit sufficiency and necessity with fewer selected units and active edges.}

\subsection{Full-Model Replacement}
\label{sec:full-model-replacement}
In this experiment, we replace all 48 attention and MLP weight matrices across the 12 transformer blocks of GPT-2 Small. The embeddings, layer-normalization modules, nonlinearities, and LM head remain unchanged. We compare with weight-sparse pretraining \citep{gao2025weightsparse} at an approximately matched budget of $26.77$M active weights (about $68\%$ sparsity relative to the dense transformer-body weight matrices); the exact training recipe and accounting are given in Appendix~\ref{app:sparse-pretrain-setting}.
Approximation error now accumulates across layers: the sparse factorization reaches held-out CE $3.90$. We therefore apply fixed-support fine-tuning, updating only the nonzero factor values while keeping the sparse supports (and hence the edge count) fixed. The resulting SWD-FT replacement reaches CE $3.44$, slightly below the matched sparse-pretraining checkpoint's $3.45$, while using under $1\%$ of its token budget ($20.6$M tokens in total, i.e.\ $4.19$M for factorization and $16.38$M for fixed-support fine-tuning, versus $2.884$B tokens for sparse pretraining). Table~\ref{tab:gpt2-fullmodel-replacement-summary} summarizes this comparison; Appendix~\ref{app:gpt2-fullmodel-ce-curve} gives the full curve.

\begin{table}[t]
    \centering
    \small
    \begin{tabular}{p{0.17\linewidth}p{0.32\linewidth}p{0.22\linewidth}p{0.13\linewidth}}
        \hline
        Method & Data used & Active nonzero weights & LM CE loss \\
        \hline
        SWD & $4.19$M calibration tokens & $26.77$M & $3.90$ \\
        SWD-FT & $4.19$M calibration + $16.38$M fine-tuning tokens & $26.77$M & $3.44$ \\
        Sparse pretrain & $2.884$B sparse-pretraining tokens & $26.77$M & $3.45$ \\
        \hline
    \end{tabular}
    \caption{GPT-2 Small full-model replacement summary at approximately matched transformer-body active nonzeros. SWD-FT keeps the SWD support fixed and refits only nonzero factor entries, achieving CE comparable to that of the sparse-pretrained model.}
    \label{tab:gpt2-fullmodel-replacement-summary}
\end{table}

\begin{figure}[t]
    \centering
    \includegraphics[width=\linewidth]{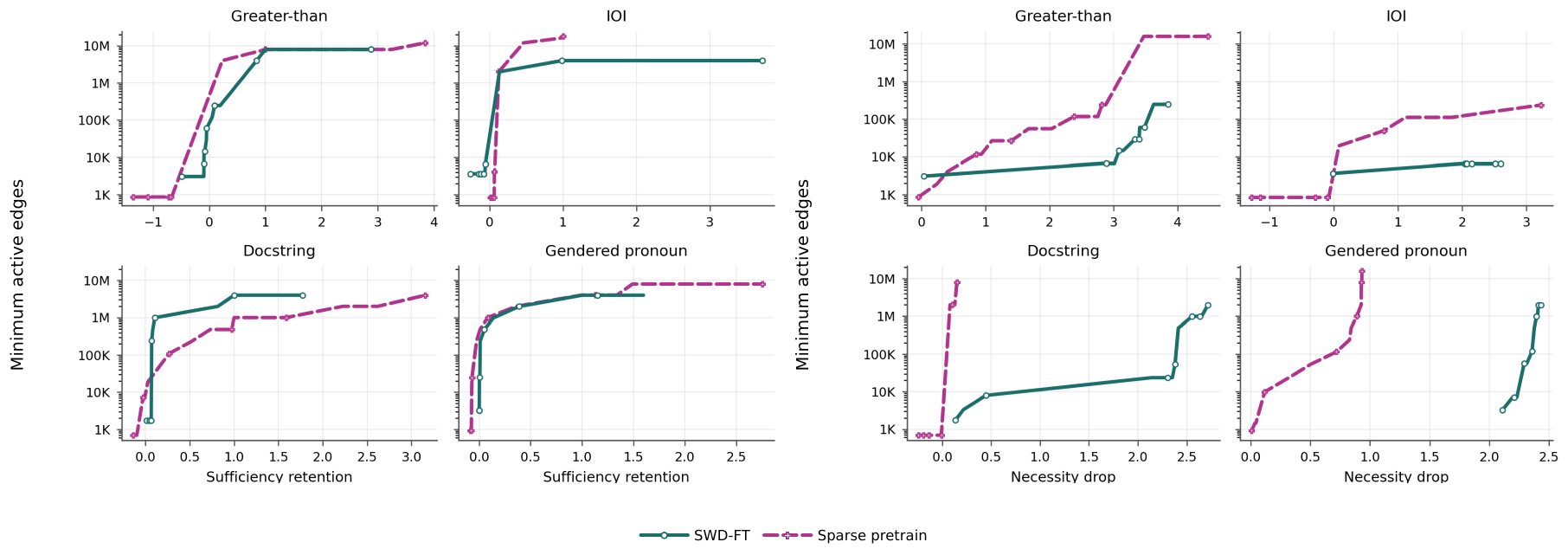}
    \caption{GPT-2 Small full-model replacement circuit results. SWD-FT remains effective across all four tasks. Sparse pretraining is comparable on greater-than and gendered-pronoun, but has near-zero or negative necessity drop on docstring and IOI.}
    \label{fig:gpt2-fullmodel-circuit}
\end{figure}

For task circuits, we prune SWD bottleneck units and the sparse-pretrained model's latent channels under their corresponding protocols (Section \ref{sec:circuit-extraction-protocol}), then compare the minimum active edges needed to reach each held-out sufficiency or necessity target, as shown in Figure~\ref{fig:gpt2-fullmodel-circuit}. At the approximately matched budget, the methods are broadly comparable on greater-than and gendered-pronoun. On docstring and IOI, the sparse-pretraining circuit results have near-zero or negative necessity drop, whereas SWD-FT remains effective on all four tasks.
\takeaway{Fixed-support fine-tuning restores replacement quality and matches sparse pretraining at an approximately equal active-weight budget and circuit cost-quality tradeoff with far fewer training tokens.}

\subsection{Ablation: Exact Dense Reparameterizations}
\label{sec:exact-dense-control-ablation}
\begin{figure}[ht]
    \centering
    \includegraphics[width=\linewidth]{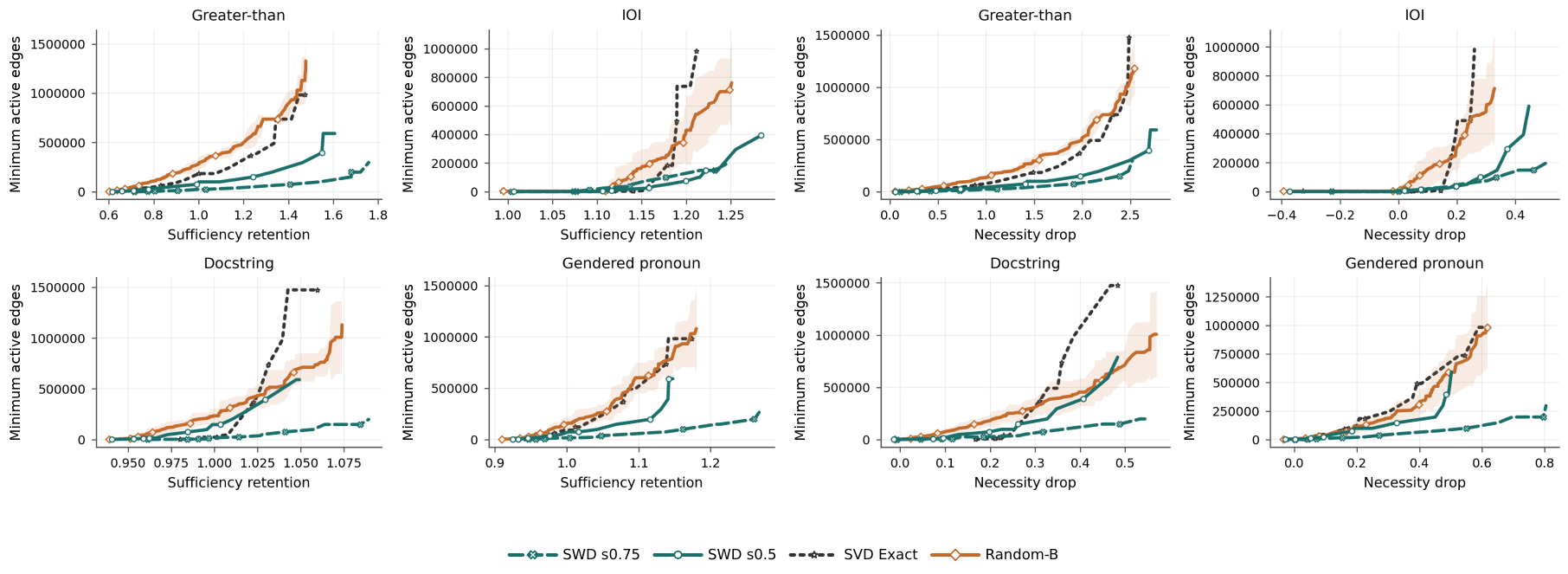}
    \caption{Exact dense reparameterization controls for the GPT-2 output-projection surface. The comparison with SWD tests whether sparse read/write structure reduces the number of active edges beyond factorization alone. See Appendix~\ref{app:exact-dense-controls} for the control settings.}
    \label{fig:gpt2-cproj-exact-dense-controls}
\end{figure}

The fact that SWD requires fewer circuit edges could arise either from its sparse read/write matrices or simply from factorizing the original matrix into intermediate units. To distinguish these explanations, we compare SWD with two exact factorizations of the same GPT-2 Small output-projection matrix. The first is a full-rank SVD, using the same exact rank-one decomposition as NaNA \citep{xue2026svd}; the second is Random-B, which uses a seeded random orthogonal matrix for $\mB$ and computes $\mA$ from $\mW$ and $\mB$. Both factorizations reproduce the target matrix exactly, but their factor matrices are dense. We apply the same task-level attribution and mean-ablation procedure to all three methods.
Figure~\ref{fig:gpt2-cproj-exact-dense-controls} shows that the exact SVD and Random-B factorizations require more active edges than SWD to retain the same task behavior under sufficiency evaluation or induce the same performance drop under necessity evaluation. Since this pattern holds for both the SVD basis and random orthogonal bases, the comparison isolates the contribution of sparse read/write structure from factorization alone. Appendix~\ref{app:exact-dense-controls} provides the construction, fidelity checks, and companion selected-unit results.

\takeaway{The exact SVD and Random-B factorizations require more active edges than SWD to preserve the same task behavior or produce the same ablation effect, isolating the contribution of sparse read/write structure.}

\subsection{Zero-Data Factorization}
\label{sec:zero_data}
The previous section shows that sparse structure helps SWD achieve low-cost circuit results. We next ask whether constructing such bottleneck units requires calibration activations. Calibration activations may be unavailable and can tie the factorization to a particular data distribution, so we evaluate the zero-data SWD variant defined in Section~\ref{sec:swd-factorization}, which depends only on checkpoint weights. Because zero-data SWD requires only the current weights, it can in principle be applied at every training step without collecting calibration activations, enabling step-by-step mechanistic analysis of how circuit structure emerges and evolves during training. We leave such training-trajectory analysis to future work. On the GPT-2 output-projection surface, we vary $s$ from $0.125$ to $0.875$ and measure two complementary notions of fidelity: relative Frobenius error $\|\mW-\mA\mB\|_F/\|\mW\|_F$ in parameter space and held-out CE delta in model behavior.

\begin{figure}[t]
    \centering
    \includegraphics[width=\linewidth]{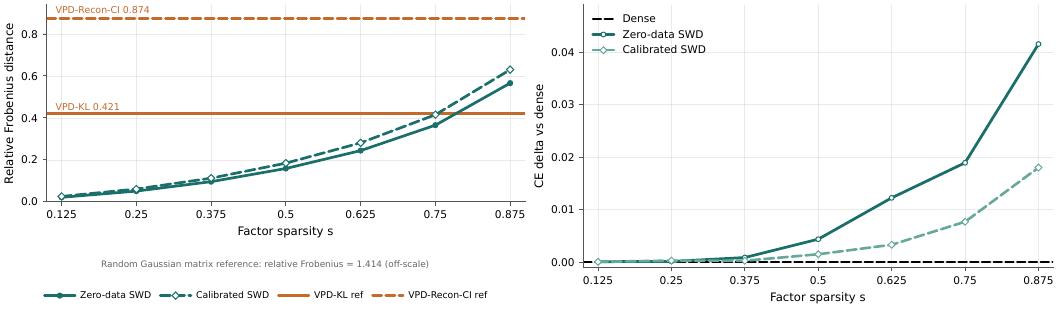}
    \caption{Parameter-space and behavioral fidelity of zero-data and activation-calibrated SWD on the GPT-2 output-projection surface. Left: relative Frobenius reconstruction error. Right: held-out CE increase. Zero-data SWD better preserves the weight matrix, whereas activation calibration better preserves model behavior under aggressive sparsification. Appendix~\ref{app:zero-data-setting} gives the factorization and evaluation settings.}
    \label{fig:zero-data-pair}
\end{figure}

As shown in Figure~\ref{fig:zero-data-pair}, zero-data SWD remains closer to $\mW$ throughout the all sparsity levels, reaching relative error $0.366$ versus $0.415$ for activation-calibrated SWD at $s=0.75$. Conversely, activation-calibrated SWD achieves lower CE under aggressive sparsification because it allocates the limited edge budget toward the distribution of activation directions. Thus zero-data SWD better preserves the raw weights, whereas calibration better preserves behavior on typical model activations. To further analyze the circuit properties of the zero-data factors, we apply the same circuit extraction protocol in Appendix~\ref{app:zero-data-task-frontiers} and show that zero-data bottlenecks remain useful for task circuits.

\takeaway{Calibration data improves replacement fidelity, but useful circuit units can still be recovered directly from checkpoint weights without activation data.}

\subsection{Qualitative Results}
\label{sec:qualitative-results}

We conclude the experimental analysis with three qualitative views of SWD bottleneck units. We first examine whether units selected by task attribution have recognizable semantic hypotheses, then test whether an individual unit supports a targeted directional edit, and finally use the factors to diagnose attention computation and interactions between units. These analyses complement the aggregate circuit results above by illustrating what the extracted units represent and how they can be manipulated.

\subsubsection{Semantic Audit of a Task Circuit}
\begin{figure}[t]
    \centering
    \includegraphics[width=0.99\linewidth]{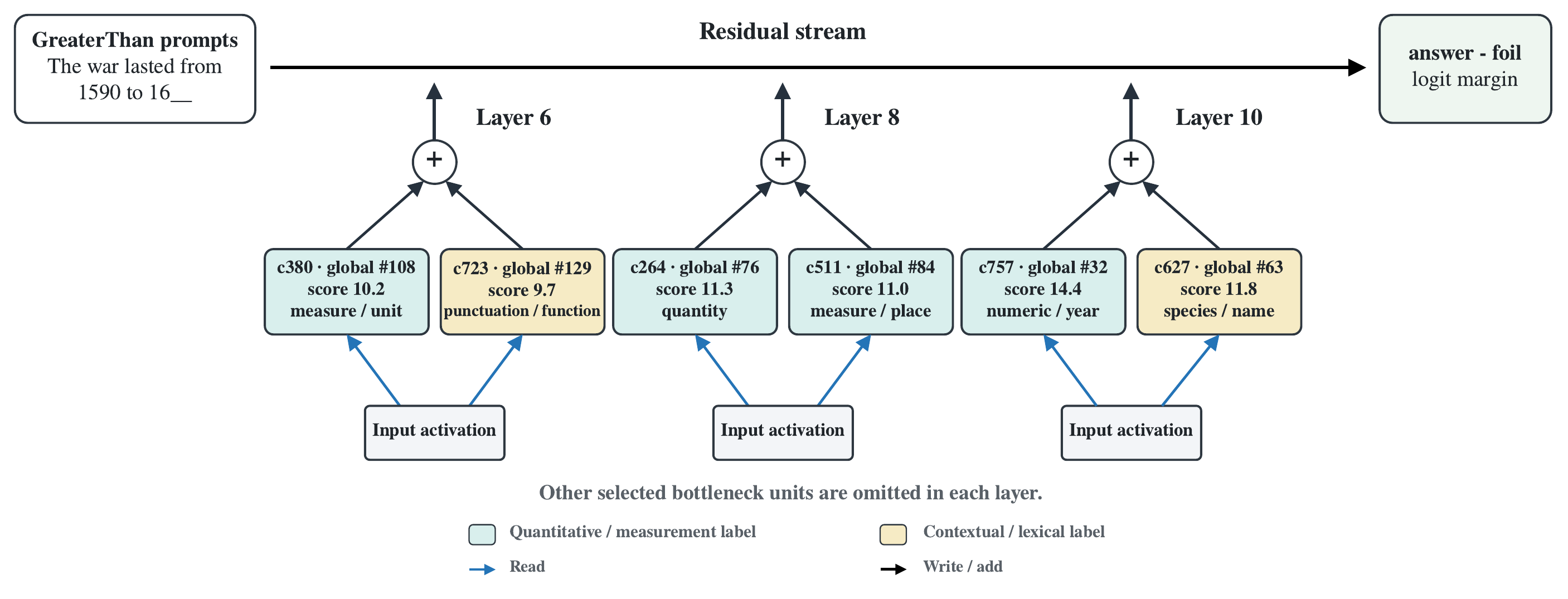}
    \caption{Semantic audit of high-attribution paths in a GreaterThan SWD circuit. Arrows show the directed computation within each factorized \texttt{mlp.c\_proj}: the MLP input activation $\vh_\ell$ is read by bottleneck unit $i$ to form $z_i=\vh_\ell\mA_{:,i}$, and the unit writes the vector contribution $z_i\mB_{i,:}$ toward the layer output. Each drawn arrow summarizes the corresponding sparse read or write connections, rather than a measured pairwise interaction between units. The $+$ node denotes vector addition of the bottleneck-unit write contributions into the residual stream. We draw the two highest-ranked units in each displayed layer and omit the remaining selected units. Nodes are selected by task-related first-order attribution; their semantic labels come from recurring patterns in task-independent, high-activation contexts and are not used for selection. Four of the six displayed units have number-, quantity-, or measurement-related hypotheses.}
    
    \label{fig:greater-than-case-study}
\end{figure}

We first audit GreaterThan on the GPT-2 Small all-layer \texttt{mlp.c\_proj} surface. We rank all $9{,}208$ bottleneck units by the train-split positive first-order attribution used for pruning, and Figure~\ref{fig:greater-than-case-study} shows the two highest-ranked units in layers 6, 8, and 10; all six lie in the global top-$512$ prefix. For each unit, we inspect its 20 highest-magnitude activation contexts from a task-independent $2{,}048$-token WikiText-2 dashboard and assign a semantic hypothesis from recurring patterns; Appendix~\ref{app:qualitative-text-activations} gives the full protocol and per-layer ranks. The resulting task-independent labels align with the task family: four of the six highlighted paths are associated with numbers, quantities, or measurements, while the remaining two capture syntax or named entities. This correspondence gives the extracted circuit a semantic interpretation that complements the held-out sufficiency and necessity results above.

\subsubsection{Targeted Bottleneck-Unit Editing}
Having examined what attribution-selected bottleneck units respond to, we next test whether an individual SWD bottleneck unit can support a directional intervention. To ensure a common base model across editing methods, we apply all edits to the original dense GPT-2 Small layer-8 \texttt{mlp.c\_proj} weight. On the prompt \texttt{The opposite of up is}, we use the read vector of bottleneck unit c205 to construct a rank-one update along the answer--foil unembedding direction defined by \texttt{ down} and \texttt{ left}. At the strongest positive setting shown in the sweep, the answer--foil margin increases by $0.216$, while the mean final-token KL on seven unrelated factual prompts is $4.02\times10^{-5}$.

\begin{figure}[t]
    \centering
    \includegraphics[width=0.68\linewidth]{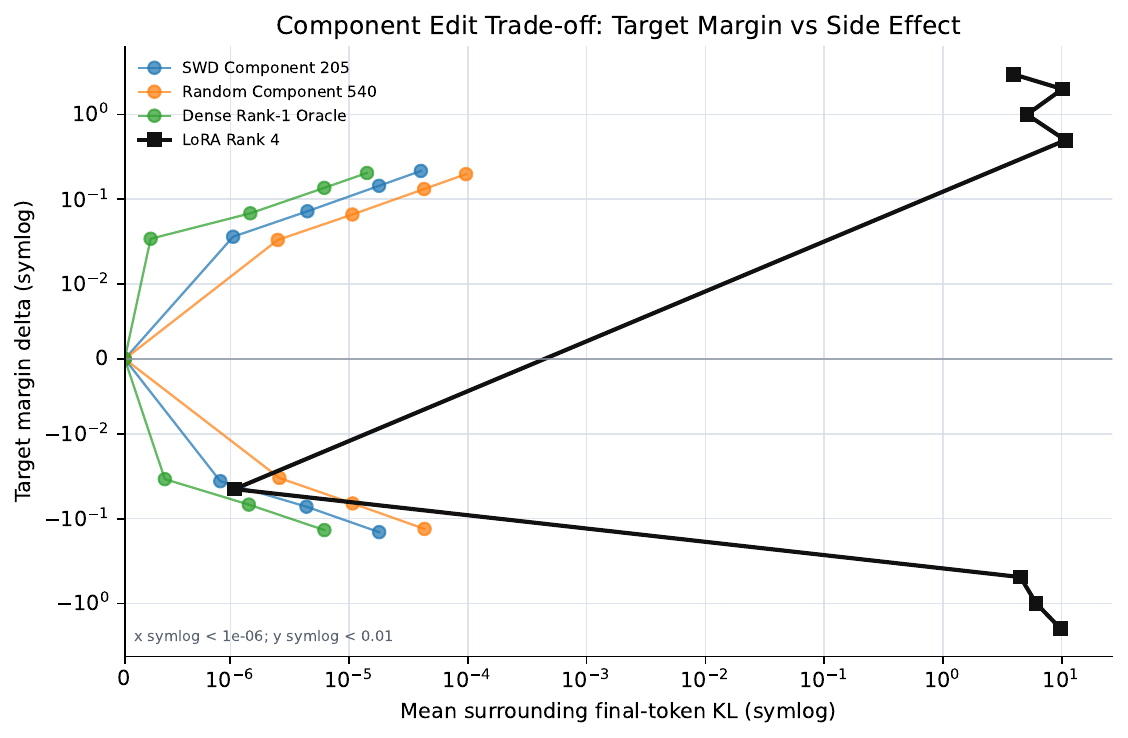}
    \caption{Target-margin change versus measured side effect for single-target edits applied to the same original dense weight matrix. The horizontal axis is mean final-token KL on seven unrelated factual prompts; the vertical axis is the change in the \texttt{ down} versus \texttt{ left} margin. Positive and negative edit settings form the upper and lower branches. At comparable margin change, farther left indicates lower measured side effect. The SWD curve uses the read direction of bottleneck unit c205; the dense rank-one oracle uses the target activation directly.}
    \label{fig:component-edit-pareto}
\end{figure}

Figure~\ref{fig:component-edit-pareto} shows the efficacy--locality tradeoff over the full edit sweep. The selected bottleneck unit c205 remains farther left than the random-unit control and rank-4 LoRA at comparable positive target-margin changes, indicating lower measured influence on the seven non-target prompts. The target-conditioned dense rank-one oracle achieves the strongest tradeoff, as expected from using the target activation directly. Thus, this example demonstrates a more precise edit through an extracted SWD bottleneck direction than through the practical controls. Appendix~\ref{app:component-editing} gives the construction and control details.

\FloatBarrier
\subsubsection{Mechanistic Diagnostics}
\begin{figure}[t]
    \centering
    \includegraphics[width=0.96\linewidth]{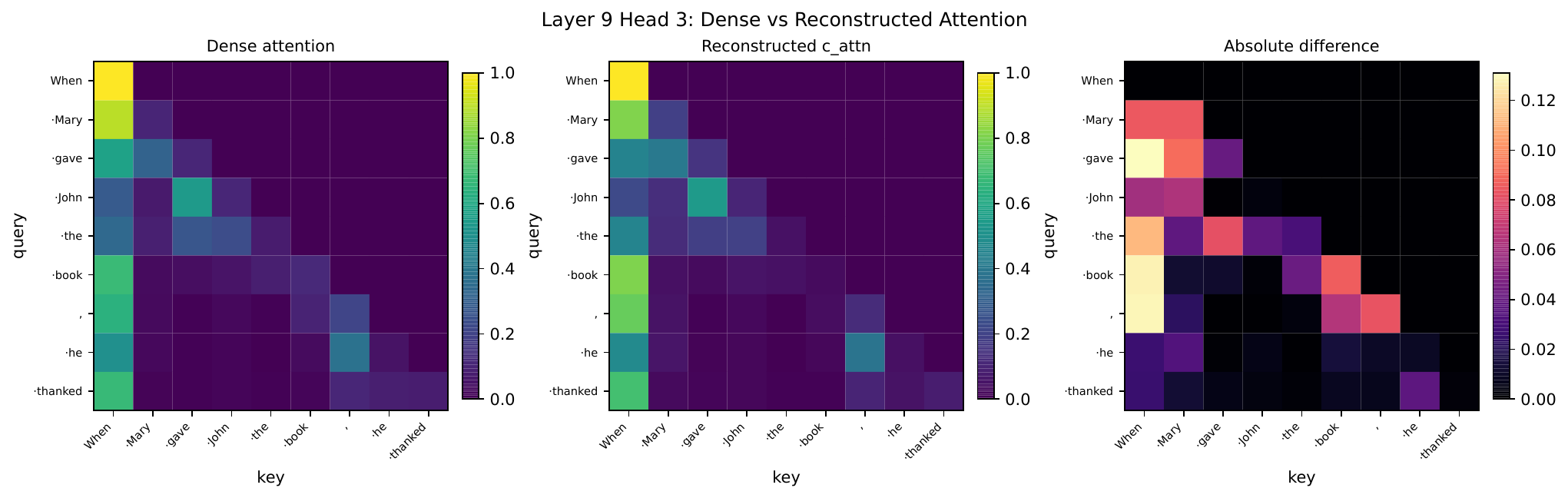}

    \vspace{3pt}
    \includegraphics[width=0.96\linewidth]{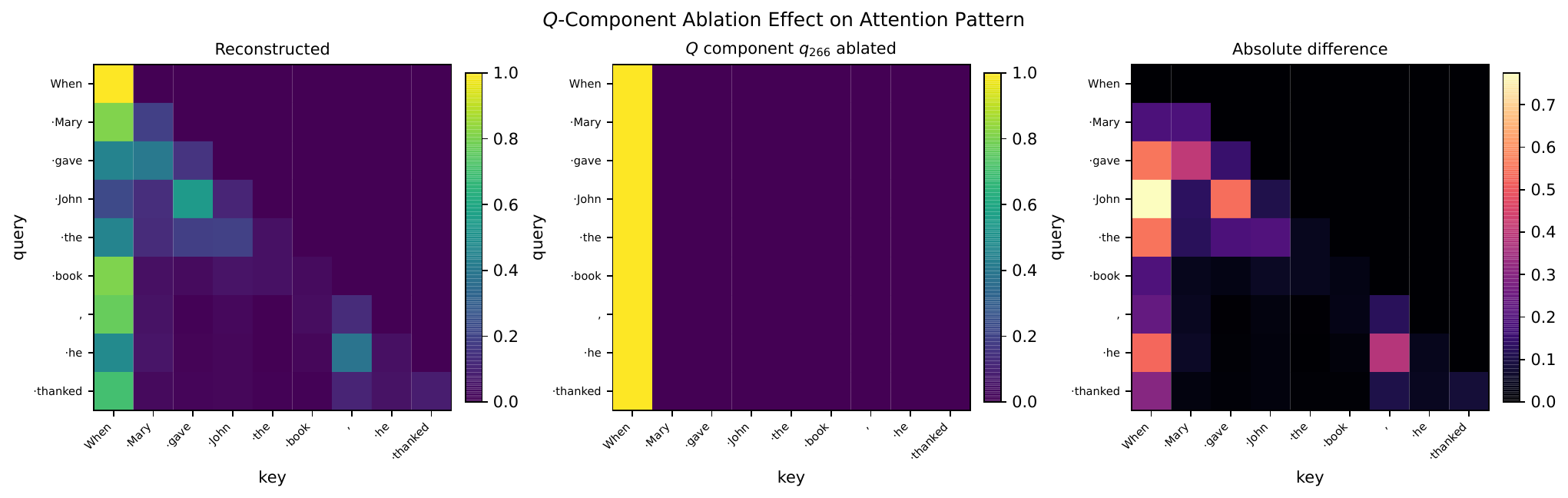}
    \caption{Prompt-level validation and intervention for layer 9, head 3. Top: dense attention, SWD-reconstructed attention, and their absolute difference; mean dense-to-reconstructed attention KL is $0.0405$.   Bottom: SWD-reconstructed attention before and after ablating query bottleneck unit q266, followed by their elementwise absolute difference. Relative to the intact reconstruction, ablating q266 produces a mean attention KL of $2.496$ and a maximum absolute change of $0.775$ in any   attention probability.}
    \label{fig:attention-replay-ablation}
\end{figure}

Beyond task-specific interpretation and editing, we use SWD factors for two complementary mechanistic diagnostics. First, adapting weight- and feature-level analyses of QK circuits \citep{elhage2021mathematical,franco2024sparseattentiondecompositionapplied,kamath2025tracing}, we screen sparse query-key bottleneck unit pairs by their static write direction interaction, replay the strongest candidates on prompts, and ablate the selected query unit. Figure~\ref{fig:attention-replay-ablation} visualizes the selected layer-9, head-3 example. The SWD reconstruction retains the dense attention structure with mean KL $0.0405$ (top), whereas removing q266 redirects attention sharply toward the first token, raising the mean KL to $2.496$ and producing a maximum probability change of $0.775$ (bottom). The contrast connects accurate reconstruction to a prompt-local causal effect of an individual bottleneck unit; Appendix~\ref{app:attention-components} gives the static screen and token-level analysis that led to this example.

\FloatBarrier
Second, because additive bottleneck units need not be functionally independent, we measure pairwise overlap after GELU and pairwise non-additivity before GELU. Post-GELU \texttt{c\_proj} contributions are nearly uncorrelated apart from a small structured tail, while pre-GELU \texttt{c\_fc} bottleneck units show broader interaction through the nonlinearity (Appendix~\ref{app:component-interactions}). Together, the attention intervention and interaction analysis extend the qualitative results from individual units to the computations and relationships in which they participate.

\takeaway{Across semantic auditing, targeted editing, and mechanistic diagnostics, SWD bottleneck units provide interpretable and manipulable units for analyzing model computation, while the observed interactions caution against treating every unit as independent or monosemantic.}

\section{Related Work}

\paragraph{Task-level circuit discovery and evaluation.}
Mechanistic interpretability (MI) seeks small computational subgraphs whose interventions explain a model behavior \citep{olah2020zoom}. Manual transformer-circuit analyses have isolated mechanisms such as induction heads, indirect-object identification, and greater-than comparison \citep{elhage2021mathematical,olsson2022context,wang2022interpretability,hanna2023how}. Causal tracing and activation patching localize behavior by exchanging or ablating internal states, while ACDC, attribution patching, and edge pruning make the search over components or edges more scalable \citep{meng2022locating,conmy2023automated,syed2023attribution,bhaskar2024edge}. Recent evaluations emphasize that circuit conclusions depend on the intervention, metric, and granularity, and therefore favor held-out causal tests and concise recovered pathways over overlap with a single presumed circuit \citep{hanna2024faithfulness,zhang2024activationpatching,miller2024transformercircuitfaithfulnessmetrics,mueller2025mib}. Whereas these methods primarily decide which nodes or edges to retain in an existing computational basis, SWD introduces an alternative basis of addressable bottleneck units inside otherwise dense linear maps. It is thus complementary to circuit-search algorithms: attribution and mean ablation can be applied to the induced units, with held-out sufficiency and necessity measuring circuit quality and selected units and active edges measuring circuit size.

\paragraph{Sparse activation features and learned replacement modules.}
Superposition makes native neurons polysemantic and motivates learning overcomplete sparse feature dictionaries \citep{elhage2022toy}. Sparse autoencoders (SAEs) decompose residual-stream or sublayer activations into sparsely active features, and large releases such as Gemma Scope make such dictionaries available across layers and model scales \citep{bricken2023monosemanticity,cunningham2023sparse,lieberum2024gemmascope}. Sparse feature circuits connect SAE features into causal graphs that support intervention and editing \citep{marks2025sparsefeaturecircuits}. Transcoders instead learn a sparse-feature module that approximates an MLP's input--output computation, avoiding some difficulties in tracing an SAE feature through the original nonlinear MLP; attribution graphs extend this replacement-based view to larger computation graphs \citep{dunefsky2024transcoders,ameisen2025circuit,kamath2025tracing}. Like these approaches, SWD exposes scalar units that can be scored, selected, and intervened on. SAEs and Transcoders, however, fit a new activation dictionary or replacement network on an activation corpus, whereas SWD factorizes a checkpoint matrix and sparsifies each unit's parameter-side read and write vectors. This construction applies uniformly to MLP and attention projections, and defines each unit as a sparse rank-one path without assuming that it is a monosemantic activation feature. We test the practical effect of this distinction through the matched-fidelity Transcoder comparison.

\paragraph{Parameter-space interpretability.}
Parameter-decomposition methods seek simpler additive components directly in model weights. Attribution-based Parameter Decomposition (APD) optimizes for parameter faithfulness, sparse per-input component use, and component simplicity; Stochastic Parameter Decomposition (SPD) develops a more scalable stochastic formulation \citep{braun2025apd,bushnaq2025spd}. VPD extends this family to language-model parameters using an input-dependent selector, and sparse attention decomposition uses SVD-derived parameter components to trace query--key computations \citep{bushnaq2026interpreting,franco2024sparseattentiondecompositionapplied}. NaNA treats each rank-one component in the exact SVD of an MLP weight matrix as a detector--effector unit (DEU). It scores each DEU by how strongly it is activated by the input and how much it supports a specified target token, then retains or removes the highest-scoring units to test their effect on the prediction \citep{xue2026svd}. SWD also constructs fixed rank-one components from model weights, but each component consists of one sparse column of $\mA$ and one sparse row of $\mB$. SWD constructs these components independently of the downstream circuit task and subsequently ranks them using a common task-margin attribution protocol; VPD instead learns input-dependent component use as part of the decomposition.

\paragraph{Sparse models and post-training pruning.}
Weight pruning shows that substantial parameter sparsity can be recovered from dense language models with little or no retraining. SparseGPT uses approximate second-order information for one-shot pruning, while Wanda scores weights using both magnitude and observed input activations \citep{frantar2023sparsegpt,sun2023wanda}. A complementary approach builds weight and activation sparsity into pretraining; such sparse-pretrained transformers can preserve more tractable circuits than comparable dense models, albeit at the cost of training a new model \citep{gao2025weightsparse}. These results motivate the hypothesis that useful sparse structure can be recovered from a dense checkpoint, but do not directly provide the intervention surface used by SWD. Pruning removes scalar edges without creating intermediate units that can be independently scored and ablated, while sparse pretraining changes the model from the start. SWD instead introduces bottleneck units with sparse incoming and outgoing edges post hoc. The sparse-pretraining comparison tests the shared sparse-connectivity hypothesis, while the exact dense-factor controls distinguish the effect of sparse edges from that of factorization alone.

\paragraph{Matrix factorization and compression.}
Classical matrix methods impose low-rank, non-negativity, or sparsity to obtain compact representations \citep{lee1999learning,mairal2010online,aspremont2007direct}. For neural-network compression, quantized sparse weight decomposition combines structural constraints, while activation-aware and truncation-aware SVD methods use calibration statistics to preserve the action of a weight matrix on likely inputs \citep{kuzmin2022quantized,yuan2023asvd,wang2025svdllm}. NaNA uses rank-one components from an exact SVD as intervention units \citep{xue2026svd}. In an exact SVD, however, the left and right singular vectors are generally dense, so each component connects to most input and output dimensions. Double Sparse Factorization (DSF) instead approximates a dense weight matrix as the product of two sparse matrices \citep{boza2025double}. SWD adopts DSF and uses calibration inputs when fitting its two sparse matrices so that the replacement preserves the outputs of the original dense matrix. The exact-SVD controls in Section~\ref{sec:exact-dense-control-ablation} apply the same attribution and ablation protocol to dense SVD components and sparse SWD components, testing whether sparse read/write connections reduce the number of selected units and active edges required to reach the same sufficiency and necessity levels.

\vspace{-0.08in}
\section{Limitations}
\label{sec:limitations}
\vspace{-0.08in}

\paragraph{Efficient extraction is not complete understanding.} SWD reaches matched replacement fidelity while using less than $1\%$ of the data that trained baselines use to train their replacements, and it exposes circuits with fewer active edges. This local-to-global gap is shared by other MI approaches: learned feature dictionaries and task-circuit methods provide tractable local objects without, by themselves, yielding a complete account of the model \citep{cunningham2023sparse,conmy2023automated,syed2023attribution,ameisen2025circuit}. Thus, our efficiency gains do not resolve the broader concern that a large nonlinear model may not admit a compact, human-understandable mechanistic account \citep{hendrycks2025misguided}. Our sufficiency and necessity results establish that selected bottleneck units causally affect held-out task margins, but they do not explain the model globally or guarantee coverage of rare, safety-relevant cases.

\paragraph{SWD circuits are local and non-unique.} Identifiability of a two-factor sparse matrix decomposition requires additional structural conditions and is, even then, only defined up to unavoidable scaling and permutation symmetries \citep{zheng2021identifiability}. SWD does not establish such conditions for its approximate learned decompositions, so the resulting bottleneck units can depend on the objective, calibration distribution, initialization, and sparsity budget. Moreover, SWD generally intervenes on an approximate replacement $\mA\mB$: matched cross-entropy and reconstruction control average drift but cannot guarantee agreement with the original dense model on every input. First-order ranking can miss behaviorally important edges, and mean-ablation conclusions can vary with the intervention and evaluation protocol \citep{hanna2024faithfulness,zhang2024activationpatching,miller2024transformercircuitfaithfulnessmetrics}; redundant or nonlinear mechanisms are also suggested by Appendix~\ref{app:component-interactions}. Recent controlled evidence further shows that exact circuit claims can change with the reported graph, pruning threshold, query/key representation, and comparison granularity, even when coarser summaries remain stable \citep{sheng2026circuitclaimsdependextracted}. Our comparisons hold the task data, ranking rule, ablation policy, top-$k$ evaluation, and cost accounting fixed across methods; this supports controlled comparisons within our chosen protocol but does not establish invariance to alternative extraction or reporting choices. The extracted circuits should therefore be interpreted as task- and protocol-specific causal explanations, not unique or complete mechanistic ground truth.

\paragraph{Scaling verification remains open.} Our Qwen3.5-27B experiment covers one matrix, full-model replacement is evaluated only on GPT-2 Small, and the semantic audit examines six model-labeled bottleneck units. Thus, reducing factorization data and circuit size does not remove the need to verify hypotheses across decompositions, prompt distributions, and adversarial edge cases. Following the top-down alternative emphasized by \citet{hendrycks2025misguided}, SWD is best used alongside behavioral and representation-level analyses rather than as a certificate of understanding or safety.

\takeaway{SWD makes local circuit hypotheses substantially cheaper to extract and test, but it does not eliminate MI's non-uniqueness, distributional, verification, or scaling limitations.}

\section{Conclusion}
\vspace{-0.08in}
Sparse Weight Decomposition turns dense pretrained projections into bottleneck units with sparse read/write connectivity that can be used directly for circuit extraction. Across GPT-2, Qwen2.5 and Qwen3.5-27B, SWD improves matched-fidelity circuit cost-quality tradeoff over learned sparse replacements. Fixed-support fine-tuning extends the method to full-model replacement, and a zero-data variant recovers useful bottleneck units from checkpoint weights alone. Exact SVD and Random-B factorizations isolate the contribution of sparse read/write structure from factorization alone. The resulting parameter-side surface supports compact circuit selection and causal testing in dense models.

\bibliographystyle{iclr2026_conference}
\bibliography{iclr2026_conference}
\clearpage
\appendix
\section{Notation}
\label{app:notation}

Table~\ref{tab:notation} consolidates the recurring notation used across the main text and appendices. We typeset scalars and indices in ordinary italic, vectors in bold lowercase, and matrices in bold uppercase; sets, datasets, models, and scalar-valued functions remain nonbold. Layer, head, and unit identifiers are zero-based, whereas displayed attribution ranks are one-based. For method-specific terms, we call an SWD intermediate coordinate a \emph{bottleneck unit}, a Transcoder hidden element a \emph{feature}, a VPD element a \emph{parameter component}, and a sparse-pretrained hidden element a \emph{latent channel}. We use \emph{unit} as the method-agnostic umbrella term in cross-method comparisons and selected-unit figures.

\begin{table}[H]
    \centering
    \small
    \renewcommand{\arraystretch}{1.5}
    \begin{tabular}{p{0.20\linewidth}p{0.70\linewidth}}
        \hline
        Symbol & Meaning \\
        \hline
        $\mW,\widehat{\mW}$ & Dense target weight matrix and its SWD replacement, $\widehat{\mW}=\mA\mB$. \\
        $\mA,\mB$ & Sparse read and write factors, $\mA\in\mathbb{R}^{d_{\mathrm{in}}\times m}$ and $\mB\in\mathbb{R}^{m\times d_{\mathrm{out}}}$. \\
        $d_{\mathrm{in}},d_{\mathrm{out}}$ & Input and output dimensions of the target matrix. \\
        $m,C$ & Numbers of SWD bottleneck units and VPD component slots, respectively. \\
        $i,k$ & $i$ indexes a candidate unit; $k$ is the number of highest-ranked units retained in a top-$k$ circuit. \\
        $\vx,\vh_{\ell,t}(\vx)$ & Input text example $\vx$ and the input activation to target module $\ell$ at token position $t$. \\
        $\mH$ & Calibration-activation matrix whose rows are input-activation vectors $\vh$. \\
        $\mG,\mI$ & Calibration Gram matrix $\mG=\mathbb{E}[\vh^\top\vh]$ and the identity matrix used by the zero-data objective. \\
        $K,s$ & Total factor-entry budget $K=\|\mA\|_0+\|\mB\|_0$ and factor sparsity $s=1-K/\|\mW\|_0$. \\
        $\mathcal{D}_{\mathrm{cal}}$ & Distribution of calibration activations used to fit SWD. \\
        $\mathcal{D}_{\tau,\mathrm{train}},\mathcal{D}_{\tau,\mathrm{test}}$ & Training and held-out evaluation splits for circuit task $\tau$. \\
        $z_i,\mu_i,a_i$ & Activation of unit $i$, its train-split mean used for ablation, and its task-attribution score. For SWD, $z_i=\vh\mA_{:,i}$. \\
        $S_k,S_{\mathrm{eff}}$ & $S_k$ is the set of the $k$ highest-ranked units; $S_{\mathrm{eff}}$ contains the selected units with at least one nonzero read edge and one nonzero write edge. \\
        $\mathbf{r}_i,\mathbf{w}_i$ & Method-specific read and write weight vectors of unit $i$; for SWD these are $\mA_{:,i}$ and $\mB_{i,:}$. \\
        $C_{\mathrm{unit}}(S),C_{\mathrm{edge}}(S)$ & Selected-unit cost $|S|$ and active-edge cost $\sum_{i\in S_{\mathrm{eff}}}(\|\mathbf{r}_i\|_0+\|\mathbf{w}_i\|_0)$. \\
        $M_{\mathrm{unpruned}}$ & Replacement model before circuit pruning. \\
        $M_{\mathrm{keep}}(S),M_{\mathrm{abl}}(S)$ & Models that respectively retain only $S$ or mean-ablate $S$. \\
        $g_{\tau}(M,\vx)$ & Per-example answer-versus-distractor logit margin for task $\tau$. \\
        $Q_{\tau}(M;\mathcal{D})$ & Average task margin of model $M$ over split $\mathcal{D}$. \\
        \hline
    \end{tabular}
    \caption{Notation used throughout the paper.}
    \label{tab:notation}
\end{table}

The support of a factor is the set of its nonzero entries, and its factor mask is the corresponding binary indicator. A unit is valid when both its read and write vectors contain at least one nonzero entry; $S_{\mathrm{eff}}$ is the valid subset of a selected set.

\section{Experimental Details}
\label{app:experimental-details}

This section consolidates the settings shared across experiments. We first specify the replacement surfaces, then give the SWD optimization, circuit-selection protocol, task data, and baseline configurations. Throughout our experiments, replacement fidelity is measured on data excluded from replacement fitting. For circuit comparisons, unit scores and mean-ablation values are estimated on \texttt{circuit\_train}, and the resulting nested top-$k$ prefixes are evaluated on \texttt{circuit\_test}.

\subsection{Models and Replacement Targets}

Table~\ref{tab:replacement-surfaces} summarizes the computational surfaces used in the paper. Within each single-matrix comparison, all methods replace the same weight matrix from the pretrained model. The MLP-replacement experiment is reported separately because a standard Transcoder replaces the entire MLP, whereas the full-model experiment replaces all attention and MLP weight matrices across the transformer blocks and therefore tests accumulated approximation error.

\begin{table}[H]
    \centering
    \small
    \setlength{\tabcolsep}{4pt}
    \begin{tabular}{p{0.3\linewidth}p{0.26\linewidth}p{0.38\linewidth}}
        \hline
        Setting & Model & Replacement surface \\
        \hline
        GPT-2 single matrix & GPT-2 Small & Layer 8 \texttt{mlp.c\_proj} \\
        Qwen2.5 scaling & Qwen2.5 0.5B, 1.5B, 3B & Middle-layer \texttt{mlp.down\_proj}: layers 12, 14, and 18, respectively \\
        Qwen3.5 single matrix & Qwen3.5-27B & Layer 31 \texttt{mlp.down\_proj} \\
        GPT-2 MLP replacement & GPT-2 Small & Layer 8 standard MLP, comprising \texttt{mlp.c\_fc} and \texttt{mlp.c\_proj} \\
        GPT-2 full-model replacement & GPT-2 Small & \texttt{attn.c\_attn}, \texttt{attn.c\_proj}, \texttt{mlp.c\_fc}, and \texttt{mlp.c\_proj} in all 12 blocks \\
        Zero-data & GPT-2 Small & Layer 8 \texttt{mlp.c\_proj}, matching the GPT-2 single-matrix experiment \\
        Qualitative audit & GPT-2 Small & \texttt{mlp.c\_proj} in all 12 blocks; selected bottleneck units are visualized from layers 6, 8, and 10 \\
        \hline
    \end{tabular}
    \caption{Models and replacement surfaces. Each comparison uses a common surface across the methods shown in that comparison.}
    \label{tab:replacement-surfaces}
\end{table}

\subsection{SWD Factorization and Fixed-Support Fine-Tuning}
\label{app:swd-optimization}

We implement SWD with the activation-aware Double Sparse Factorization solver of \citet{boza2025double}. Standard decompositions use inner width $m=\min(d_{\mathrm{in}},d_{\mathrm{out}})$. The following configuration is fixed across a sparsity sweep; only the total nonzero budget $K$ changes.

\begin{itemize}[leftmargin=1.5em,itemsep=0.25em,topsep=0.35em]
    \item \textbf{Budget and initialization.} The square factor receives density $0.16$ for square targets and $0.25$ for rectangular targets, with the remaining entries in $K$ assigned to the other factor. We rescale the target using the diagonal of the calibration Gram matrix, initialize the square factor to the identity and the other factor to a magnitude-pruned rescaled target, and initialize the dual variables to zero.
    \item \textbf{Alternating optimization.} Each outer iteration updates both factors in turn. We use 40 outer iterations for the GPT-2 single-matrix and full-model experiments and 8 for the Qwen single-matrix experiments.
    \item \textbf{ADMM block updates.} Each block update uses 5 inner iterations, penalty $\rho=1$, and ridge coefficient $10^{-2}$. Hard thresholding recomputes the prescribed support during the first 2 inner iterations and then holds it fixed. At outer iteration $t$ of $T$, the first solve uses $\rho_{\mathrm{init}}=\min\{1,t/(T-3)\}^{3}$; subsequent inner solves use $\rho=1$.
    \item \textbf{Finalization and precision.} After alternating optimization, 20 fixed-support ADMM iterations refit the selected entries of one factor while holding the other factor and both factor supports fixed. Factorization uses float32 with TF32 matrix multiplication disabled. This local finalization is distinct from the model-level fixed-support fine-tuning used in the full-model experiment below.
    \item \textbf{Data used to fit SWD replacements.} Each point on an SWD replacement-quality curve corresponds to an independently fitted SWD replacement using the stated amount of calibration data; the points are not successive checkpoints from one training run. Table~\ref{tab:swd-data-circuit-checkpoints} separates the data ranges evaluated in these curves from the fixed SWD replacements subsequently used for circuit extraction.
\end{itemize}

\begin{table}[H]
    \centering
    \small
    \setlength{\tabcolsep}{4pt}
    \begin{tabular}{p{0.20\linewidth}p{0.23\linewidth}p{0.48\linewidth}}
        \hline
        Experiment & CE-curve data (tokens) & Fixed replacement used for circuit extraction \\
        \hline
        \multirow{2}{=}{GPT-2 single matrix}
        & \multirow{2}{=}{1,024--4.19M}
        & $s{=}0.5$: 16,384 tokens; CE delta $0.000889$ \\
        & & $s{=}0.75$: 16,384 tokens; CE delta $0.008292$ \\
        \cmidrule(lr){1-3}
        \multirow{2}{=}{Qwen2.5-0.5B}
        & \multirow{2}{=}{1,024--33.55M}
        & $s{=}0.5$: 1,024 tokens; CE delta $0.001222$ \\
        & & $s{=}0.75$: 1.05M tokens; CE delta $0.005010$ \\
        \cmidrule(lr){1-3}
        \multirow{2}{=}{Qwen2.5-1.5B}
        & \multirow{2}{=}{1,024--33.55M}
        & $s{=}0.5$: 1,024 tokens; CE delta $0.000733$ \\
        & & $s{=}0.75$: 1.05M tokens; CE delta $0.001222$ \\
        \cmidrule(lr){1-3}
        \multirow{2}{=}{Qwen2.5-3B}
        & \multirow{2}{=}{1,024--33.55M}
        & $s{=}0.5$: 2,048 tokens; CE delta $0.000733$ \\
        & & $s{=}0.75$: 1.05M tokens; CE delta $0.000611$ \\
        \cmidrule(lr){1-3}
        \multirow{2}{=}{Qwen3.5-27B}
        & \multirow{2}{=}{1,024--16.78M}
        & $s{=}0.5$: 2,048 tokens; CE delta $-0.000427$ \\
        & & $s{=}0.75$: 1.05M tokens; CE delta $-0.000448$ \\
        \cmidrule(lr){1-3}
        \multirow{2}{=}{GPT-2 MLP replacement}
        & \multirow{2}{=}{1,024--33.55M}
        & $s{=}0.5$: 16,384 tokens; CE delta $0.004003$ \\
        & & $s{=}0.75$: 16,384 tokens; CE delta $0.020491$ \\
        \cmidrule(lr){1-3}
        GPT-2 full model & 4.19M (factorization) + 16.38M (fine-tuning) & Final SWD-FT replacement (20.57M total tokens) \\
        \cmidrule(lr){1-3}
        Zero-data SWD & 0 & Fixed zero-data factors at each evaluated sparsity \\
        \hline
    \end{tabular}
    \caption{Data used for SWD replacement-quality results and the fixed SWD replacements used for circuit extraction. CE delta is replacement CE minus dense-model CE. The third column identifies the fixed replacement whose units are subsequently scored and ablated; it does not count the task data used for circuit scoring and evaluation.}
    \label{tab:swd-data-circuit-checkpoints}
\end{table}

In the primary GPT-2 and Qwen2.5 single-matrix comparisons, $s=0.5$ is the CE-matched SWD setting and $s=0.75$ is a higher-sparsity reference. Both fixed SWD settings are included in the Qwen3.5-27B circuit evaluation. In the MLP-replacement experiment, $s=0.5$ is the CE-matched SWD setting and $s=0.75$ is a higher-sparsity reference; SWD composes the two projection factorizations as specified in Appendix~\ref{app:fullmlp-alignment}. The full-model experiment instead compares methods at a matched active-weight budget, and zero-data SWD uses no calibration data (Appendix~\ref{app:zero-data-setting}).

\paragraph{Circuit cost.}
Each selected unit is one SWD bottleneck unit. Its read and write vectors are $\mA_{:,i}$ and $\mB_{i,:}$, so its active-edge count is $\|\mA_{:,i}\|_0+\|\mB_{i,:}\|_0$.

For the full-model experiment, approximation errors from the 48 replaced weight matrices accumulate. We therefore apply model-level fixed-support fine-tuning after sparse factorization. All pretrained parameters and binary factor masks remain frozen, and gradients update only factor values at existing nonzero locations. Fine-tuning thus preserves both the $26{,}770{,}800$-nonzero budget and the read/write connectivity used for circuit-cost accounting.

\begin{itemize}[leftmargin=1.5em,itemsep=0.25em,topsep=0.35em]
    \item \textbf{Objective and data.} Fixed-support fine-tuning minimizes autoregressive next-token cross-entropy on 4.19M unique training tokens. We train with batch size 8 for 2,000 steps, totaling 16.38M consumed tokens.
    \item \textbf{Optimizer.} We use AdamW with peak learning rate $5\times10^{-7}$, $(\beta_1,\beta_2)=(0.9,0.95)$, $\epsilon=0.1$, and weight decay $10^{-3}$.
    \item \textbf{Schedule and gradient control.} The learning rate warms up linearly over the first 1\% of steps and then decays linearly to zero. Before each optimizer step, the global factor-gradient RMS is normalized to $1.0$.
    \item \textbf{Precision and seed.} Fine-tuning uses bfloat16 training, float32 evaluation, and seed 0.
\end{itemize}

\subsection{Unit Scoring and Circuit Selection}
\label{app:scoring-protocol}

Candidate units are scored under the unpruned replacement. SWD exposes bottleneck units, Transcoders expose hidden features, and the VPD variants expose parameter components. For the sparse-pretrained model, which has no explicit replacement bottleneck, we prune latent channels following its native setup \citep{gao2025weightsparse}.

Let $g_{\tau}(M,\vx)$ be the task logit margin for example $\vx$, computed from the next-token logits as the average correct-answer logit minus the average distractor-answer logit, and let $z_i(\vx)$ be the activation of candidate unit $i$ (for SWD, $z_i(\vx)=\vh_{\ell,t}(\vx)\mA_{:,i}$ at the task-scoring position). The train-split mean activation used for ablation is
\[
    \mu_i = \mathbb{E}_{\vx \in \mathcal{D}_{\tau,\mathrm{train}}}\left[z_i(\vx)\right].
\]
The default score is positive first-order task-margin attribution \citep{bushnaq2026interpreting,syed2023attribution},
\[
    a_i = \mathbb{E}_{\vx \in \mathcal{D}_{\tau,\mathrm{train}}}
    \left[\max\!\left(0,\, z_i(\vx)\frac{\partial g_{\tau}}{\partial z_i(\vx)}\right)\right].
\]
Here $\partial g_{\tau}/\partial z_i$ is the gradient of the task margin with respect to unit $i$'s activation, so $a_i$ accumulates only positive first-order contributions to the margin. We sort each method's candidate units once by $a_i$ on \texttt{circuit\_train}; this fixed ranking defines the nested top-$k$ prefixes evaluated on held-out \texttt{circuit\_test}.

\subsection{Data Sources and Splits}
\label{app:data-sources}

\subsubsection{Language-Model Data}

All data-dependent replacement methods use FineWeb-Edu for fitting or training \citep{lozhkov2024fineweb-edu}. SWD uses subsets of this corpus to fit its sparse factors, whereas the trained baselines use it for optimization. Replacement quality is evaluated on a fixed, disjoint FineWeb-Edu split shared by all methods within each comparison. Accordingly, data used counts calibration tokens for SWD and optimizer-consumed tokens for the trained baselines. Method-specific data amounts and the checkpoints used for circuit extraction are reported in Appendices~\ref{app:swd-optimization} and~\ref{app:baseline-settings}. The zero-data experiment uses no text data.

\subsubsection{Circuit-Task Data}

Circuit extraction uses fixed public examples from Edge-Pruning \citep{bhaskar2024edge} and auto-circuit \citep{conmy2023automated,miller2024transformercircuitfaithfulnessmetrics}. These task examples are separate from the FineWeb-Edu data above and are not counted as replacement fitting or training data. The training split is used to estimate unit scores and mean-ablation values; the test split is used only to evaluate the resulting fixed top-$k$ sets.

\begin{table}[H]
    \centering
    \small
    \setlength{\tabcolsep}{3pt}
    \begin{tabular}{p{0.16\linewidth}p{0.27\linewidth}p{0.15\linewidth}p{0.15\linewidth}p{0.15\linewidth}}
        \hline
        Task & Public source & GPT-2 & Qwen2.5 & Qwen3.5-27B \\
        \hline
        Greater-than
        & \href{https://github.com/princeton-nlp/Edge-Pruning/tree/a918efd8fffcabe9b5853f6e236722e9ec0f4688}{Edge-Pruning, \texttt{a918efd8}}
        & 80,000 / 12,240
        & Not used
        & Not used \\
        IOI
        & \href{https://github.com/UFO-101/auto-circuit/tree/8fc5afd88b0231c8579a00c2a41c07b4eb5d950a}{auto-circuit, \texttt{8fc5afd8}}
        & 600 / 200
        & 600 / 200
        & 600 / 200 \\
        Docstring
        & \href{https://github.com/UFO-101/auto-circuit/tree/8fc5afd88b0231c8579a00c2a41c07b4eb5d950a}{auto-circuit, \texttt{8fc5afd8}}
        & 600 / 200
        & 600 / 200
        & 600 / 200 \\
        Gendered pronoun
        & \href{https://github.com/princeton-nlp/Edge-Pruning/tree/a918efd8fffcabe9b5853f6e236722e9ec0f4688}{Edge-Pruning, \texttt{a918efd8}}
        & 3,024 / 378
        & 3,024 / 378
        & 2,887 / 370 \\
        \hline
    \end{tabular}
    \caption{Public circuit-task sources and train/test example counts. The linked repository revisions fix the source data used in our experiments.}
    \label{tab:circuit-data-sources}
\end{table}

GPT-2 uses all four tasks. Qwen2.5 and Qwen3.5-27B use IOI, docstring, and gendered pronoun. Greater-than is omitted for Qwen because its numerical answers do not satisfy the single-token next-token evaluation used in these experiments. The Qwen3.5-27B counts report the examples retained after model-specific tokenization compatibility checks.

\subsection{Baseline Configurations}
\label{app:baseline-settings}

Shared data-usage axes count optimizer-replay tokens for trained baselines and calibration tokens for SWD. They compare data used and do not equate optimization steps or FLOPs.
We match replacement CE to within $0.001$ unless a curve is explicitly included as a reference with its CE delta reported.

Table~\ref{tab:baseline-comparison} summarizes the representation, sparsity, optimization, data requirements, and scope of each method. The subsections below provide the exact configurations and checkpoints used in each experiment.

\subsubsection{Transcoder}
\label{app:transcoder-settings}

The Transcoder baselines use sparse hidden-feature replacements for the same target surface as the corresponding SWD comparison. Below we report the settings needed to identify the plotted Transcoder lines and the checkpoints used for circuit extraction.

\paragraph{Circuit cost.}
Each selected unit is one Transcoder hidden feature. Its active edges are the nonzero encoder and decoder weights connected to that feature.

\paragraph{GPT-2 single-matrix replacement.}
For the GPT-2 single-matrix experiment, the Transcoder replaces the layer 8 MLP output projection.
\begin{itemize}[leftmargin=1.4em,itemsep=0.1em,topsep=0.2em]
    \item \textbf{Target:} GPT-2 Small layer 8 \texttt{mlp.c\_proj}.
    \item \textbf{Widths:} TC3k: Transcoder with hidden width 3072, TC12k: Transcoder with hidden width 12,288.
    \item \textbf{Feature sparsity penalty:} L1 coefficient $5\times10^{-5}$, seed 42.
    \item \textbf{Replacement-quality curve:} measured from 1,024 to 4.19M training tokens.
    \item \textbf{Checkpoint for circuit extraction:} 2.048M training tokens; CE deltas are 0.000529 for TC3k and 0.000979 for TC12k.
\end{itemize}

\paragraph{Qwen2.5 down-projection scaling.}
For the Qwen2.5 scaling experiment, each Transcoder replaces the middle-layer MLP down-projection of the corresponding model size: layer 12 for Qwen2.5-0.5B, layer 14 for Qwen2.5-1.5B, and layer 18 for Qwen2.5-3B.
\begin{itemize}[leftmargin=1.4em,itemsep=0.1em,topsep=0.2em]
    \item \textbf{Widths:} TC3k and TC12k, with hidden widths 3,072 and 12,288.
    \item \textbf{Feature sparsity penalty:} L1 coefficient $5\times10^{-5}$, seed 42.
    \item \textbf{Replacement-quality curve:} measured from 1,024 to 33.55M training tokens.
    \item \textbf{Checkpoints for circuit extraction:} Qwen2.5-0.5B uses 25.17M training tokens for both widths, with CE deltas 0.001654 for TC3k and 0.002032 for TC12k. Qwen2.5-1.5B uses 33.55M tokens for TC3k and 25.17M for TC12k, with CE deltas $-0.000057$ and $-0.000098$, respectively. Qwen2.5-3B uses 25.17M tokens for both widths, with CE deltas 0.001516 for TC3k and 0.001229 for TC12k.
\end{itemize}

\paragraph{Qwen3.5-27B down projection.}
For Qwen3.5-27B, each Transcoder replaces layer 31 \texttt{mlp.down\_proj}.
\begin{itemize}[leftmargin=1.4em,itemsep=0.1em,topsep=0.2em]
    \item \textbf{Widths:} TC24k and TC48k, with hidden widths 24,576 and 49,152.
    \item \textbf{Feature sparsity penalty:} L1 coefficient $5\times10^{-5}$, seed 42.
    \item \textbf{Replacement-quality curves:} measured from 1,024 tokens; TC24k is shown through 134.22M training tokens and TC48k through 67.11M training tokens.
    \item \textbf{Checkpoints for circuit extraction:} 134.22M training tokens for TC24k and 67.11M training tokens for TC48K, with CE deltas 0.000398 and 0.000695, respectively.
\end{itemize}

\paragraph{GPT-2 MLP replacement.}
For the MLP-replacement comparison, we use the standard TC24k Transcoder setting with hidden width 24,576, L1 coefficient $1.4\times10^{-4}$, and seed 42. Circuit extraction uses only the 6.29M-token checkpoint, with CE delta 0.003456. The complete replacement-quality trajectory is reported in Appendix~\ref{app:fullmlp-alignment}.

\subsubsection{VPD-KL and VPD-Recon-CI}
\label{app:vpd-recon-ci-setting}

VPD-KL uses the original VPD component parameterization, learned causal-importance (CI) mask, and model-level KL fidelity term. VPD-Recon-CI retains the same parameterization and mask but replaces this term with local activation reconstruction. The differing fidelity terms are
\[
\begin{aligned}
\mathcal{L}_{\mathrm{fid}}^{\mathrm{VPD\text{-}KL}}
&=
\mathbb{E}_{\vx,t}
\!\left[
D_{\mathrm{KL}}
\!\left(
p_{\mathrm{dense},t}(\cdot\mid \vx)
\,\|\,
p_{\mathrm{VPD\text{-}KL},t}(\cdot\mid \vx)
\right)
\right], \\
\mathcal{L}_{\mathrm{fid}}^{\mathrm{VPD\text{-}Recon\text{-}CI}}
&=
\mathbb{E}_{\vx,t,j}
\!\left[
\left(
\widehat{y}^{\mathrm{CI}}_{\ell,t,j}(\vx)-y_{\ell,t,j}(\vx)
\right)^2
\right].
\end{aligned}
\]
Here $\vx$ denotes a token sequence, $t$ a next-token position, $\ell$ the target module, and $j$ one coordinate of that module's output. The distributions $p_{\mathrm{dense},t}$ and $p_{\mathrm{VPD\text{-}KL},t}$ are the dense and VPD-KL next-token distributions. The values $y_{\ell,t,j}$ and $\widehat{y}^{\mathrm{CI}}_{\ell,t,j}$ are the corresponding dense and CI-masked replacement outputs; the expectation in the reconstruction objective averages over all batch elements, token positions, and output coordinates, matching the elementwise MSE used in training. The first objective constrains the final model output; the second directly constrains the local replacement. The equations isolate the fidelity terms; mask-minimality regularization is specified below. The CI network predicts a per-component mask, after which the masked rank-one components are recombined into the replacement output. VPD-KL is therefore reported only as a replacement-quality reference when its CE cannot be matched.

\paragraph{Circuit cost.}
Each selected unit is one VPD parameter component. Its active edges are the nonzero entries in that component's read and write vectors. The CI predictor is not part of this read/write edge count.

\paragraph{GPT-2 single-matrix replacement.}
For the GPT-2 single-matrix experiment, VPD-Recon-CI replaces the layer 8 MLP output projection.
\begin{itemize}[leftmargin=1.4em,itemsep=0.1em,topsep=0.2em]
    \item \textbf{Target:} GPT-2 Small layer 8 \texttt{mlp.c\_proj}.
    \item \textbf{Components and CI predictor:} $C=3{,}072$ components, seed 0, and a vector-MLP CI predictor of hidden width 128.
    \item \textbf{Objective:} MSE between the CI-masked replacement and dense projection outputs, plus a CI minimality penalty of $10^{-4}$.
    \item \textbf{Checkpoint for replacement quality and circuit extraction:} 4.19M training tokens, with CE delta 0.000873. We reuse this fixed checkpoint for both replacement-quality and circuit results.
\end{itemize}

\paragraph{Qwen2.5 down-projection scaling.}
For the Qwen2.5 scaling experiment, VPD-Recon-CI replaces the same middle-layer MLP down-projection used by SWD and the Transcoders.
\begin{itemize}[leftmargin=1.4em,itemsep=0.1em,topsep=0.2em]
    \item \textbf{Targets:} layer 12 \texttt{mlp.down\_proj} for Qwen2.5-0.5B, layer 14 for Qwen2.5-1.5B, and layer 18 for Qwen2.5-3B.
    \item \textbf{Components and objective:} $C=12{,}288$ components and the same local CI-masked reconstruction objective as in the GPT-2 single-matrix experiment.
    \item \textbf{Checkpoints for circuit extraction:} 16.78M training tokens for Qwen2.5-0.5B, 4.19M for Qwen2.5-1.5B, and 16.78M for Qwen2.5-3B. Their CE deltas are 0.000729, 0.001186, and 0.000906, respectively. The corresponding circuit results are reported in Appendix~\ref{app:qwen-vpdrecon-extension}.
\end{itemize}

\paragraph{Qwen3.5-27B down projection.}
For Qwen3.5-27B, VPD-Recon-CI replaces layer 31 \texttt{mlp.down\_proj}.
\begin{itemize}[leftmargin=1.4em,itemsep=0.1em,topsep=0.2em]
    \item \textbf{Components and CI predictor:} $C=12{,}288$ components, seed 0, and a vector-MLP CI predictor of hidden width 12.
    \item \textbf{Objective:} MSE between the CI-masked replacement and dense projection outputs, plus a CI minimality coefficient of $3\times10^{-4}$.
    \item \textbf{Checkpoint for circuit extraction:} 200.00M training tokens, with CE delta 0.001847. It is retained as the parameter-decomposition reference; the corresponding circuit results are reported in Appendix~\ref{app:qwen35-circuit}.
\end{itemize}

\paragraph{GPT-2 MLP replacement.}
For the MLP-replacement experiment, VPD-Recon-CI replaces the complete layer 8 MLP.
\begin{itemize}[leftmargin=1.4em,itemsep=0.1em,topsep=0.2em]
    \item \textbf{Components and objective:} $C=3{,}072$ components with the local MLP-reconstruction objective.
    \item \textbf{Replacement-quality curve:} measured from 1,024 to 16.78M training tokens.
    \item \textbf{Checkpoint for circuit extraction:} 4.19M training tokens, with CE delta 0.003522.
    \item \textbf{VPD-KL reference:} $C=3{,}072$, seed 0, and the native model-level KL objective. Its replacement-quality curve is measured from 4,096 to 1B training tokens; it is not used for circuit extraction. Full results are reported in Appendix~\ref{app:fullmlp-alignment}.
\end{itemize}

\subsubsection{Sparse-Pretrained Reference}
\label{app:sparse-pretrain-setting}

The sparse-pretraining reference is used in the GPT-2 Small full-model comparison. Unlike the other baselines, it is a separately trained sparse model with GPT-2 Small dimensions, rather than a sparse replacement of the Hugging Face GPT-2 checkpoint.

\paragraph{Circuit cost.}
For circuit extraction, the 3,072 post-GELU MLP channels in each layer form 36,864 candidate units; attention and residual channels are not candidate units. Each selected unit is one MLP hidden channel. Its active edges are the nonzero incoming and outgoing weights connected to that channel. For channel $j$, this count is
\[
 \left\|\mW_{\mathrm{c\_fc}}[:,j]\right\|_0+
 \left\|\mW_{\mathrm{c\_proj}}[j,:]\right\|_0+
 \mathbf{1}\!\left[b_{\mathrm{c\_fc},j}\ne0\right].
\]

\paragraph{GPT-2 full-model comparison.}
\begin{itemize}[leftmargin=1.4em,itemsep=0.1em,topsep=0.2em]
    \item \textbf{Architecture:} 12 layers, residual width 768, 12 attention heads of width 64, MLP width 3,072, context length 1,024, and vocabulary size 50,257. The model uses learned positional embeddings, LayerNorm, tied token-embedding and language-model-head weights, biases, zero dropout, and standard \texttt{nn.GELU}. Activation sparsity is disabled. The use of \texttt{nn.GELU} differs from the \texttt{gelu\_new} implementation in the Hugging Face GPT-2 checkpoint used by SWD.
    \item \textbf{Training data:} FineWeb-Edu documents are separated by end-of-sequence tokens and packed into 1,024-token examples; a disjoint FineWeb-Edu split is held out for evaluation. The global batch contains 128 examples, or 131,072 token presentations per optimizer update.
    \item \textbf{Evaluation checkpoints:} The replacement-quality and circuit results use separately trained models with identical configurations, each evaluated after 2,883,584,000 token presentations. Their held-out CE values are 3.450415 and 3.450323, respectively. The complete training schedule contains 4,999,872,512 token presentations, so the evaluated models are not terminal checkpoints.
    \item \textbf{Matched active-weight budget:} At the evaluated data point, the scheduled density is 0.315194486. The 48 transformer-block matrices contain 26,770,920 nonzeros (realized density 0.315194307), compared with 26,770,800 nonzeros in SWD-FT; the difference is 120 weights. This accounting excludes biases, normalization parameters, embeddings, and the tied language-model head. Across all trainable two-dimensional matrices, including token and positional embeddings, 39,184,479 of 124,318,464 entries are nonzero.
\end{itemize}

\paragraph{Optimization and sparsification.}
Training uses seed 0, bfloat16, and AdamW with base learning rate $3\times10^{-4}$, $(\beta_1,\beta_2)=(0.9,0.95)$, $\epsilon=0.1$, and weight decay $10^{-3}$. Let $n$ denote cumulative token presentations, with $n_0=49{,}938{,}432$, $n_1=3{,}999{,}793{,}152$, and $N=4{,}999{,}872{,}512$. The scheduled nonzero density is
\[
 d(n)=
 \begin{cases}
 1,&n<n_0,\\
 0.2^{(n-n_0)/(n_1-n_0)},&n_0\leq n<n_1,\\
 0.2,&n\geq n_1.
 \end{cases}
\]
Thus density remains one during the first 1\% of training, decays exponentially to the target density 0.2 by 80\% of training, and then remains fixed. The learning rate is
\[
 \operatorname{lr}(n)=3\times10^{-4}\frac{h(n)}{\sqrt{d(n)}},
 \qquad
 h(n)=
 \begin{cases}
 n/n_0,&n<n_0,\\
 1-(n-n_0)/(N-n_0),&n\geq n_0.
 \end{cases}
\]
After every optimizer update, each trainable two-dimensional matrix is magnitude-projected to its scheduled nonzero budget while retaining at least one weight per output row. The gradient routine computes a global RMS over all available parameter-gradient entries and, when this RMS exceeds $10^{-8}$, divides all gradients by $\mathrm{RMS}+10^{-5}$.

\section{Single-Matrix Replacement Results}

\subsection{Replacement Quality beyond Cross-Entropy}
\label{app:original-vpd-kl-longtail}

Section~\ref{sec:single_layer_replacement} measures replacement quality primarily by cross-entropy delta. We additionally report two complementary metrics on GPT-2 Small layer 8 \texttt{mlp.c\_proj} and the three Qwen2.5 \texttt{mlp.down\_proj} surfaces to verify that the comparison is not specific to CE.

\paragraph{Metrics.} Let $\mY_{\mathrm{dense}}$ and $\mY_{\mathrm{rep}}$ stack the dense and replacement outputs of the target projection over all evaluated examples and token positions. The \emph{activation relative MSE} is a local, single-layer measure of how well the replacement reproduces that projection's output,
\[
    \mathrm{relMSE}
    = \frac{\lVert \mY_{\mathrm{rep}} - \mY_{\mathrm{dense}}\rVert_F^2}
           {\lVert \mY_{\mathrm{dense}}\rVert_F^2}.
\]
The numerator and denominator are aggregated over the complete evaluation set before taking the ratio; we do not average per-example relative errors.
The \emph{KL} is a global measure that inserts the replacement back into the full model and compares next-token distributions on a held-out language-model split,
\[
    \mathrm{KL}
    = \mathbb{E}_{t}\,\mathrm{KL}\!\big(p^{\mathrm{dense}}_t \,\|\, p^{\mathrm{rep}}_t\big),
\]
with the dense model as the reference distribution and the expectation taken over held-out tokens $t$. Lower is better for both metrics. The x-axis reports data used in tokens, rather than compute, and uses a broken log scale so the short-horizon comparison and the VPD-KL long tail are both visible. We report activation relative MSE first because it measures the replaced layer directly, and KL second because it measures how that local error propagates to the model's output.

Both metrics reproduce the CE ordering across all single-matrix replacements. SWD reaches low error with little data, whereas Transcoder and VPD-Recon-CI approach it after roughly $10^{6}$ optimizer-replay tokens. VPD-KL is the weakest local replacement because its native objective targets model-level logits instead of the projection output (Appendix~\ref{app:vpd-recon-ci-setting}). No checkpoint on its extended GPT-2 trajectory, from 4,096 to 1.6384B tokens, meets the CE-matching criterion, showing that the fidelity gap is not an early-training artifact.

\begin{figure}[H]
    \centering
    \includegraphics[width=\linewidth]{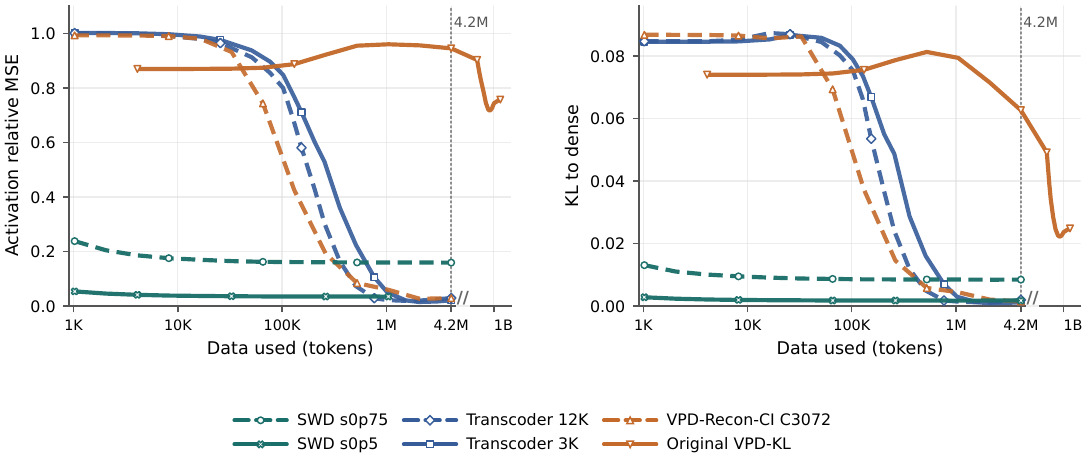}
    \caption{GPT-2 Small layer 8 \texttt{mlp.c\_proj} single-matrix replacement quality for all methods. Left: activation relative MSE (local). Right: KL to the dense model (global). Lower is better; both panels use the same broken-x policy so the short-horizon comparison and the VPD-KL long tail are visually aligned. VPD-KL is a native-objective replacement-quality reference line and is not a circuit baseline.}
    \label{fig:original-vpd-kl-activation}
\end{figure}

\begin{figure}[H]
    \centering
    \includegraphics[width=\linewidth]{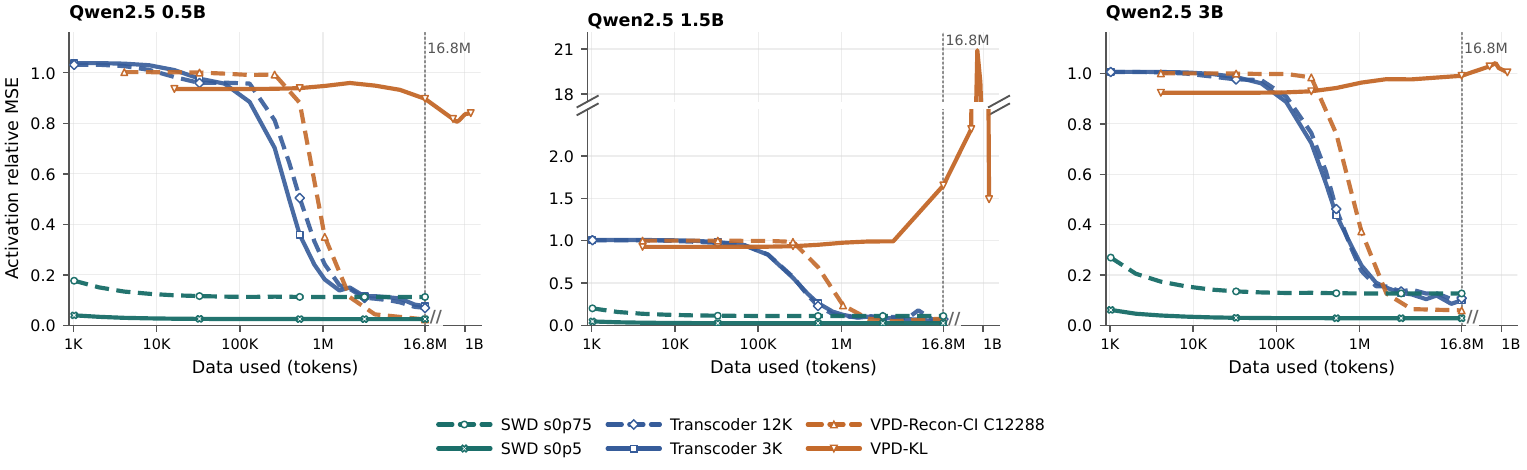}
    \caption{Qwen2.5 0.5B/1.5B/3B single-matrix \texttt{mlp.down\_proj} activation relative MSE for all methods (lower is better). The panels use the same broken-axis style as the KL companion: 0.5B and 3B use a broken x-axis after the 16.8M-token shared horizon, and 1.5B also uses a y-axis break to keep the low-error region readable despite the VPD-KL long-tail spike. VPD-KL is included to contextualize replacement quality.}
    \label{fig:qwen-relmse-companion}
\end{figure}

\begin{figure}[H]
    \centering
    \includegraphics[width=\linewidth]{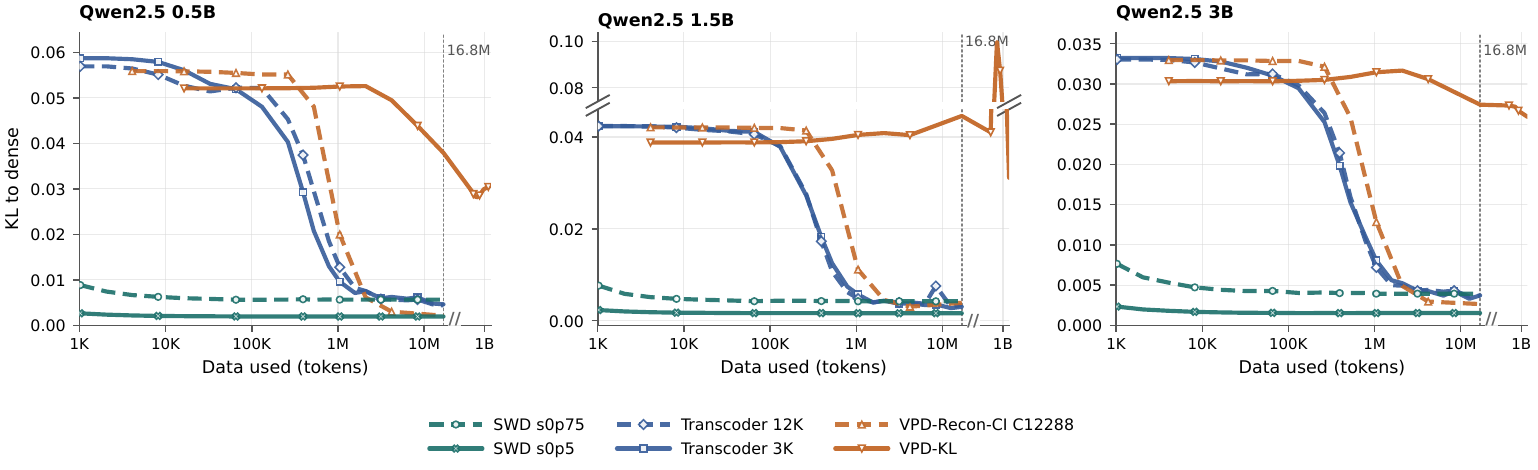}
    \caption{Qwen2.5 0.5B/1.5B/3B single-matrix \texttt{mlp.down\_proj} KL to the dense model for all methods (lower is better), with the same broken-axis convention as Figure~\ref{fig:qwen-relmse-companion}.}
    \label{fig:qwen-kl-companion}
\end{figure}

\subsection{GPT-2 Selected-Unit Circuit Results}
\label{app:gpt2-cproj-unit-companion}

\begin{figure}[H]
    \centering
    \includegraphics[width=\linewidth]{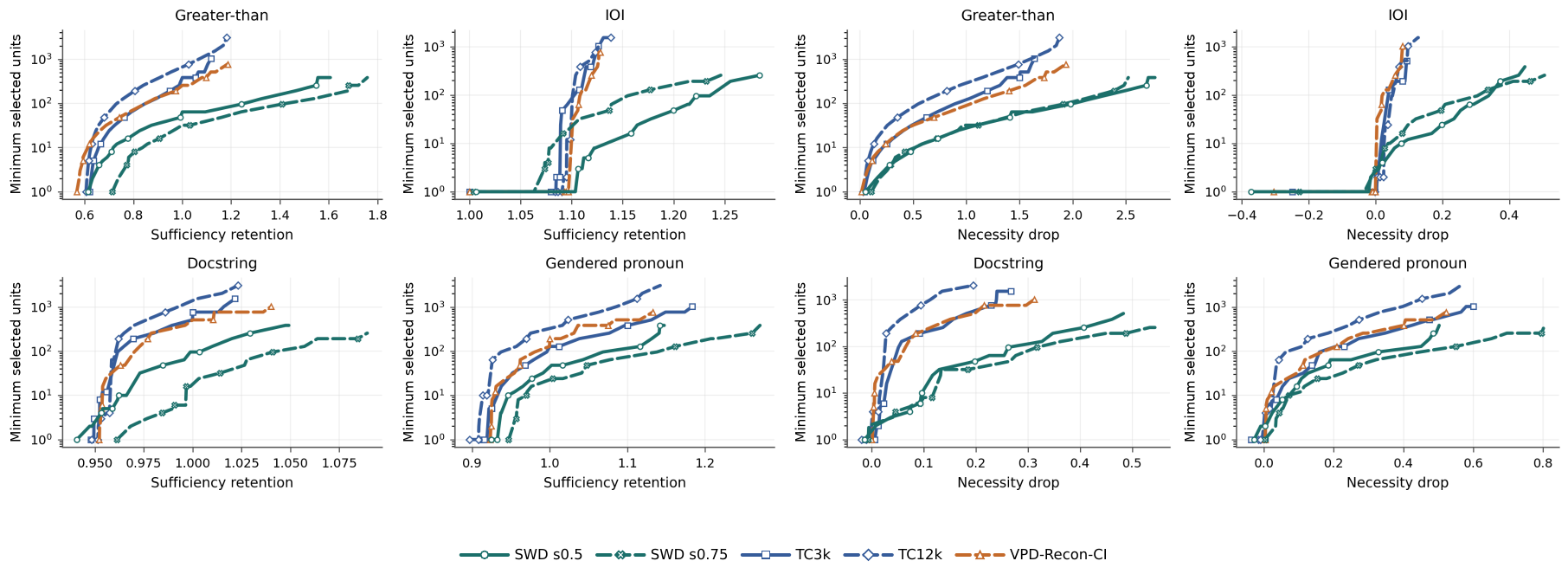}
    \caption{Companion to Figure~\ref{fig:gpt2-cproj-circuit} with selected units as the cost axis. The x-axis is the target threshold and the y-axis is the minimum selected units needed to reach it (lower-right is better). Left: sufficiency. Right: necessity drop. The same SWD advantage holds on the selected-unit axis.}
    \label{fig:gpt2-cproj-circuit-units}
\end{figure}

Figure~\ref{fig:gpt2-cproj-circuit-units} presents the GPT-2 Small layer 8 \texttt{mlp.c\_proj} circuit results from Figure~\ref{fig:gpt2-cproj-circuit}, using selected units rather than active edges as the cost axis. The main CE-matched comparison uses SWD ($s{=}0.5$), TC3k, TC12k, and VPD-Recon-CI; a sparser SWD ($s{=}0.75$) is retained as a non-CE-matched sparse control.

\subsection{Robustness under Zero Ablation}
\label{app:gpt2-zero-ablation}

The main circuit results replace ablated units with their mean activation on
\texttt{circuit\_train}. To test whether SWD's advantage depends on using mean
ablation rather than zero ablation, we repeat the GPT-2 single-matrix circuit
evaluation from Section~\ref{sec:single_layer_replacement} using zero ablation. We keep the
method checkpoints, unit rankings, top-$k$ sets, task examples, and definitions
of selected units and active edges fixed. The only change is the intervention
value: for sufficiency, units outside the selected set are set to zero; for
necessity, the selected units are set to zero. No unit scores or rankings are
recomputed.

\begin{figure}[H]
    \centering
    \includegraphics[width=\linewidth]{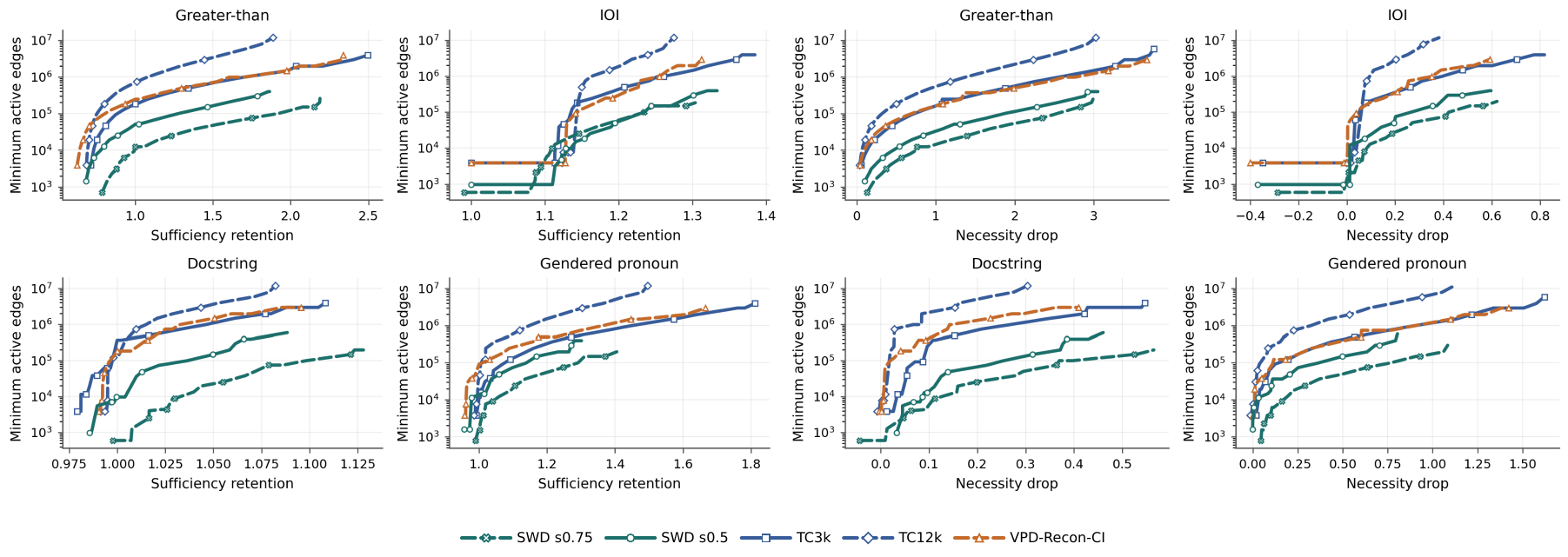}
    \caption{GPT-2 single-matrix circuit results under zero ablation, using minimum active edges as the cost. The checkpoints, unit rankings, and top-$k$ sets are identical to those in Figure~\ref{fig:gpt2-cproj-circuit}; only the ablation value changes from the training-set mean to zero. Left: sufficiency. Right: necessity drop.}
    \label{fig:gpt2-cproj-zero-ablation-edges}
\end{figure}

Figure~\ref{fig:gpt2-cproj-zero-ablation-edges} shows that SWD's active-edge
advantage persists under zero ablation. At sufficiency retention of at least
$1.0$, SWD with $s=0.75$ has the lowest observed active-edge cost on all four
tasks; on GreaterThan, it requires $12{,}303$ active edges, compared with
$184{,}320$ for TC3k. The necessity results show the same overall separation,
with the SWD curves reaching comparable necessity drops at lower active-edge
costs than the trained baselines.

\begin{figure}[H]
    \centering
    \includegraphics[width=\linewidth]{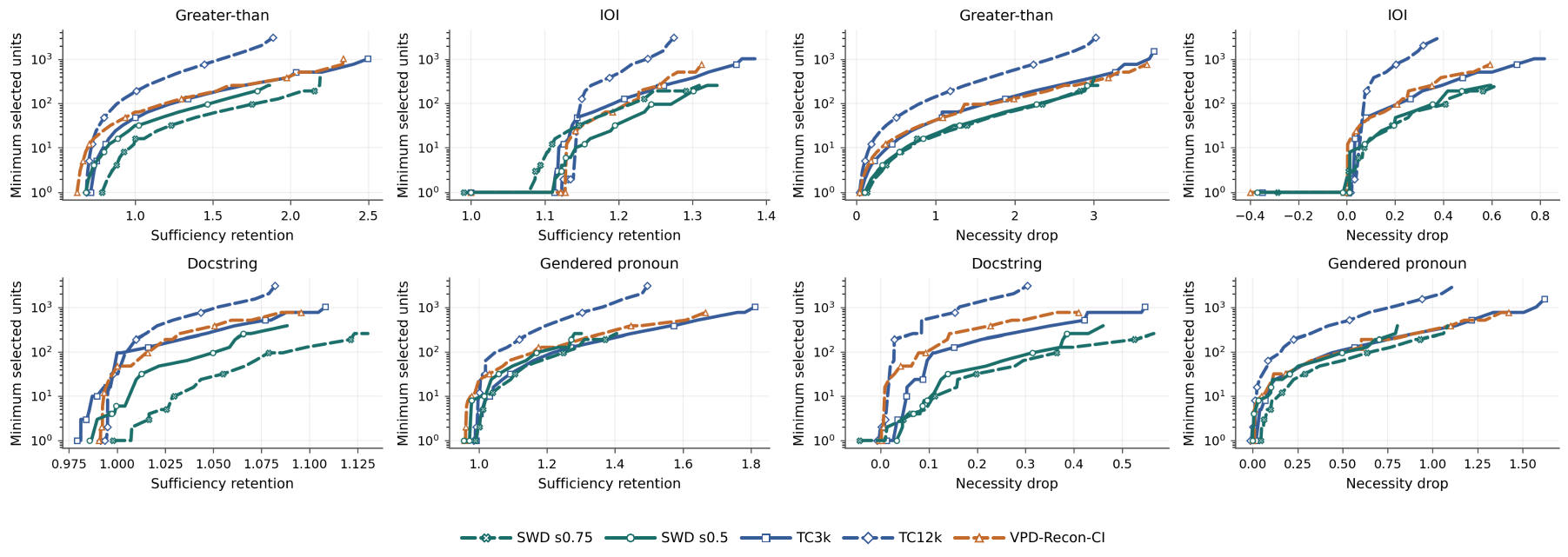}
    \caption{The same fixed-ranking zero-ablation comparison using minimum selected units as the cost. Left: sufficiency. Right: necessity drop.}
    \label{fig:gpt2-cproj-zero-ablation-units}
\end{figure}

The selected-unit comparison in
Figure~\ref{fig:gpt2-cproj-zero-ablation-units} shows the same pattern.
Together, these results show that SWD's advantage in the GPT-2 single-matrix
experiment is not specific to using mean activation as the ablation reference.

\subsection{Qwen2.5 Circuit Results}
\label{app:qwen-vpdrecon-extension}

\begin{figure}[H]
    \centering
    \includegraphics[width=0.98\linewidth]{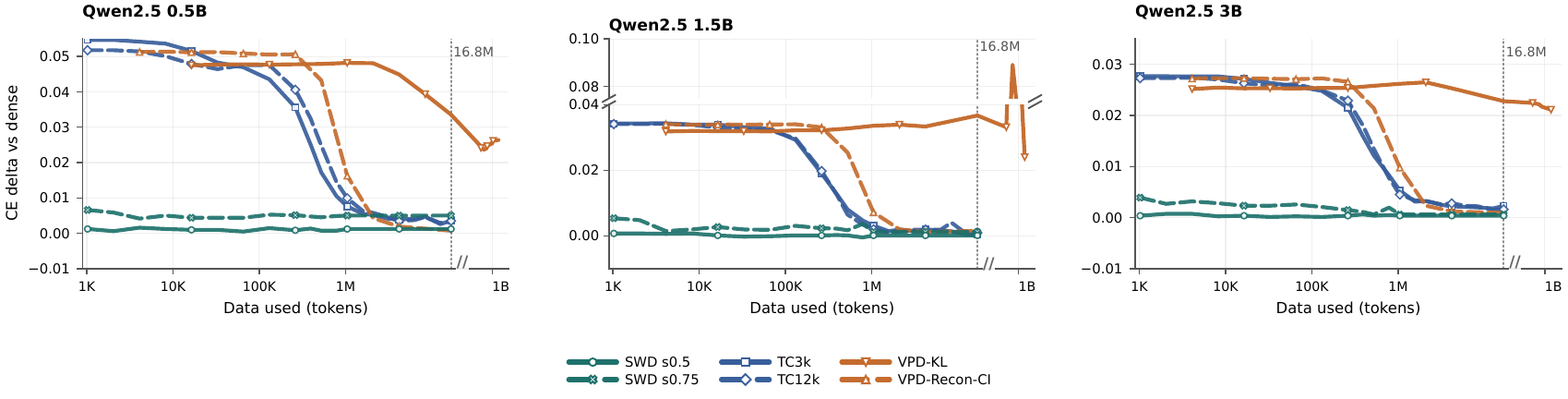}
    \caption{Qwen2.5 single-matrix replacement quality across 0.5B, 1.5B, and 3B models. SWD reaches low CE delta with substantially fewer tokens than the trained baselines.}
    \label{fig:qwen25-ce-all-sizes-app}
\end{figure}

\begin{figure}[H]
    \centering
    \includegraphics[width=\linewidth]{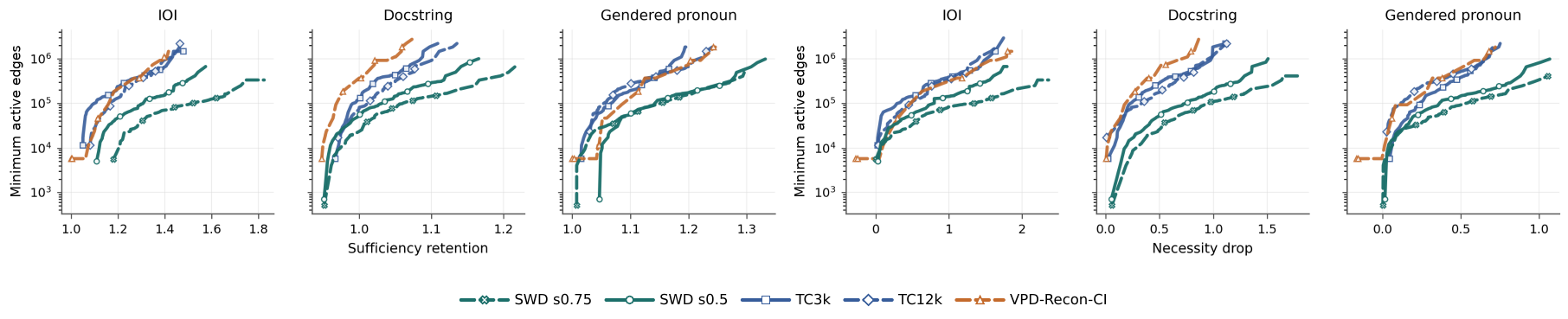}
    \caption{Qwen2.5 0.5B single-matrix circuit results, with active edges as the cost axis (lower-right is better). Left: sufficiency. Right: necessity drop.}
    \label{fig:qwen05-vpdrecon-circuit}
\end{figure}

\begin{figure}[H]
    \centering
    \includegraphics[width=\linewidth]{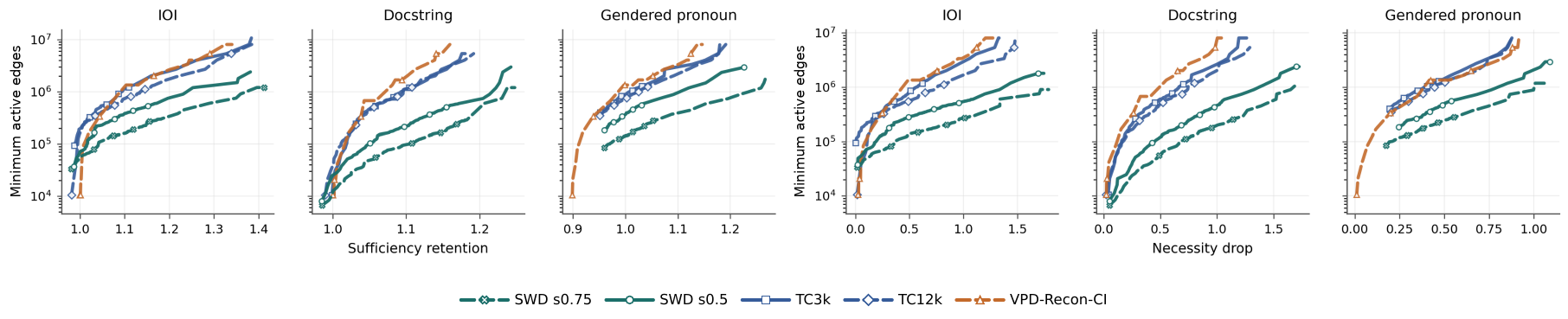}
    \caption{Qwen2.5 1.5B single-matrix circuit results, with active edges as the cost axis (lower-right is better). Left: sufficiency. Right: necessity drop. The 3B results are in the main text (Figure~\ref{fig:qwen-circuit}).}
    \label{fig:qwen15-vpdrecon-circuit}
\end{figure}

The Qwen2.5 comparison targets the middle-layer \texttt{mlp.down\_proj} in each model size. VPD-Recon-CI uses $C=12{,}288$ components and the same local reconstruction objective as in the GPT-2 experiment. Table~\ref{tab:qwen-vpdrecon-setting} specifies the target modules and fixed checkpoints used for circuit extraction. VPD-KL appears only in the replacement-quality figures because its fidelity is insufficient for a controlled circuit comparison. Figure~\ref{fig:qwen25-ce-all-sizes-app} reports replacement quality for all three model sizes. Figures~\ref{fig:qwen05-vpdrecon-circuit} and~\ref{fig:qwen15-vpdrecon-circuit} complement the 3B result in the main text with the 0.5B and 1.5B circuit results.

\begin{table}[H]
    \centering
    \small
    \setlength{\tabcolsep}{4pt}
    \begin{tabular}{p{0.4\linewidth}p{0.3\linewidth}p{0.12\linewidth}}
        \hline
        Model and target module & Checkpoint for circuit extraction & CE delta \\
        \hline
        Qwen2.5 0.5B, layer 12 \texttt{mlp.down\_proj} & 16.78M tokens used & 0.000729 \\
        Qwen2.5 1.5B, layer 14 \texttt{mlp.down\_proj} & 4.19M tokens used & 0.001186 \\
        Qwen2.5 3B, layer 18 \texttt{mlp.down\_proj} & 16.78M tokens used & 0.000906 \\
        \hline
    \end{tabular}
    \caption{Qwen2.5 VPD-Recon-CI settings. All rows use $C=12{,}288$ components and replace one \texttt{mlp.down\_proj}.}
    \label{tab:qwen-vpdrecon-setting}
\end{table}

\begin{figure}[H]
    \centering
    \includegraphics[width=\linewidth]{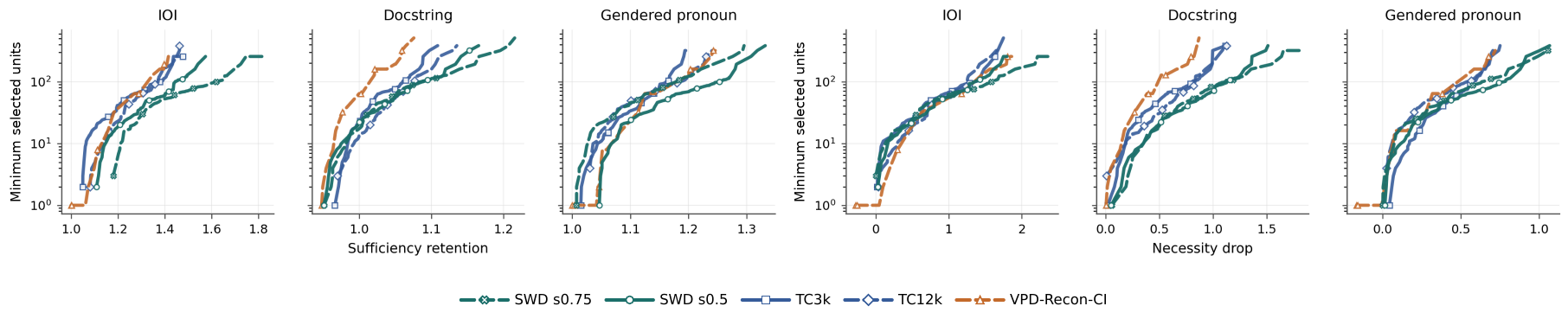}
    \caption{Qwen2.5 0.5B single-matrix circuit results, with selected units as the cost axis (lower-right is better). Companion to Figure~\ref{fig:qwen05-vpdrecon-circuit}. Left: sufficiency. Right: necessity drop.}
    \label{fig:qwen05-vpdrecon-circuit-units}
\end{figure}

\begin{figure}[H]
    \centering
    \includegraphics[width=\linewidth]{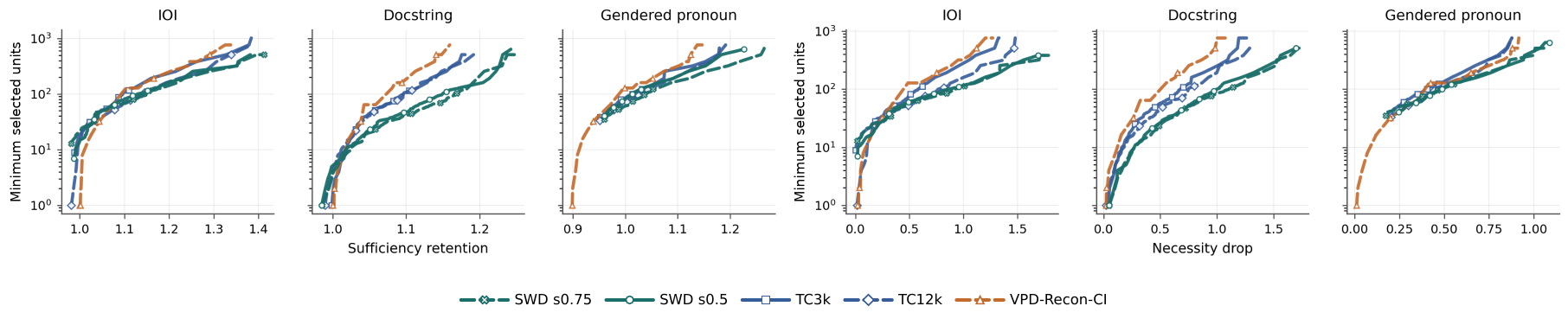}
    \caption{Qwen2.5 1.5B single-matrix circuit results, with selected units as the cost axis (lower-right is better). Companion to Figure~\ref{fig:qwen15-vpdrecon-circuit}. Left: sufficiency. Right: necessity drop.}
    \label{fig:qwen15-vpdrecon-circuit-units}
\end{figure}

\begin{figure}[H]
    \centering
    \includegraphics[width=\linewidth]{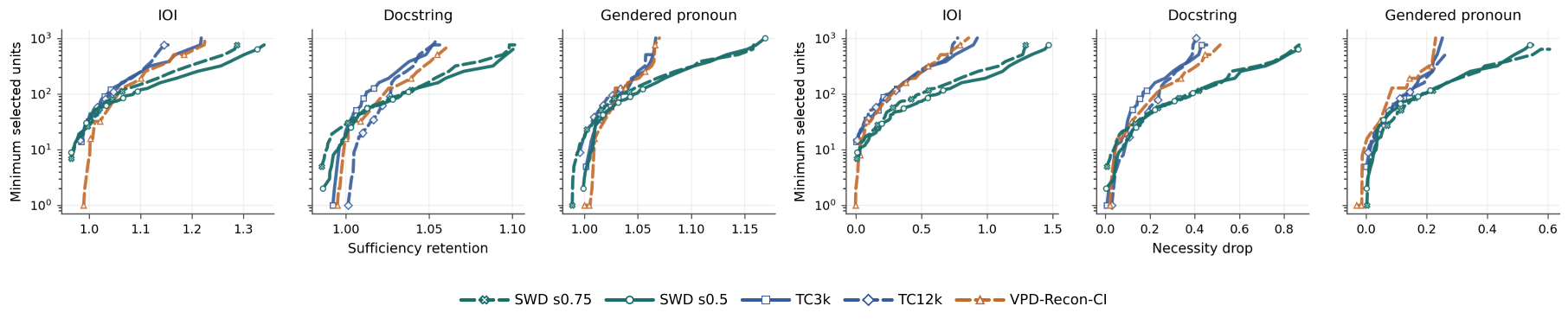}
    \caption{Qwen2.5 3B single-matrix circuit results, with selected units as the cost axis (lower-right is better). Companion to the 3B active-edge results in the main text (Figure~\ref{fig:qwen-circuit}). Left: sufficiency. Right: necessity drop.}
    \label{fig:qwen3b-vpdrecon-circuit-units}
\end{figure}

Figures~\ref{fig:qwen05-vpdrecon-circuit-units}--\ref{fig:qwen3b-vpdrecon-circuit-units} recast the same three Qwen2.5 circuit sweeps with selected units as the cost axis, separating selected-unit count from read/write connectivity, as in the GPT-2 companion (Figure~\ref{fig:gpt2-cproj-circuit-units}).

\subsection{Qwen3.5-27B Circuit Results}
\label{app:qwen35-circuit}

\begin{figure}[H]
    \centering
    \includegraphics[width=\linewidth]{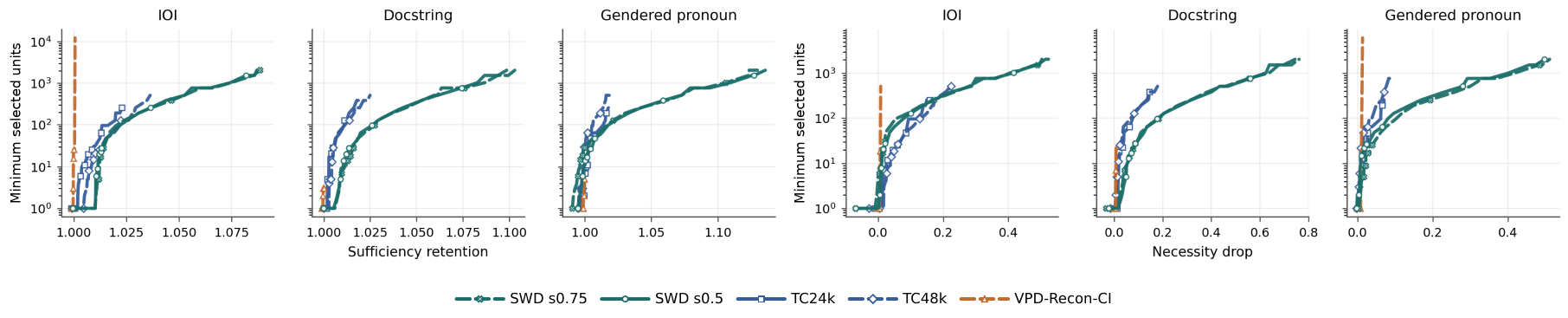}
    \caption{Companion to Figure~\ref{fig:qwen35-circuit} with selected units as the cost axis. The x-axis is the target threshold and the y-axis is the minimum selected units needed to reach it (lower-right is better). Left: sufficiency. Right: necessity drop.}
    \label{fig:qwen35-circuit-units}
\end{figure}

The Qwen3.5-27B comparison targets layer 31 \texttt{mlp.down\_proj}. All methods use the same IOI, docstring, and gendered-pronoun task data and the circuit-selection protocol in Appendix~\ref{app:scoring-protocol}. Figure~\ref{fig:qwen35-circuit} reports active edges, and Figure~\ref{fig:qwen35-circuit-units} reports the same circuit results using selected units. Table~\ref{tab:qwen35-circuit-settings} lists the fixed checkpoints used for circuit extraction.

\begin{table}[H]
    \centering
    \small
    \setlength{\tabcolsep}{4pt}
    \begin{tabular}{p{0.23\linewidth}p{0.25\linewidth}p{0.27\linewidth}p{0.12\linewidth}}
        \hline
        Method & Setting & Checkpoint for circuit extraction & CE delta \\
        \hline
        SWD & $s=0.5$ & 2,048 calibration tokens & $-0.000427$ \\
        SWD & $s=0.75$ & 1.05M calibration tokens & $-0.000448$ \\
        Transcoder & width 24,576 & 134.22M training tokens & $0.000398$ \\
        Transcoder & width 49,152 & 67.11M training tokens & $0.000695$ \\
        VPD-Recon-CI & $C=12{,}288$ & 200.00M training tokens & $0.001847$ \\
        \hline
    \end{tabular}
    \caption{Qwen3.5-27B checkpoints used for circuit extraction. CE delta is measured against the same dense model and held-out language-model data.}
    \label{tab:qwen35-circuit-settings}
\end{table}

Relative to SWD with $s=0.5$, the CE-delta differences are $0.000022$ for SWD with $s=0.75$, $0.000824$ for TC24k, $0.001121$ for TC48k, and $0.002274$ for VPD-Recon-CI. TC48k is slightly outside the $0.001$ matching tolerance. VPD-Recon-CI has the largest difference and is retained as a reference for the parameter-decomposition baseline.

\section{Full-Model Replacement}
\label{app:gpt2-fullmodel-ce-curve}

\begin{figure}[H]
    \centering
    \includegraphics[width=0.78\linewidth]{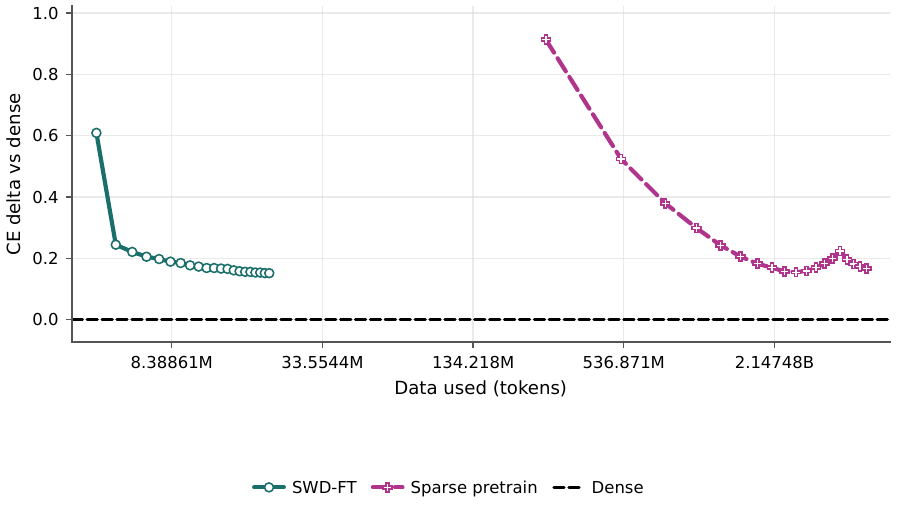}
    \caption{Held-out CE of GPT-2 Small full-model replacements versus data used in tokens (lower is better). The SWD-FT curve begins at the 4.19M-token factorization point and adds fixed-support fine-tuning tokens thereafter; the sparse-pretraining curve counts tokens consumed during sparse pretraining. The dotted horizontal line marks the dense model. At an approximately matched active-weight budget, factorizing and fine-tuning an existing dense checkpoint brings SWD-FT to CE $3.44$ after $20.6$M tokens, less than $1\%$ of the $2.884$B sparse-pretraining tokens used to reach CE $3.45$.}
    \label{fig:gpt2-fullmodel-ce-loss}
\end{figure}

Figure~\ref{fig:gpt2-fullmodel-ce-loss} shows the replacement-quality trajectory behind Table~\ref{tab:gpt2-fullmodel-replacement-summary}. The SWD point before fine-tuning appears at 4.19M calibration tokens with CE 3.90. Fixed-support fine-tuning then lowers the CE along the SWD-FT curve without changing the active-weight budget, reaching CE 3.44 at 20.6M total tokens. This matches the sparse-pretraining baseline's CE 3.45, which is reached after 2.884B tokens.

\section{MLP Replacement}
\label{app:fullmlp-alignment}

The main single-matrix experiment constrains every method to replace \texttt{mlp.c\_proj}. Standard Transcoders normally replace the entire MLP, so we also compare on the GPT-2 Small layer 8 MLP input--output map. For SWD, this means factorizing the weight matrices of the two linear projections, \texttt{mlp.c\_fc} and \texttt{mlp.c\_proj}, while keeping the original GELU nonlinearity $\phi$ between them:
\[
    \widehat{\mathrm{MLP}}(\vh)
    = \phi\!\left(\vh\mA_{\mathrm{fc}}\mB_{\mathrm{fc}}+\vb_{\mathrm{fc}}\right)
      \mA_{\mathrm{proj}}\mB_{\mathrm{proj}}+\vb_{\mathrm{proj}}.
\]
Thus, SWD changes only the two linear projections, while the nonlinearity is the original GELU used by the dense model.

\begin{table}[H]
    \centering
    \small
    \setlength{\tabcolsep}{4pt}
    \begin{tabular}{p{0.20\linewidth}p{0.43\linewidth}p{0.29\linewidth}}
        \hline
        Method & Replacement configuration & Checkpoint for circuit extraction \\
        \hline
        SWD & Sequential factorization of \texttt{c\_fc} and \texttt{c\_proj}; $s\in\{0.5,0.75\}$ & 16,384 tokens used \\
        Transcoder & Standard MLP Transcoder (Appendix~\ref{app:transcoder-settings}) & 6.29M tokens used; CE delta 0.003456\\
        VPD-Recon-CI & Local MLP-reconstruction objective; $C=3{,}072$ & 4.19M tokens used\\
        VPD-KL & Native KL-logit objective; $C=3{,}072$, seed 0 & Replacement only \\
        \hline
    \end{tabular}
    \caption{Fixed configurations for the GPT-2 Small layer 8 MLP-replacement comparison. SWD tokens are calibration tokens; Transcoder and VPD-Recon-CI tokens are optimizer replay. VPD-KL is excluded from circuit extraction because its replacement fidelity is not matched.}
    \label{tab:gpt2-fullmlp-vpdrecon-setting}
\end{table}

Replacement-quality trajectories share a 16.78M-token horizon; the VPD-KL curve is extended to 1B tokens to test whether longer optimization closes its fidelity gap. Across CE delta, KL, and activation relative MSE, it remains the weakest replacement. As in the single-matrix experiment, we retain it for replacement-quality context but exclude it from task-circuit pruning.

\begin{figure}[H]

    \centering
    \includegraphics[width=\linewidth]{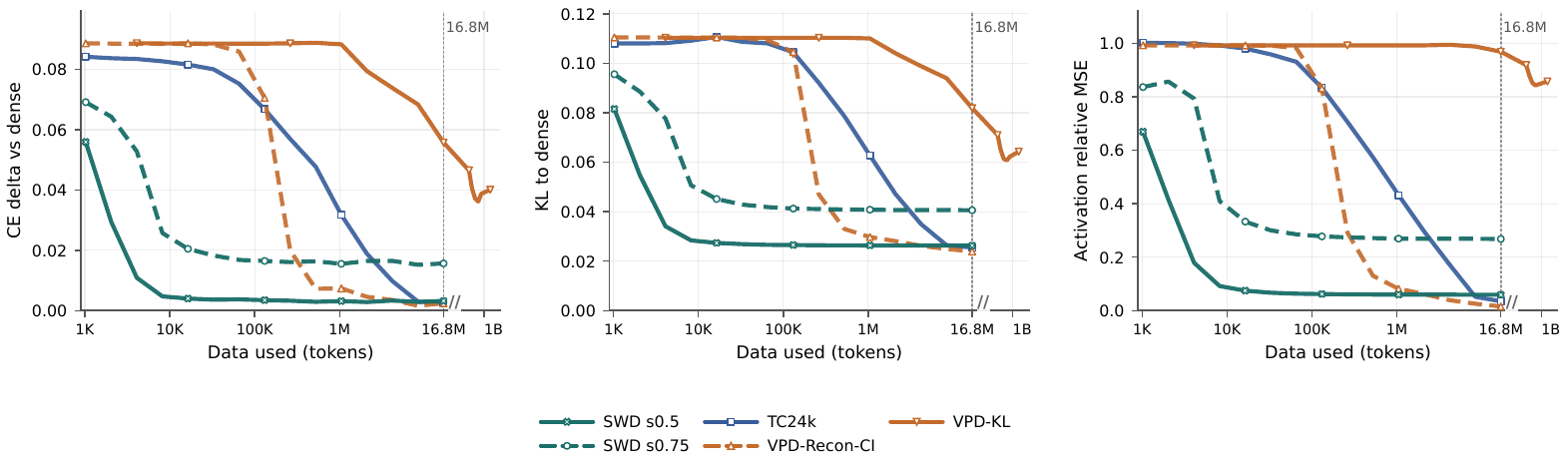}
    \caption{GPT-2 Small layer 8 MLP replacement quality for all methods (lower is better in every panel). Left: CE delta versus the dense model. Middle: KL to the dense model. Right: activation relative MSE at the target module. The broken x-axis keeps the 1K--16.8M shared comparison region readable and shows only the VPD-KL tail beyond the break. The native-objective VPD-KL curve serves as a replacement-quality reference and is excluded from the circuit-pruning figures; it begins at its first 4,096-token checkpoint.}
    \label{fig:gpt2-fullmlp-vpdrecon-token-quality}
\end{figure}

\begin{figure}[H]
    \centering
    \includegraphics[width=\linewidth]{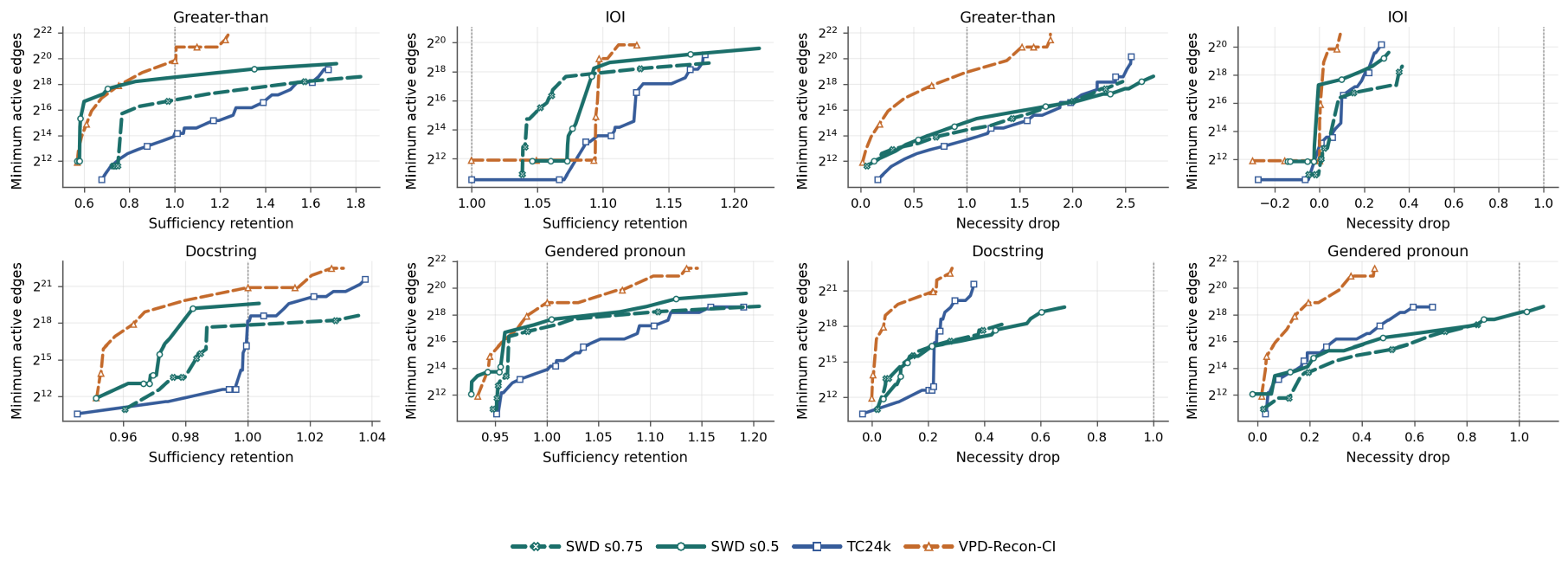}
    \caption{GPT-2 layer 8 MLP-replacement circuit results for the setting in Table~\ref{tab:gpt2-fullmlp-vpdrecon-setting}. For each curve, the x-axis is the target threshold and the y-axis is the minimum active edges needed to reach it (lower-right is better). Left: sufficiency. Right: necessity drop. Transcoder and VPD-Recon-CI use the fixed CE-matched 6.29M- and 4.19M-token checkpoints, respectively.}
    \label{fig:gpt2-fullmlp-circuit}
\end{figure}

\begin{figure}[H]
    \centering
    \includegraphics[width=\linewidth]{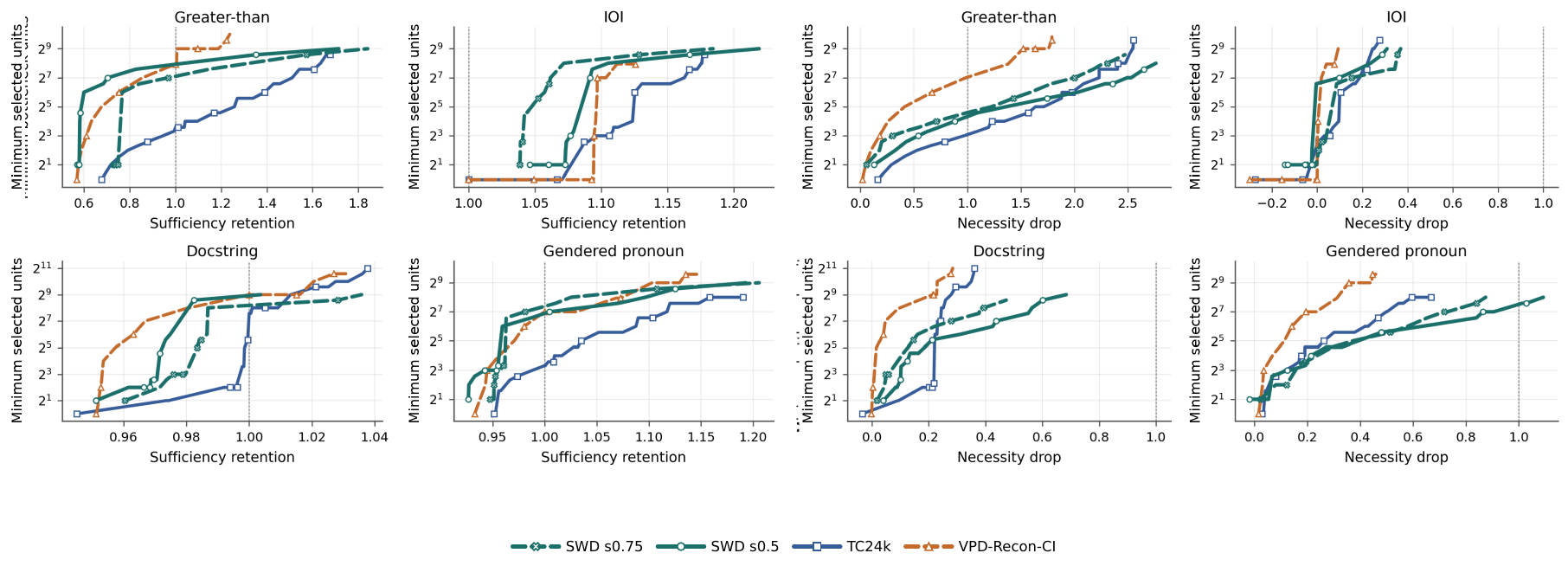}
    \caption{Companion to Figure~\ref{fig:gpt2-fullmlp-circuit} with selected units as the cost axis (lower-right is better), separating selected-unit count from read/write connectivity.}
    \label{fig:gpt2-fullmlp-total-units}
\end{figure}

We then apply the same circuit extraction protocol to the fixed MLP-replacement checkpoints in Table~\ref{tab:gpt2-fullmlp-vpdrecon-setting}. The primary matched comparison uses SWD with $s=0.5$, a single TC24k checkpoint at 6.29M tokens, and the 4.19M-token VPD-Recon-CI checkpoint; SWD with $s=0.75$ provides a higher-sparsity reference. Figures~\ref{fig:gpt2-fullmlp-circuit} and~\ref{fig:gpt2-fullmlp-total-units} report the resulting cost-quality tradeoff using active edges and selected units as the cost axes. On this broader surface, the results are more mixed than in the single-matrix setting: SWD remains competitive on the active-edge axis and often requires fewer edges to reach the same necessity drop, while the standard MLP Transcoder is competitive in several selected-unit and sufficiency comparisons. This comparison complements the single-matrix results by testing the same protocol on the MLP-replacement surface.

\section{Exact-Dense Reparameterization Controls}
\label{app:exact-dense-controls}

The exact dense controls in Figure~\ref{fig:gpt2-cproj-exact-dense-controls} reparameterize GPT-2 Small layer-8 \texttt{mlp.c\_proj} without approximation or sparsification. They test whether bottleneck re-expression alone yields SWD's low-cost circuits. The SVD control uses the same exact rank-one decomposition used by NaNA \citep{xue2026svd}. To measure how sparsity in the read and write vectors affects the number of selected units and active edges required, we keep the task data, positive first-order task-margin attribution, mean ablation, and figure plotting fixed across SVD, Random-B, and SWD. This shared protocol compares dense and sparse read/write connections without changing how units are ranked or ablated.
\paragraph{Control configurations.}
\begin{itemize}[leftmargin=1.4em,itemsep=0.1em,topsep=0.2em]
    \item \textbf{SWD ($s=0.5$):} fixed CE-matched decomposition on the same \texttt{mlp.c\_proj} surface; the primary matched-fidelity SWD line.
    \item \textbf{SWD ($s=0.75$):} higher-sparsity SWD reference on the same \texttt{mlp.c\_proj} surface.
    \item \textbf{SVD Exact:} full-rank dense SVD of \texttt{mlp.c\_proj} (rank 768, $d_{\mathrm{in}}=3072$, $d_{\mathrm{out}}=768$); an exact dense control that is neither low-rank nor sparse.
    \item \textbf{Random-B (seed mean):} random orthogonal exact dense control with $\mB=\mQ$, $\mA=\mW\mQ^\top$. Results are averaged over 10 runs (seeds 0--9), and the band shows one standard deviation above and below the mean.
\end{itemize}
Both dense controls use the same cost accounting: $k$ bottleneck units, or $k(d_{\mathrm{in}}+d_{\mathrm{out}})$ active edges. They exactly reparameterize the GPT-2 layer 8 \texttt{mlp.c\_proj} matrix, whereas the SWD lines expose sparse read/write structure.

For the full-rank SVD control, the audited effective orientation is
\[
    \mW_{\mathrm{eff}} = \mA_{\mathrm{svd}}\mB_{\mathrm{svd}},
    \qquad
    \mA_{\mathrm{svd}} = \mU\mSigma,\quad \mB_{\mathrm{svd}} = \mV^\top,
\]
using all rank-768 singular directions. For the Random-B control,
\[
    \mB_{\mathrm{rand}} = \mQ,
    \qquad
    \mA_{\mathrm{rand}} = \mW_{\mathrm{eff}}\mQ^\top,
    \qquad
    \mW_{\mathrm{eff}} = \mA_{\mathrm{rand}}\mB_{\mathrm{rand}},
\]
where $\mQ$ is a seeded random orthogonal matrix from Gaussian QR. In both cases $\mW_{\mathrm{eff}}$ reproduces the original \texttt{mlp.c\_proj} map exactly.

\begin{figure}[H]
    \centering
    \includegraphics[width=\linewidth]{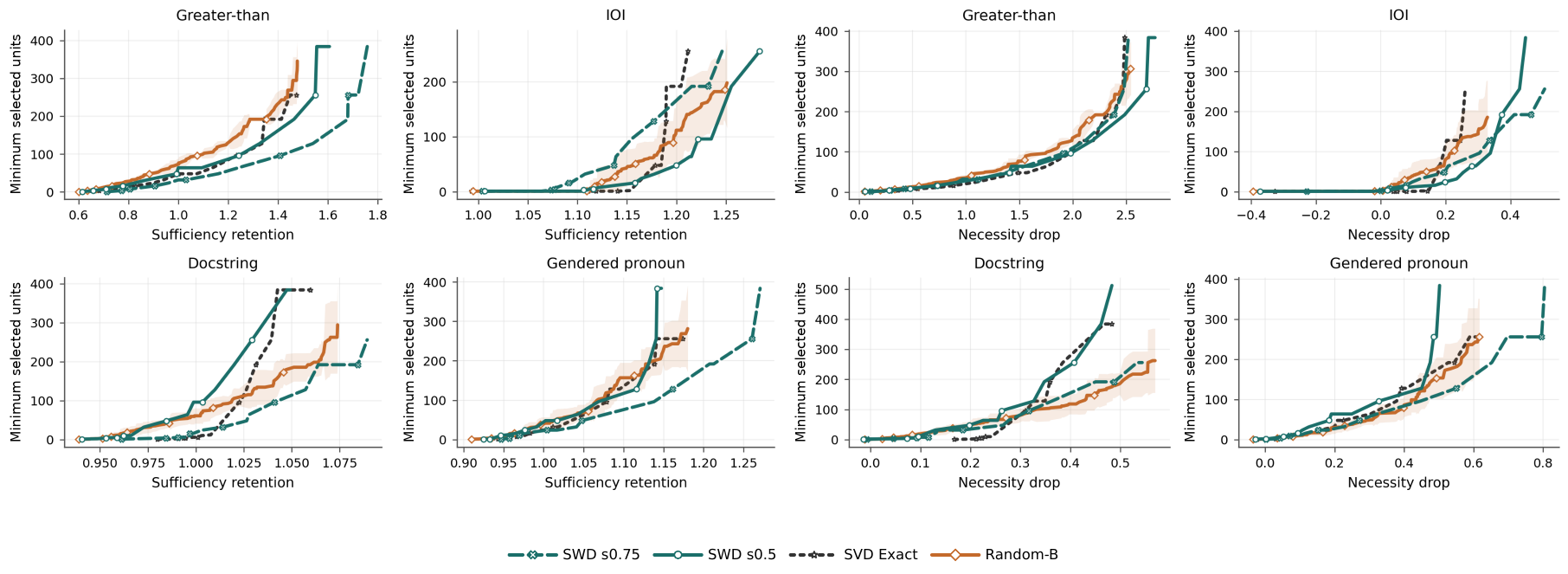}
    \caption{Companion to Figure~\ref{fig:gpt2-cproj-exact-dense-controls} with selected units as the cost axis (lower-right is better). Left: sufficiency. Right: necessity drop. Dense controls become competitive with SWD at specific low thresholds or tasks, but do not generally reproduce its low-cost circuit.}
    \label{fig:gpt2-cproj-exact-dense-controls-units}
\end{figure}

Both controls are numerically exact: SVD has mean CE delta $1.58\times10^{-6}$ and KL $1.81\times10^{-7}$; across 10 Random-B seeds, CE delta ranges from $-2.48\times10^{-6}$ to $3.40\times10^{-6}$ and mean KL is $1.89\times10^{-7}$. Their need for more active edges therefore cannot be attributed to poorer replacement fidelity. On the selected-unit axis (Figure~\ref{fig:gpt2-cproj-exact-dense-controls-units}), SVD and Random-B are competitive on IOI and at low docstring necessity levels, while SWD requires fewer units in most other regions. The larger and more consistent difference appears on the active-edge axis, where SWD's sparse read/write factors reduce the number of active connections per selected unit.

\clearpage
\section{Zero-Data Analysis}

\subsection{Factorization and Evaluation Setting}
\label{app:zero-data-setting}

To isolate the role of calibration activations, we keep the standard SWD solver, factor dimensions, and nonzero-budget schedule fixed and replace only the activation Gram matrix with the identity:
\[
    \mH^\top \mH \leftarrow \mI,
    \qquad
    \min_{\mA,\mB}\|\mW-\mA\mB\|_F^2
    \quad \mathrm{s.t.}\quad
    \|\mA\|_0+\|\mB\|_0\leq K .
\]
No calibration activations are used to choose the supports or fit their surviving values. We factorize the GPT-2 Small layer 8 \texttt{mlp.c\_proj} from the main single-matrix experiment over $s\in\{0.125,0.25,0.375,0.5,0.625,0.75,0.875\}$. Larger $s$ leaves fewer active scalars. The main-text fidelity plots compare each resulting product $\mA\mB$ with the corresponding activation-calibrated factorization at the same sparsity.

\subsection{Circuit Results}
\label{app:zero-data-task-frontiers}

\begin{figure}[H]
    \centering
    \includegraphics[width=\linewidth]{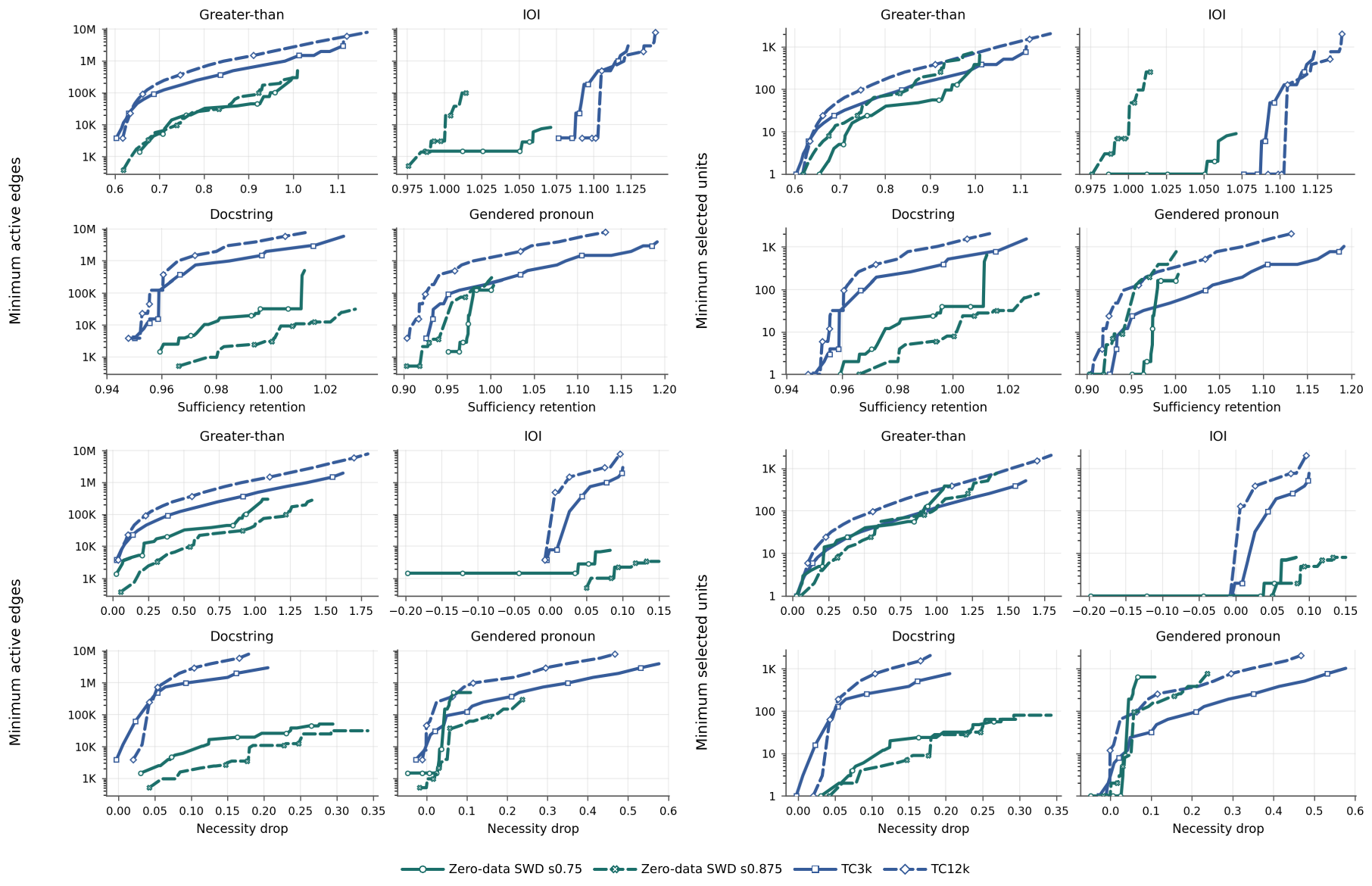}
    \caption{Zero-data identity-Gram SWD task-circuit results compared with fixed Transcoder reference settings. Each curve gives the minimum cost reaching the plotted threshold (lower is better). Top: sufficiency. Bottom: necessity drop. Left: active edges. Right: selected units.}
    \label{fig:zero-data-task-frontiers}
\end{figure}

Replacement fidelity alone does not establish that the bottleneck units remain useful after circuit pruning. We therefore apply the common unit-scoring and top-$k$ protocol from Section~\ref{sec:circuit-extraction-protocol} to every zero-data sparsity checkpoint. For each of greater-than, IOI, docstring, and gendered-pronoun, the bottleneck-unit ranking and ablation means are computed on \texttt{circuit\_train}, then frozen before sufficiency and necessity are evaluated on \texttt{circuit\_test}. Figure~\ref{fig:zero-data-task-frontiers} reports the minimum active edges and selected units needed to reach each held-out quality threshold. The Transcoder curves use the fixed checkpoints from the single-matrix experiments in Appendix~\ref{app:transcoder-settings} as contextual references.

The task results in Figure~\ref{fig:zero-data-task-frontiers} suggest that zero-data SWD factors are not merely good matrix approximations; their bottleneck units remain useful for circuit extraction after pruning. It also shows that many zero-data task-sparsity settings retain high held-out sufficiency at small cost: at $s=0.375$, the docstring circuit reaches $0.962$ test sufficiency with only 3 selected units, and the greater-than circuit reaches $0.986$ with 256 units. Sufficiency is easiest to retain on greater-than, docstring, and IOI, while gendered-pronoun is the weakest task, with fewer high-sufficiency low-cost points.

Useful circuit structure can therefore be recovered directly from pretrained weights without calibration data. Activation-aware SWD remains stronger under aggressive sparsification, but the zero-data result shows that its bottleneck units are not solely an artifact of a chosen calibration distribution.

\section{GreaterThan Case Study: Semantic Audit}
\label{app:qualitative-text-activations}



This appendix provides additional details for the semantic audit in Section~\ref{sec:qualitative-results}. We rank all $9{,}208$ valid \texttt{mlp.c\_proj} bottleneck units in GPT-2 Small using the GreaterThan attribution score defined in Appendix~\ref{app:scoring-protocol}. Figure~\ref{fig:greater-than-case-study} shows the two highest-ranked units in layers 6, 8, and 10, and Table~\ref{tab:greater-than-layer-ranks} extends the comparison to the top three units in each layer. The global ranks are computed over the full candidate pool.

For each bottleneck unit, we then evaluate 2,048 task-independent WikiText-2 tokens and collect the 20 surrounding text passages associated with its largest activation magnitudes. GPT-5.5 summarizes the recurring patterns in these passages into a short semantic hypothesis. The passages and semantic labels are not used for unit ranking or circuit selection.

\begin{table}[H]
    \centering
    \small
    \begin{tabular}{c c c c c l}
        \hline
        Layer & Bottleneck unit & Layer order & Global rank & Score & Semantic hypothesis \\
        \hline
        6 & c380 & 1 & 108 & 10.2 & measure/unit \\
        6 & c723 & 2 & 129 & 9.7 & punctuation/function \\
        6 & c558 & 3 & 152 & 9.3 & mixed lexical \\
        8 & c264 & 1 & 76 & 11.3 & quantity \\
        8 & c511 & 2 & 84 & 11.0 & measure/place \\
        8 & c728 & 3 & 123 & 9.8 & event/time \\
        10 & c757 & 1 & 32 & 14.4 & numeric/year \\
        10 & c627 & 2 & 63 & 11.8 & species/name \\
        10 & c481 & 3 & 85 & 10.9 & numeric/year \\
        \hline
    \end{tabular}
    \caption{Top-three GreaterThan bottleneck units in each displayed layer. Layer order is used only for visualization; global rank is over all $9{,}208$ candidates. Semantic hypotheses are task-independent and do not affect ranking.}
    \label{tab:greater-than-layer-ranks}
\end{table}

The six displayed bottleneck units have global ranks 32--129 and all fall within the global top-$512$ circuit prefix. Table~\ref{tab:qualitative-text-activations} gives representative activation contexts for these units. Four respond to numbers, quantities, or measurements, while the other two respond to punctuation or species names. The dominant pattern is therefore quantitative, with two units reflecting the surrounding syntactic or lexical context.

\begin{table}[H]
    \centering
    \small
    \begin{tabular}{p{0.16\linewidth}p{0.22\linewidth}p{0.52\linewidth}}
        \hline
        Bottleneck unit & Semantic hypothesis & Top-activating excerpts \\
        \hline
        L6 c380 & measure/unit & ``carapace length of 80--85 \textbf{millimetres}''; ``usually 23--38 \textbf{cm}''; ``a carapace length of 15 \textbf{mm}''. \\
        L6 c723 & punctuation/function & Repeated comma-token activations in clauses such as ``higher than that of H. americanus\textbf{,} and the European species'' and around numeric punctuation. \\
        L8 c264 & quantity & ``estimated that only \textbf{1} larva''; ``not normally deeper than \textbf{50 m} (160 ft)''; ``published in \textbf{1758}''. \\
        L8 c511 & measure/place & ``\textbf{48 kilometres or 30 miles northwest} of Gothenburg''; ``for up to \textbf{12 months}''; ``body length up to \textbf{60 centimetres}''. \\
        L10 c757 & numeric/year & ``published in \textbf{1758}''; ``Homarus Weber, \textbf{1795}''; ``Fabricius, \textbf{1775}''. \\
        L10 c627 & species/name & Repeated activations on species-name fragments in ``H. \textbf{americanus}'', ``H. \textbf{gammarus}'', and ``\textbf{Homarus} americanus''. \\
        \hline
    \end{tabular}
    \caption{Representative task-independent top-activation contexts for the six bottleneck units rendered in Figure~\ref{fig:greater-than-case-study}. Bold marks the token or phrase most relevant to each tentative semantic hypothesis.}
    \label{tab:qualitative-text-activations}
\end{table}
Overall, the semantic audit gives the GreaterThan circuit a more concrete interpretation. Most displayed bottleneck units respond to numbers, years, quantities, or measurements in task-independent text, consistent with their high GreaterThan attribution. Together with the held-out sufficiency and necessity results, this shows that the selected SWD bottleneck units are both behaviorally important and semantically recognizable.

\clearpage
\section{Targeted Editing of a Single SWD Bottleneck Unit}
\label{app:component-editing}

Our design adapts the model-editing template of making a localized low-rank weight update and evaluating both target efficacy and locality on unrelated inputs \citep{meng2022locating}. To remove replacement-fidelity differences from the comparison, every method starts from the same original dense GPT-2 Small layer-8 \texttt{mlp.c\_proj} weight $\mW$. The SWD intervention uses the read direction of one bottleneck unit to materialize the dense rank-one update induced by changing that unit's write direction. We include a random-unit control, a target-conditioned dense rank-one oracle, and rank-4 LoRA \citep{hu2022lora}. The source SWD factorization uses $s=0.75$, 768 bottleneck units, seed 0, and 131,072 WikiText-2 calibration tokens \citep{merity2017pointer}.

The main circuit experiments establish that SWD bottleneck units support targeted ablation. Here we ask a complementary question: can the read direction of one unit support a localized directional edit? We study the prompt
\begin{quote}
\texttt{The opposite of up is}\qquad
answer: \texttt{ down},\quad foil: \texttt{ left}.
\end{quote}
The dense model assigns the answer an answer--foil logit margin of $3.261101$. Let $\va_c=\mA_{:,c}\in\mathbb{R}^{3072}$ be the read vector of bottleneck unit $c$, let $\vh^\star\in\mathbb{R}^{1\times3072}$ be the final-position input to \texttt{mlp.c\_proj}, and let
\[
    \vd=\vu_{\mathrm{ans}}-\vu_{\mathrm{foil}},
    \qquad
    \overline{\vd}=\frac{\vd}{\lVert\vd\rVert_2},
\]
where $\vu_{\mathrm{ans}},\vu_{\mathrm{foil}}\in\mathbb{R}^{768}$ are the column-vector forms of the corresponding unembedding rows. For requested local answer-direction shift $\delta$, the SWD-guided edit is
\[
    \mW_c(\delta)
    =
    \mW+\alpha_c(\delta)\va_c\overline{\vd}^{\top},
    \qquad
    \alpha_c(\delta)
    =
    \frac{\delta}
    {(\vh^\star\va_c)\lVert\vd\rVert_2}.
\]
It therefore satisfies
\[
    \vh^\star\bigl(\mW_c(\delta)-\mW\bigr)\vd=\delta.
\]
This is the materialized dense increment induced by the corresponding write-row update, since
\[
    \mB'_{c,:}
    =
    \mB_{c,:}+\alpha_c(\delta)\overline{\vd}^{\top}
    \quad\Longrightarrow\quad
    \mA\mB'-\mA\mB
    =
    \alpha_c(\delta)\mA_{:,c}\overline{\vd}^{\top}.
\]
We apply this increment to the original dense $\mW$, rather than using $\mA\mB$ as the base, so that all editing methods are evaluated from the same model. The requested-shift grid is
\[
    \delta\in\{-2,-1,-0.5,0,0.5,1,2,3\}.
\]

Unit c205 was selected by screening eight candidate prompt--answer--foil triples for positive target activation and answer-direction write alignment, retaining the top 24 positive units per prompt and incorporating stored task importance and rank in the selection score. The \emph{random-unit} control c540 was sampled with seed 17 from units whose absolute target activation lies between the 35th and 85th percentiles. It uses the same answer direction and requested-shift calibration as c205. The \emph{target-conditioned dense rank-one oracle} replaces the SWD read vector by the normalized target activation,
\[
    \vr_{\mathrm{dense}}
    =
    \frac{\vh^{\star\top}}{\lVert\vh^\star\rVert_2},
    \qquad
    \mW_{\mathrm{dense}}(\delta)
    =
    \mW+\alpha_{\mathrm{dense}}(\delta)
    \vr_{\mathrm{dense}}\overline{\vd}^{\top},
\]
with $\alpha_{\mathrm{dense}}(\delta)$ calibrated by the same local answer-direction constraint. Rank-4 LoRA uses $\Delta\mW=\mL_A\mL_B$, with $\mL_A\in\mathbb{R}^{3072\times4}$ initialized as $0.01\mathcal{N}(0,1)$ and $\mL_B\in\mathbb{R}^{4\times768}$ initialized to zero. A separate LoRA update is trained for each requested final-margin shift using AdamW for 180 steps with learning rate $0.08$, zero weight decay, and an $L_2$ coefficient of $10^{-4}$. The Pareto comparison uses the realized final-margin change of every method.

At requested local shift $\delta=3$, the c205 coefficient is $\alpha_{205}=0.423220$, and the realized final answer--foil margin change is $0.215797$. Mean dense-to-edited final-token KL over the seven non-target prompts is $4.02375\times10^{-5}$. On the separate 504-token global evaluation sample, the corresponding mean KL is $1.96415\times10^{-5}$.

Figure~\ref{fig:component-edit-pareto} shows the expected ordering. The target-conditioned dense rank-one oracle gives the strongest tradeoff. The c205 edit is next: it opens farther vertically and remains farther left than the random-unit control. Rank-4 LoRA attains larger absolute margin changes, but at substantially higher non-target KL. Thus, on this audited target and probe set, an SWD-guided bottleneck-unit edit provides a more precise intervention than the random-unit and LoRA controls, though it does not match the target-conditioned dense rank-one oracle.


\clearpage
\section{Attention Bottleneck Units: From Static QK Geometry to Prompt-Local Effects}
\label{app:attention-components}

The workflow here combines the QK-circuit decomposition of attention \citep{elhage2021mathematical}, sparse weight-space attention decomposition for circuit tracing \citep{franco2024sparseattentiondecompositionapplied}, and feature-level QK attribution \citep{kamath2025tracing}. Because a pre-softmax attention score is bilinear in its query- and key-side inputs, an additive feature decomposition rewrites that score as a sum of query--key feature-pair terms; prior work interprets those terms on concrete prompts and validates them with causal interventions. SWD supplies parameter-side query and key bottleneck units for the same analysis. We additionally begin with a reconstruction check so that any unit-level interpretation is tied to an attention pattern that remains close to the dense head.

The full-model experiment in Section~\ref{sec:full-model-replacement} replaces every attention and MLP weight matrix across the transformer blocks. We use those factors to test whether sparse Q and K bottleneck units expose attention computations that are active on real tokens and sensitive to intervention. Following the motivation above, the analysis proceeds from reconstruction fidelity, to a weight-space candidate screen, to prompt-level replay and ablation.

\subsection{Attention Factorization Fidelity}

We first measure activation-weighted relative reconstruction SSE for the packed Q/K/V weight matrix \texttt{attn.c\_attn} and the attention output weight matrix \texttt{attn.c\_proj} in all 12 GPT-2 Small layers. As shown in Figure~\ref{fig:attention-reconstruction-static-screen} (top), the Q/K/V reconstruction error remains below $4\%$ in every layer. The attention output matrix is less accurate in the middle layers, where its error reaches approximately $6\%$. The prompt-level analysis below therefore uses the better-preserved \texttt{attn.c\_attn} factors and separately checks the reconstructed attention pattern against the dense head.

\subsection{Static QK Candidate Screen}

For one head $h$, write the reconstructed query and key at token positions $t$ and $u$ as
\[
\begin{aligned}
    \vq_t^h &= \sum_a z_a(\vh_t)\vb^{\mathrm{Q}}_{a,h},\\
    \vk_u^h &= \sum_b z_b(\vh_u)\vb^{\mathrm{K}}_{b,h}.
\end{aligned}
\]
Expanding their scaled dot product separates each attention score into bottleneck-unit-pair terms:
\[
\begin{aligned}
    \operatorname{score}_{tu}^h
        &= \sum_{a,b}z_a(\vh_t)z_b(\vh_u)\gamma_{ab}^h,\\
    \gamma_{ab}^h
        &= \frac{\langle \vb^{\mathrm{Q}}_{a,h},\vb^{\mathrm{K}}_{b,h}\rangle}{\sqrt{64}}.
\end{aligned}
\]
The static coefficient $\gamma_{ab}^h$ measures the interaction between a Q write direction and a K write direction; the activation product $z_a(\vh_t)z_b(\vh_u)$ determines whether that interaction contributes on a particular prompt and token pair.

\begin{figure}[H]
    \centering
    \includegraphics[width=0.74\linewidth]{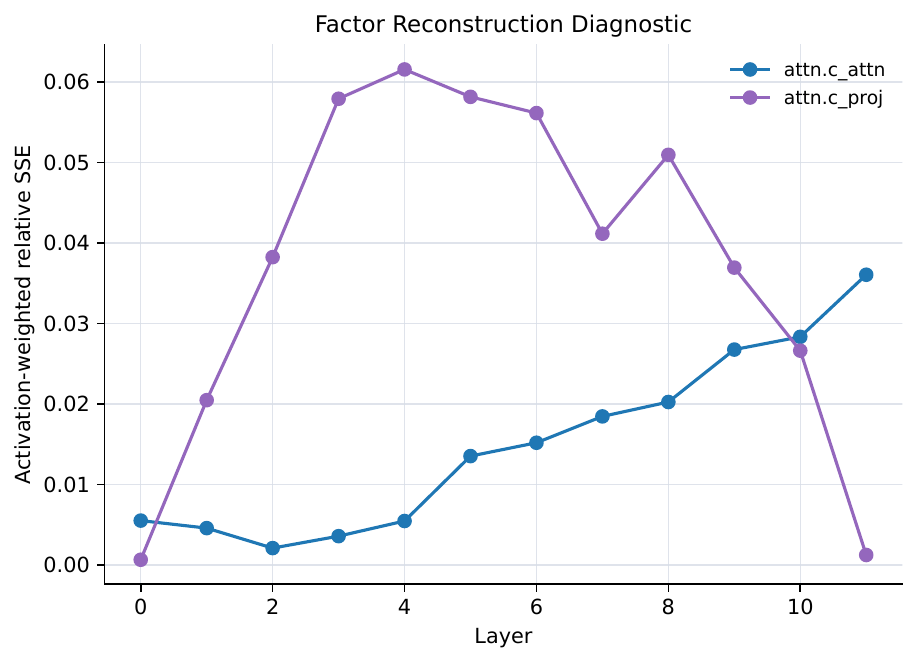}

    \vspace{4pt}
    \includegraphics[width=0.96\linewidth]{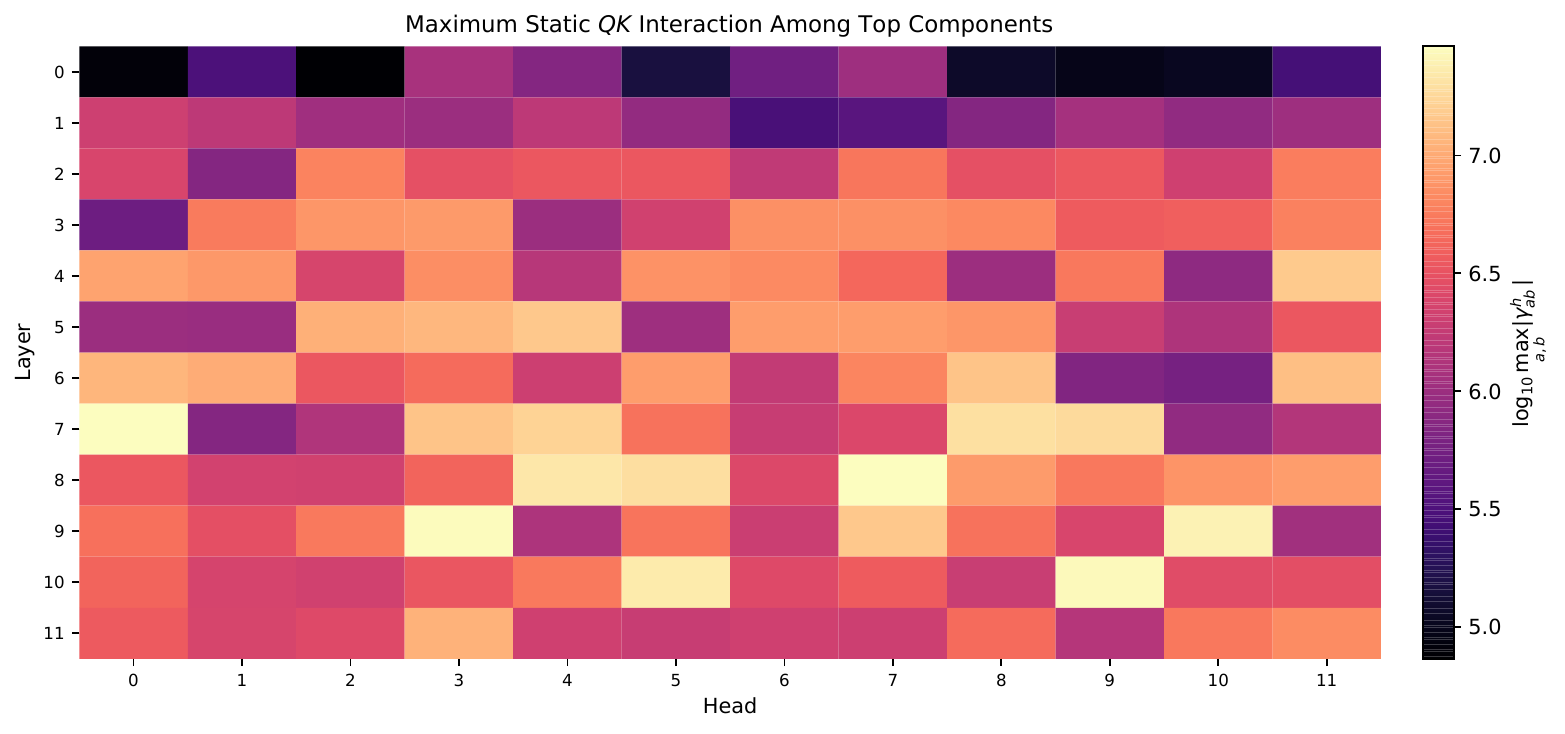}
    \caption{Attention reconstruction and static QK candidate screen. Top: activation-weighted relative reconstruction SSE for the packed Q/K/V and attention output matrices across GPT-2 Small. Bottom: the largest absolute static QK coefficient among the retained $16\times16$ bottleneck-unit pairs in each layer and head.}
    \label{fig:attention-reconstruction-static-screen}
\end{figure}

For each of 12 layers and 12 heads, we retain the 16 Q and 16 K bottleneck units with the largest head-specific write norms, evaluate all 256 cross-pairs, and summarize the head by its largest $|\gamma_{ab}^h|$. Figure~\ref{fig:attention-reconstruction-static-screen} (bottom) shows that the strongest candidates occur mainly in later layers. We advance the five strongest pairs in every head to prompt replay.

\subsection{Prompt Replay and Bottleneck-Unit Intervention}

We evaluate the five strongest static pairs from each layer and head on three prompts, for
\[
    12\ \text{layers}
    \times 12\ \text{heads}
    \times 5\ \text{pairs}
    \times 3\ \text{prompts}
    =2{,}160
\]
prompt--head--pair evaluations. The clearest response occurs on \texttt{When Mary gave John the book, he thanked} in layer 9, head 3, for pair q266$\times$k64.

\begin{figure}[H]
    \centering
    \includegraphics[width=0.72\linewidth]{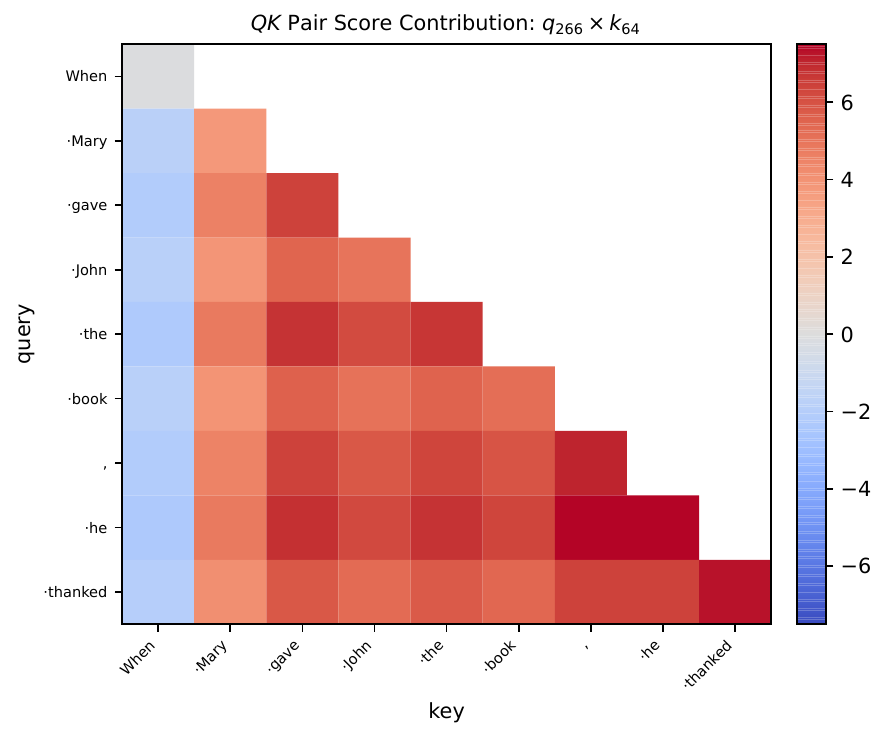}
    \caption{Token-level pre-softmax score contribution of q266$\times$k64 in layer 9, head 3. Each cell shows $z_{266}(\vh_t)z_{64}(\vh_u)\gamma_{266,64}^h$ for one causally valid query--key pair; the signed mean is $+4.26$.}
    \label{fig:qk-pair-contribution}
\end{figure}

Before interpreting the pair, we compare the dense attention pattern with the pattern obtained after replacing the packed Q/K/V weight matrix by its SWD reconstruction. Figure~\ref{fig:attention-replay-ablation} (top) shows that the main attention structure is retained; the mean dense-to-reconstructed attention KL is $0.0405$. We then evaluate the full pair contribution
\[
    z_{266}(\vh_t)z_{64}(\vh_u)\gamma_{266,64}^h
\]
at every causally valid query--key position. As shown in Figure~\ref{fig:qk-pair-contribution}, the pair contributes negatively toward the first key, \texttt{When}, and positively toward most later keys, with signed mean $+4.26$ over valid positions.

Finally, we remove q266 from the reconstructed Q slice and recompute the head. Attention shifts sharply toward the first token: relative to the intact reconstructed pattern, mean attention KL rises to $2.496$ and the maximum probability change is $0.775$ (Figure~\ref{fig:attention-replay-ablation}, bottom). This reversal agrees with the signed contribution map: removing q266 eliminates the unit's suppression of \texttt{When} and its support for later keys.

Together, these checks close the chain from static geometry to causal effect. SWD closely reconstructs the selected head, static QK geometry identifies q266$\times$k64, prompt replay shows a coherent token-level contribution, and ablating q266 reverses that pattern. This is a single selected attention case, but it shows that SWD's attention bottleneck units can expose a concrete and manipulable part of a head's computation.


\section{Interaction Patterns among MLP Bottleneck Units}
\label{app:component-interactions}

Feature-based circuit analyses express an MLP output as a sum of component contributions \citep{dunefsky2024transcoders}, while work on superposition shows that the corresponding features can overlap \citep{elhage2022toy}. More generally, feature interactions can be measured as departures from additivity introduced by nonlinear activations \citep{tsang2018detecting}. We therefore examine two complementary phenomena: correlation between additive output contributions after GELU, and contextual non-additivity when bottleneck-unit contributions are combined before GELU.

We analyze the two MLP weight matrices separately in GPT-2 Small layer 8. For the output matrix \texttt{mlp.c\_proj}, bottleneck-unit contributions are combined linearly after GELU, so we measure correlation between their residual-stream contributions. For the input matrix \texttt{mlp.c\_fc}, the vector contributions of the bottleneck units are summed before GELU, so we measure their pairwise contextual finite difference across the nonlinearity.

\subsection{\texttt{c\_proj}: Linear Contributions Are Usually Weakly Correlated}

\begin{figure}[H]
    \centering
    \includegraphics[width=0.72\linewidth]{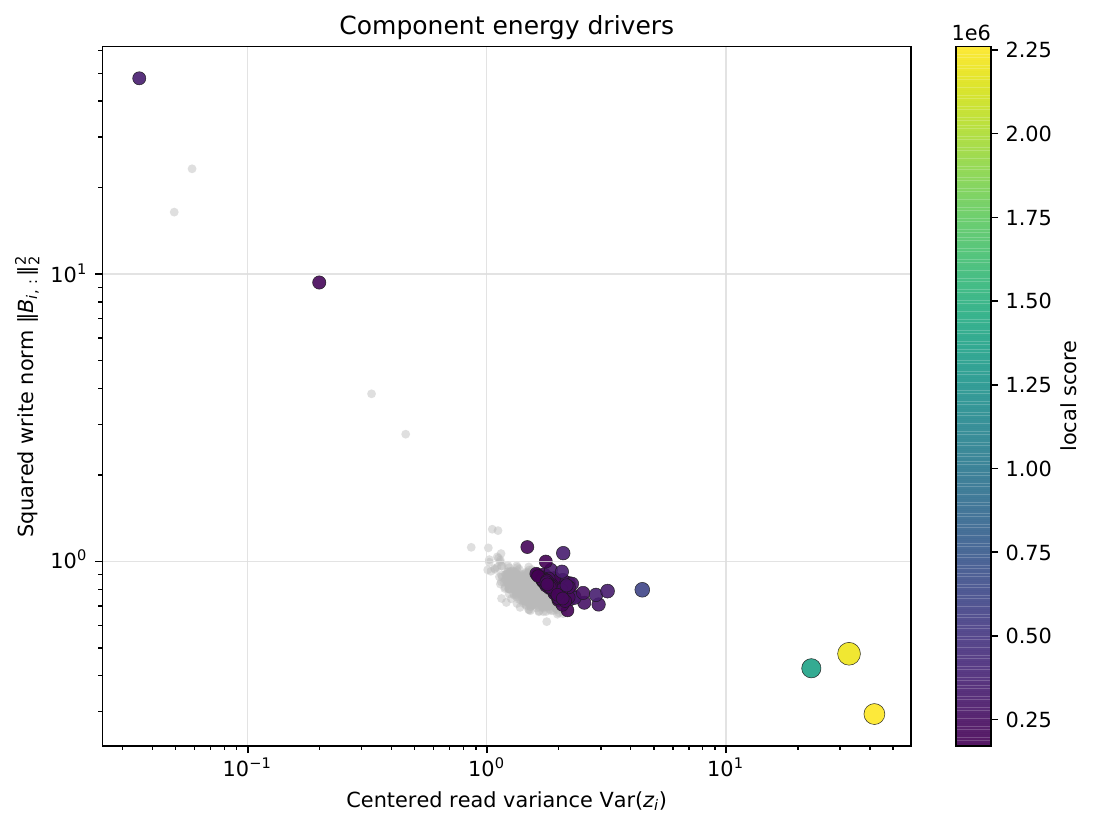}

    \vspace{4pt}
    \includegraphics[width=\linewidth]{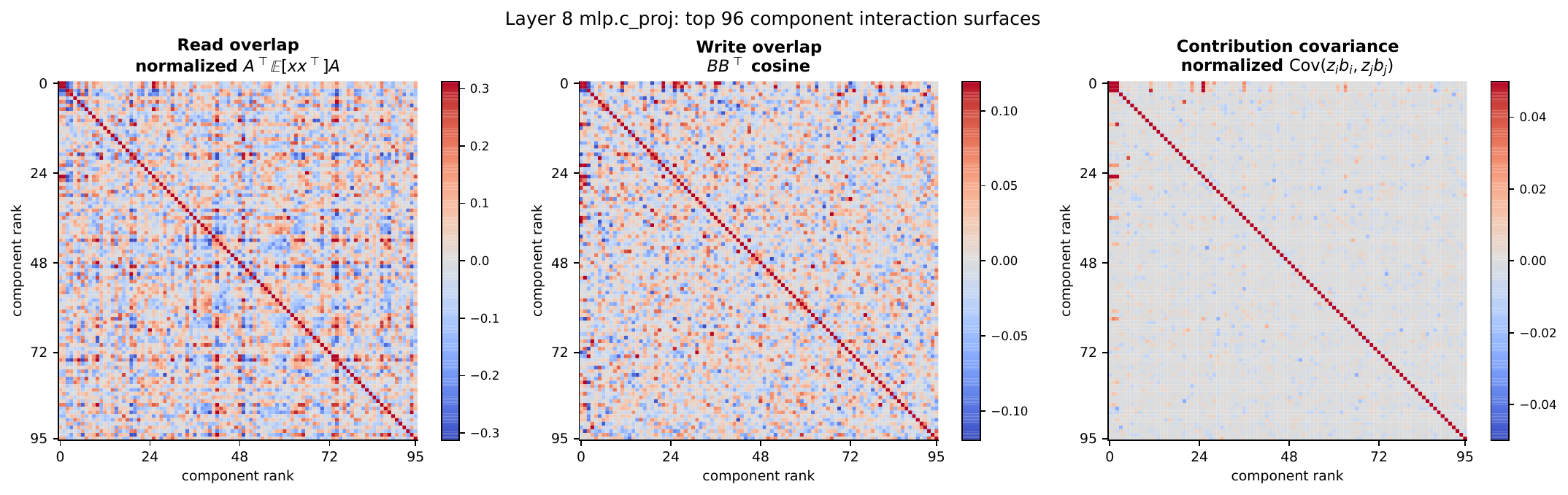}
    \caption{Selection and overlap structure for layer-8 \texttt{mlp.c\_proj}. Top: read variance versus squared write norm; the 96 bottleneck units with highest output-variance energy are highlighted. Bottom: read correlation, write-direction cosine, and complete contribution correlation for those units. Read and write structure is visible separately but largely disappears when combined into full residual contributions.}
    \label{fig:cproj-overlap-structure}
\end{figure}

The linear analysis uses all 768 bottleneck units of the layer-8 \texttt{mlp.c\_proj} factorization with $s=0.75$ and 2,048 held-out token activations. Bottleneck unit $i$ reads a scalar $z_i=\vh\mA_{:,i}$ and contributes the vector $z_i\mB_{i,:}$ to the residual stream. We define its output-variance energy as
\[
    \operatorname{Var}(z_i)\lVert \mB_{i,:}\rVert_2^2.
\]
The 96 highest-energy bottleneck units are retained for the readable matrix views in Figure~\ref{fig:cproj-overlap-structure}; distributional statistics use all 768 units.

For each bottleneck-unit pair, we measure absolute token-level read correlation, absolute cosine similarity between write directions, and the correlation of the complete residual contributions. Read and write overlap are each common in isolation, but they rarely occur for the same pair. Across all units, the 95th-percentile absolute read correlation is $0.207$ and the 95th-percentile absolute write cosine is $0.071$, whereas the 95th-percentile absolute contribution correlation is only $0.0049$; even its 99th percentile is $0.0089$.

\begin{figure}[H]
    \centering
    \includegraphics[width=\linewidth]{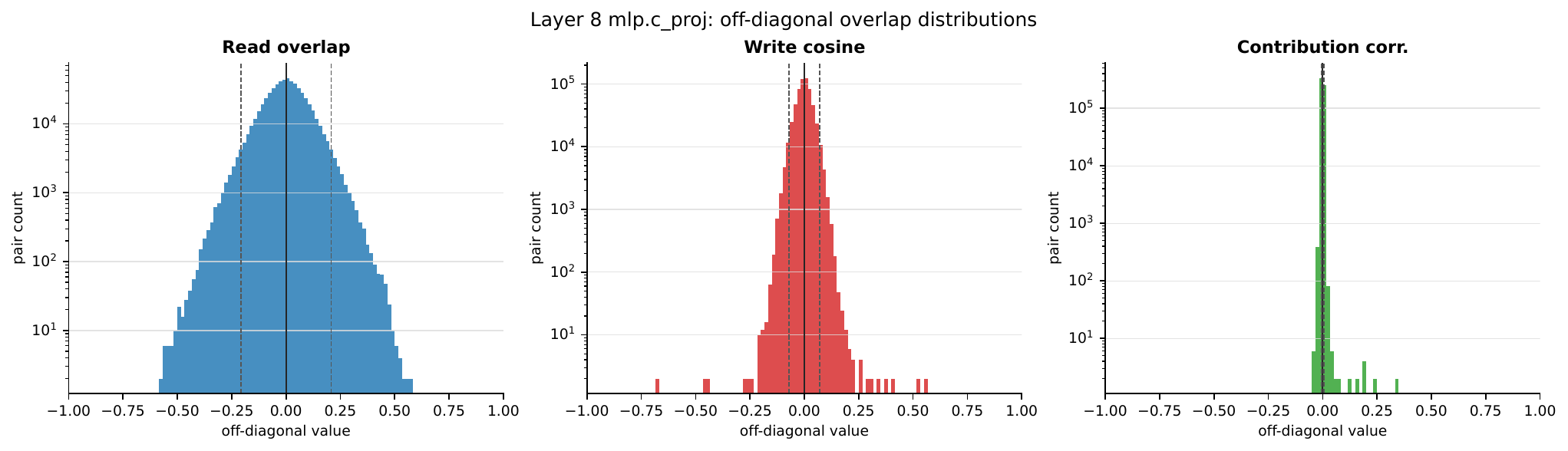}

    \vspace{4pt}
    \includegraphics[width=0.82\linewidth]{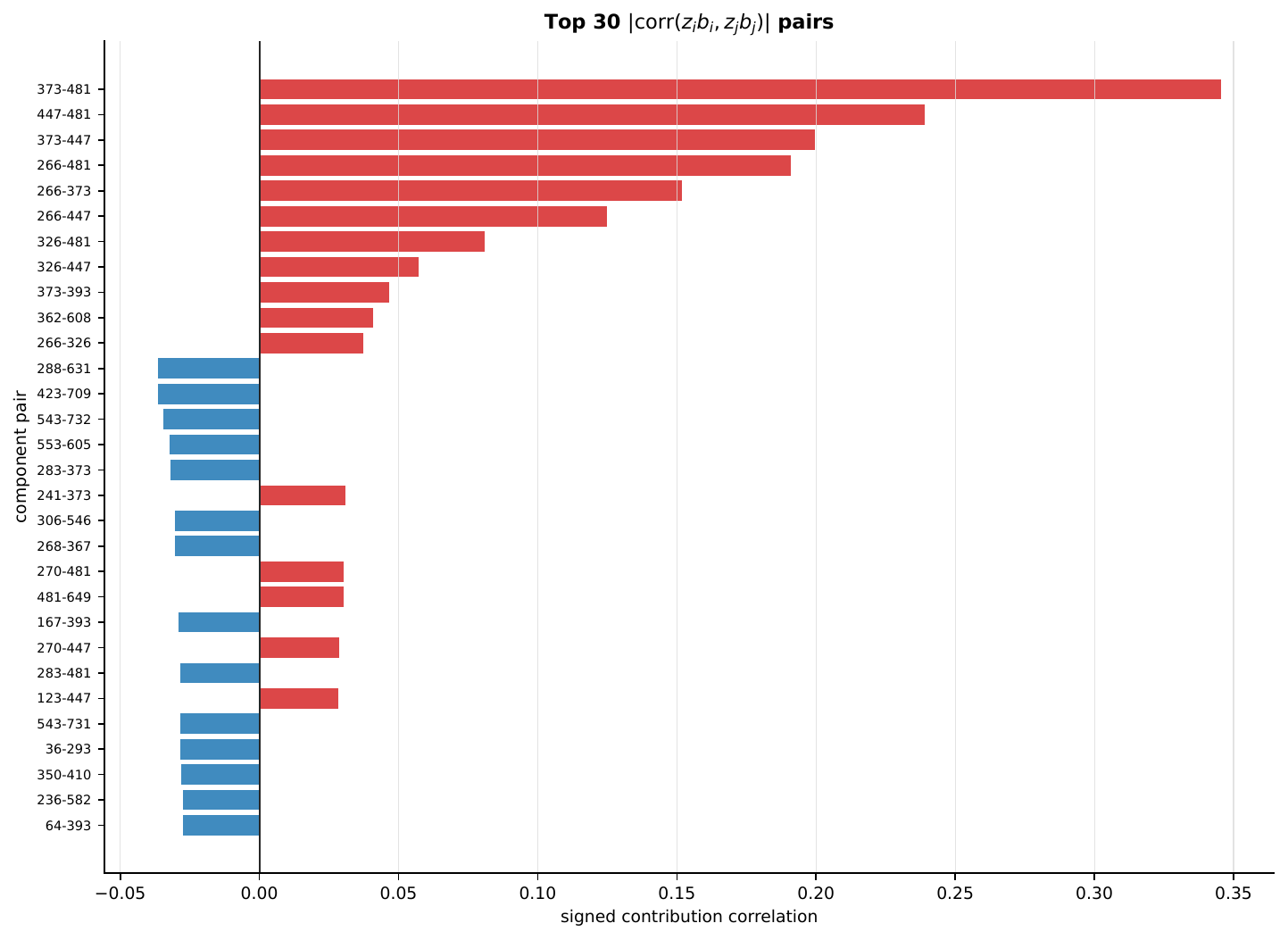}
    \caption{Linear bottleneck-unit dependence in layer-8 \texttt{mlp.c\_proj}. Top: off-diagonal pair distributions over all 768 units. Bottom: the 30 pairs with largest absolute contribution correlation. Most contribution correlations concentrate near zero, while a small tail is organized around recurring units; c373--c481 reaches $0.345$.}
    \label{fig:cproj-overlap-tail}
\end{figure}

The exceptional tail is compact. Bottleneck units 373, 447, 481, and 266 recur among the strongest positive pairs; the leading c373--c481 pair has contribution correlation $0.345$. In that pair, opposed read patterns combine with opposed write directions, so the two sign reversals yield positively correlated residual contributions. Thus, linear \texttt{c\_proj} bottleneck units are usually weakly correlated as complete pathways, with a small set of structured exceptions.

\subsection{\texttt{c\_fc} and GELU: Contextual Pair Interactions}

The nonlinear diagnostic uses the 1,536-bottleneck-unit layer-8 \texttt{mlp.c\_fc} factorization and 8,192 held-out token activations. We replay the tokens with the lower-layer \texttt{c\_fc} replacements installed, so the layer-8 inputs follow the sequentially reconstructed model. To obtain a readable pair matrix, we retain 128 units with the largest factor-rescaling-invariant pre-GELU downstream energy,
\[
    R_i
    =
    \mathbb{E}_t[z_{t,i}^2]
    \left\lVert \mB_{i,:}\mW_{\mathrm{proj}}\right\rVert_2^2,
\]
giving $128\times127/2=8{,}128$ unordered pairs.

Let $\vc_{t,i}=z_{t,i}\mB_{i,:}$ be unit $i$'s vector contribution to the \texttt{c\_fc} output, and let
\[
    \vh_t
    =
    \vb_{\mathrm{fc}}+\sum_a\vc_{t,a}
\]
be the complete reconstructed preactivation. For each pair, we compute the contextual finite difference
\[
\begin{aligned}
    \boldsymbol{\Delta}_{t,ij}
    ={}&
    \operatorname{GELU}(\vh_t)
    -\operatorname{GELU}(\vh_t-\vc_{t,i})
    -\operatorname{GELU}(\vh_t-\vc_{t,j})\\
    &+\operatorname{GELU}(\vh_t-\vc_{t,i}-\vc_{t,j}).
\end{aligned}
\]
We then project this interaction through the original MLP output matrix,
$\vq_{t,ij}=\boldsymbol{\Delta}_{t,ij}\mW_{\mathrm{proj}}$. Define the contextual marginal residual effect of unit $i$ as
\[
    \vm_{t,i}
    =
    \left[
    \operatorname{GELU}(\vh_t)
    -\operatorname{GELU}(\vh_t-\vc_{t,i})
    \right]\mW_{\mathrm{proj}},
    \qquad
    E_i=\sum_t\lVert\vm_{t,i}\rVert_2^2.
\]
The normalized interaction magnitude is
\[
    R_{ij}
    =
    \frac{
        \left(\sum_t\lVert\vq_{t,ij}\rVert_2^2\right)^{1/2}
    }{
        (E_iE_j)^{1/4}
    }.
\]

\begin{figure}[t]
    \centering
    \includegraphics[width=0.72\linewidth]{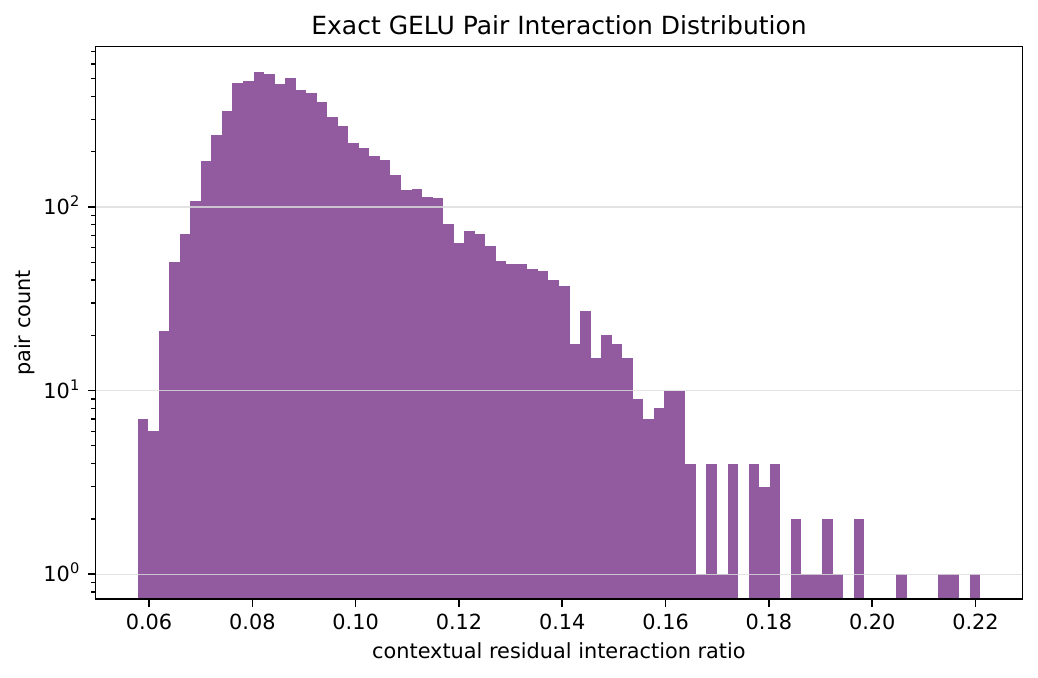}

    \vspace{4pt}
    \includegraphics[width=0.74\linewidth]{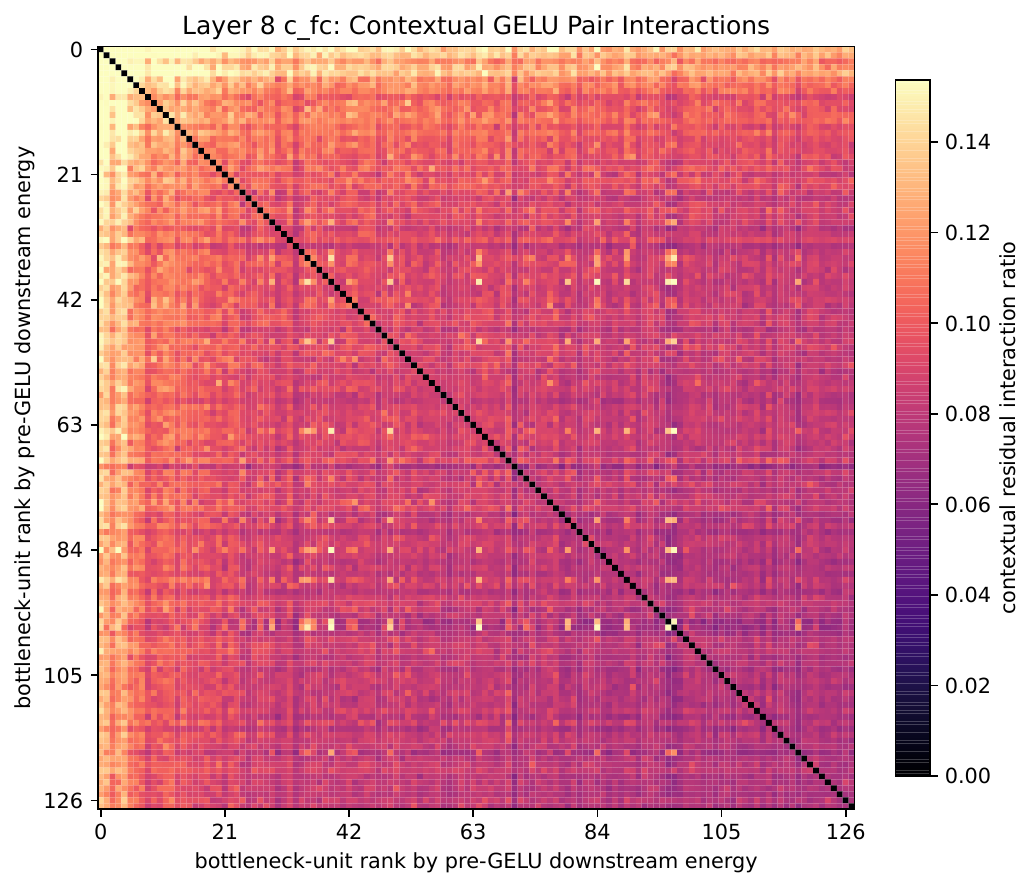}
    \caption{Contextual GELU-mediated interactions among 128 high-energy layer-8 \texttt{mlp.c\_fc} bottleneck units. Top: distribution of residual-stream interaction ratios over all 8,128 unordered pairs. Bottom: the same ratios ordered by pre-GELU downstream energy. Interactions are broadly distributed, with stronger values concentrated around several recurring units and a smaller set of pair-specific cells.}
    \label{fig:cfc-gelu-interactions}
\end{figure}

Pair interactions are widespread but typically moderate. Across the 8,128 pairs, mean $R$ is $0.093$, the median is $0.089$, the 95th percentile is $0.132$, the 99th percentile is $0.154$, and the maximum is $0.221$. In total, 2,130 pairs have $R\geq0.10$, 113 have $R\geq0.15$, and four have $R\geq0.20$. Figure~\ref{fig:cfc-gelu-interactions} shows a dense central band together with a thinner upper tail. The brighter first rows and columns indicate that several high-energy units interact moderately with many partners, while isolated bright cells reveal additional pair-specific effects.

\begin{figure}[H]
    \centering
    \includegraphics[width=0.82\linewidth]{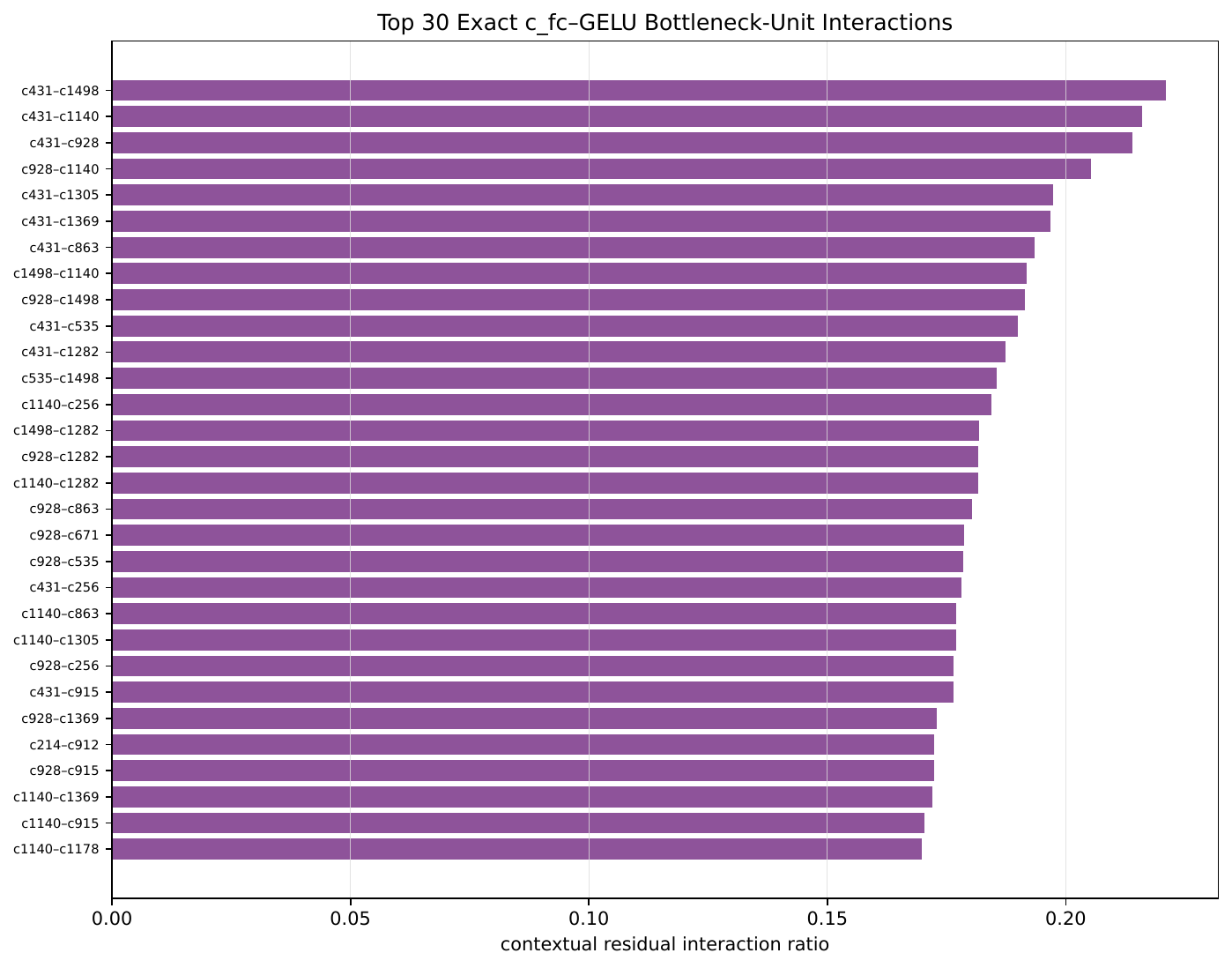}
    \caption{The 30 strongest contextual GELU-mediated pairs, ranked by residual-stream interaction ratio. Units c431, c928, c1140, and c1498 recur throughout the upper tail. The leading c431--c1498 pair reaches $R=0.221$; its interaction residual RMS is $2.096$, compared with marginal residual RMS values $11.127$ and $8.089$ for the two units.}
    \label{fig:cfc-gelu-top-pairs}
\end{figure}

The two MLP locations therefore show complementary structure. After GELU, most \texttt{c\_proj} bottleneck paths make weakly correlated residual-stream contributions, with a compact set of exceptions. Across GELU, the \texttt{c\_fc} finite differences reveal a dense field of moderate interactions and a thinner tail organized around recurring high-energy units. Thus, output-projection paths are usually well described by their additive contributions, whereas the effects of some input-projection paths are more informative when examined together with their strongest interaction partners.

\end{document}